\documentclass[sn-basic]{sn-jnl}

\jyear{2022}%

\theoremstyle{thmstyleone}%
\newtheorem{theorem}{Theorem}
\newtheorem{proposition}[theorem]{Proposition}%

\theoremstyle{thmstyletwo}%
\newtheorem{example}{Example}%
\newtheorem{remark}{Remark}%

\theoremstyle{thmstylethree}%
\newtheorem{definition}{Definition}%

\usepackage{amsmath}
\usepackage{amsthm}
\usepackage{float} 

\DeclareMathOperator*{\argmax}{arg\,max}
\DeclareMathOperator*{\argmin}{arg\,min}
\newcommand{\E}{\mathbb{E}}

\newcommand{\CE}{\mathsf{CE}}
\newcommand{\ECE}{\mathsf{ECE}}
\newcommand{\PL}{\mathsf{PL}}

\newcommand{\ch}{\hat{c}} 
\newcommand{\Ch}{\hat{C}}
\newcommand{\ph}{\hat{p}} 
\newcommand{\Ph}{\hat{P}}
\newcommand{\yb}{\bar{y}}

\newcommand{\pb}{\bar{p}}

\newcommand{\vB}{\mathbf{B}} 
\newcommand{\vH}{\mathbf{H}}
\newcommand{\vA}{\mathbf{A}}
\newcommand{\vOne}{\mathbf{1}}

\newcommand{\cC}{\mathcal{C}} 
\newcommand{\cB}{\mathcal{B}}
\newcommand{\cH}{\mathcal{H}}
\newcommand{\cA}{\mathcal{A}}
\newcommand{\cX}{\mathcal{X}}
\newcommand{\sR}{\mathbb{R}} 

\usepackage[normalem]{ulem}
\usepackage{color}
\definecolor{orange}{RGB}{255,127,0}
\definecolor{brown}{RGB}{150,70,0}
\definecolor{green}{RGB}{127,255,127}
\definecolor{mediumgreen}{RGB}{63,167,63}
\definecolor{darkgreen}{RGB}{0,127,0}
\definecolor{blue}{RGB}{127,127,255}
\definecolor{lightblue}{RGB}{150,150,255}
\definecolor{darkblue}{RGB}{0,0,127}
\definecolor{red}{RGB}{255,90,90}
\definecolor{grey}{RGB}{127,127,127}
\definecolor{pink}{RGB}{255,180,180}

\newcommand{\modified}[1]{{\textcolor{black} {#1}}}
\newcommand{\inserted}[1]{{\textcolor{black}{#1}}}
\newcommand{\insertnew}[1]{{\textcolor{black}{#1}}}

\usepackage[utf8]{inputenc} 
\usepackage[T1]{fontenc}    
\usepackage{hyperref}       
\usepackage{url}            
\usepackage{booktabs}       
\usepackage{amsfonts}       
\usepackage{nicefrac}       
\usepackage{microtype}      
\usepackage{xcolor}         

\usepackage{adjustbox}
\usepackage{subcaption}

\newtheorem{innercustomgeneric}{\customgenericname}
\providecommand{\customgenericname}{}
\newcommand{\newcustomtheorem}[2]{%
  \newenvironment{#1}[1]
  {%
   \renewcommand\customgenericname{#2}%
   \renewcommand\theinnercustomgeneric{##1}%
   \innercustomgeneric
  }
  {\endinnercustomgeneric}
}

\newcustomtheorem{customthm}{Theorem}
\newcustomtheorem{customlemma}{Lemma}

\begin{document}


\title[Fit-on-Test View on Evaluating Calibration of Classifiers]{On the Usefulness of the Fit-on-Test View on Evaluating Calibration of Classifiers}



\author*[1]{\fnm{Markus} \sur{K\"angsepp}}\email{markus.kangsepp@ut.ee}

\author[1]{\fnm{Kaspar} \sur{Valk}}\email{kaspar.valk@ut.ee}

\author[1]{\fnm{Meelis} \sur{Kull}}\email{meelis.kull@ut.ee}

\affil[1]{\orgdiv{Institute of Computer Science}, \orgname{University of Tartu}, \orgaddress{\street{Narva mnt}, \city{Tartu}, \postcode{51009}, \state{Tartumaa}, \country{Estonia}}}

\abstract{
Calibrated uncertainty estimates are essential for classifiers used in safety-critical applications.
\insertnew{
If a classifier is uncalibrated, then there is a unique way to calibrate its uncertainty using the idealistic true calibration map corresponding to this classifier. 
Although the true calibration map is typically unknown in practice, it can be estimated with many post-hoc calibration methods which fit some family of potential calibration functions on a validation dataset.
This paper examines the connection between such post-hoc calibration methods and calibration evaluation.
Despite the negative connotations of fitting on test data in machine learning, we claim that fitting calibration maps on test data as part of the calibration evaluation process is a method worth considering, and we refer to this view as fit-on-test.
This view enables the usage of any post-hoc calibration method as an evaluation measure, unlocking missed opportunities in development of evaluation methods. We prove that even ECE, which is the most common calibration evaluation method, is actually a fit-on-test measure. This observation leads us to a new method of tuning the number of bins in ECE with cross-validation. Fitting on test data can lead to test-time overfitting, and therefore, we discuss the limitations and concerns with the fit-on-test view.
Our contributions also include: (1) enhancement of reliability diagrams with diagonal filling; (2) development of new calibration map families PL and PL3; and (3) an experimental study of which families perform strongly both as post-hoc calibrators and calibration evaluators.}}




\maketitle

\section{Introduction}

When classifiers are incorporated into safety-critical applications, it is essential that the predictions of these classifiers would involve reliable uncertainty estimates. 
If the predictions are over-confident, then this can cause costly errors, such as an autonomous vehicle getting into an accident.
If the predictions are under-confident, then this can result in a failure of the system to fulfill its task, e.g. an autonomous car would move too slowly to mitigate the over-estimated risks.
Therefore, classifiers are expected to report calibrated uncertainty in the form of class probability estimates.
A probabilistic classifier is considered calibrated, if in the groups of similar predictions the average prediction is in an agreement with the actual class proportions. 
For example, in binary classification this implies that if the classifier predicts 80\% probability to be positive for each of a set of 100 instances, then 80 of these instances are expected to be truly positives.
Most learning algorithms result in classifiers that are not well-calibrated and need dedicated post-hoc calibration methods to be applied \citep{niculescu2005predicting,Guo2017}. 

Progress in developing methods to get calibrated classifiers can only be made if we have reliable methods for evaluating calibration.
In binary classification, the most common way of estimating a classifier's calibration is through reliability diagrams and ECE (estimated calibration error, also known as the expected calibration error\footnote{Following \citet{sweep}, we prefer `estimated' to avoid confusion with the true calibration error which involves an expectation.})  \citep{murphy1977reliability,Broecker2011EstimatingRA, naeini2015}.
In reliability diagrams, many instances with similar predicted probabilities are binned together to get an estimate of the true calibrated probability in each bin by averaging the corresponding class labels. 
However, there is no consensus, how to place the bins in reliability diagrams and how many bins there should be \citep{sweep}. 
Usually 10, 15, or 20 bins are used \citep{naeini2015,Guo2017}.  
Bins are placed with equal width, so that they take up an equal chunk in the probability space, or with equal size, so that they each contain an equal number of predictions. 
The choice of binning can drastically impact the shape of the reliability diagram and alter the estimated calibration error~\citep{sweep, kumar2019neurips, Nixon2019MeasuringCI}. 
Failure to measure calibration reliably leads to problems deciding which classifier is better calibrated or which method of post-hoc calibration is better. This in turn harms the performance of safety-critical systems.

\insertnew{Even though there are multiple existing works published on methods of evaluating calibration, i.e. \cite{Widmann2019CalibrationTI,Zhang2020MixnMatchEA}, these have not exploited the more direct link between post-hoc calibration and evaluation, which we refer to as the fit-on-test paradigm.} According to this paradigm, any post-hoc calibration method can be repurposed for calibration evaluation by applying it on the test data (not on the validation data as in post-hoc calibration) and then using it as a plug-in estimator of calibration error. 


The contributions of this paper are the following:
\begin{itemize}
    \item We introduce the fit-on-test paradigm of evaluating calibration, showing that any post-hoc calibration method can also be used for evaluating calibration (Section~\ref{subsec:fot_estimation});
    \item We prove that the classical binning-based ECE measure follows from the fit-on-test paradigm using a particular calibration map family (Section~\ref{subsec:ece_is_fot});
    \item Exploiting this fact, we show how cross-validation can be used for optimising the number of bins in ECE (Section~\ref{sec:cv_n_bins});
    \item We demonstrate shortcomings in the common visualisations of reliability diagrams and propose \emph{reliability diagrams with diagonal filling} (Section~\ref{subsec:fot_reliability});
    \item Using the fit-on-test paradigm, we develop new methods PL and PL3 of evaluating calibration using continuous piecewise linear functions (Section~\ref{sec:PL_RD});
    \item We clarify the methodology of assessing calibrators and calibration evaluators (Section~\ref{sec:eval_methods_of_cal}) and introduce the usage of pseudo-real data for this purpose (Section~\ref{sec:experiments});
    \item We perform experimental comparisons to find out which families of calibration maps result in better post-hoc calibration, better reliability diagrams, and better approximations of calibration errors (Section~\ref{sec:experiments}).
     \insertnew{\item We discuss the limitations of the fit-on-test paradigm (Section~\ref{sec:discussion}).}
\end{itemize}

\section{Related Work}

\insertnew{In this section, an overview of different approaches to calibration error evaluation is given. 
Evaluation of calibration has been the main focus of several works after the introduction of reliability diagrams and ECE~\citep{murphy1977reliability, Broecker2011EstimatingRA, naeini2015}.
To start with, \citet{vaicenavicius2019evaluating} proposed a more general definition of calibration and a method to perform statistical calibration tests based on binning.
\citet{Widmann2019CalibrationTI} proposed the kernel calibration error for calibration evaluation in multi-class classification.
\citet{sweep} proposed a method which chooses the maximal number of bins such that it leads to a monotonically increasing reliability diagram. 
\citet{popordanoska2022consistent} proposed estimating multi-class calibration error using kernel density estimation with Dirichlet kernels. 
\citet{popordanoska2023consistent} proposed Kullback-Leibler calibration error, allowing one to estimate all proper calibration errors and refinement terms.}

\insertnew{
The research on evaluating calibration has gone hand-in-hand with the research on post-hoc calibration, which aims to learn a calibration map transforming the classifier's output probabilities into calibrated probabilities.
Many papers contributed to both post-hoc calibration and evaluation.
\citet{naeini2015} proposed BBQ and used ECE to evaluate calibration.
\citet{Guo2017} proposed temperature, vector and matrix scaling and used the reliability diagrams and ECE for evaluating confidence in multi-class classification. Confidence stands for the probability of a target class, leaving out all the other probabilities and class-wise relations. This is also referred to as top-label calibration error by \citet{kumar2019neurips}.
\citet{Kull2019BeyondTS} proposed Dirichlet calibration and the notion of classwise-calibration error. Classwise-calibration error measures the calibration error for each class separately.
\citet{kumar2019neurips} proposed scaling-binning calibration, a new debiasing method for ECE and the notion of marginal calibration error. 
Marginal calibration error is similar to classwise-calibration error defined concurrently with 
\citet{Kull2019BeyondTS}, adding a possibility to control how much each class counts towards the error.
\citet{Zhang2020MixnMatchEA} proposed generic Mix-n-Match calibration strategies and used kernel density estimation (KDE) for estimating calibration error.
\citet{spline_gupta21} proposed a calibration evaluation metric based on the Kolmogorov-Smirnov test and a calibration method based on fitting splines.
\citet{xiong2023proximitybias} proposed proximity calibration (procal) for confidence calibration and proximity-informed ECE (PIECE). PIECE divides instances based on the representation space distance into proximity groups and uses this information to measure the calibration error of different proximity groups separately.
}

\section{Notation and Background}
\label{sec:estimating_CE}

\subsection{True Calibration Error}

\begin{figure}
    \begin{center}
    \begin{subfigure}[b]{0.33\columnwidth}
        \caption{}
        \includegraphics[width=\textwidth]{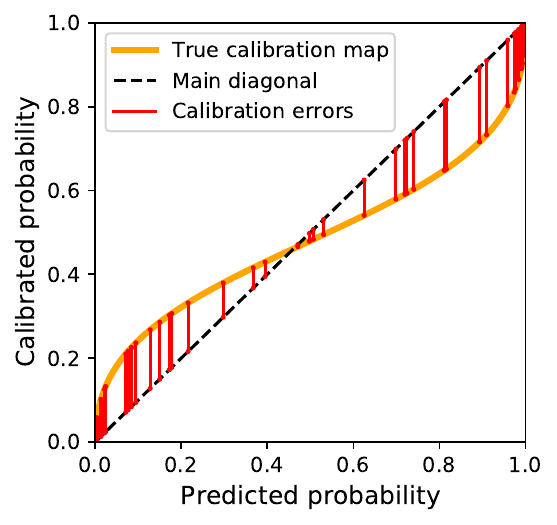}
        \label{fig:true_calibration_map}
    \end{subfigure}%
    ~ %
    \begin{subfigure}[b]{0.33\columnwidth}
        \caption{}
        \includegraphics[width=\textwidth]{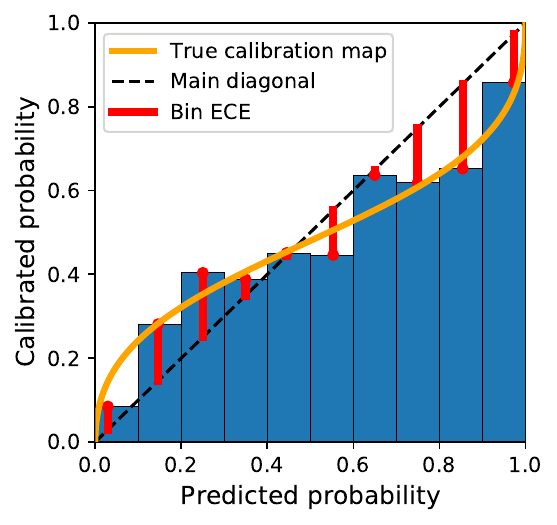}
        \label{fig:reliability_diagram_classic}
    \end{subfigure}%
    
    \caption{(a) True calibration map (orange line) vs the predicted probabilities (dashed line). Connecting lines show instance-wise miscalibration. (b) Reliability diagram consists of bars (blue) with the height of average label. The red lines show the error between the mean labels and predicted probabilities in each bin. 
    The diagrams are made with synthetic data (3000 data points, \modified{stratified sample of 50 data points from the bins shown for instance-wise errors, }\inserted{see Appendix~\ref{subsec:syn_experiments} for more details}).
    }
    \label{fig:1_calmap_reldiag}
    \end{center}
\end{figure}

We present the methods for binary classification, but Section~\ref{sec:multiclass} shows applicability to multi-class classification as well.
Consider a binary classifier $f:\cX\to[0,1]$ predicting the probabilities of instances to be positive.
Let $X\in\cX$ be a randomly drawn instance, $Y\in\{0,1\}$ its true class, and let us denote the model's predictions with $\Ph=f(X)$.
Every classifier $f$ has a corresponding \emph{true calibration map}, which could be used to perfectly calibrate the model: $c^*_f(\ph)=\E[Y\mid \Ph=\ph]$ (also known as the canonical calibration function \citep{vaicenavicius2019evaluating}).
For evaluation of calibration, consider a test dataset with instances $x_1,\dots,x_n\in\cX$ and true labels $y_1,\dots,y_n\in\{0,1\}$, and denote the predictions by $\ph_i=f(x_i)$.
The true calibration error (CE) is the model's average violation of calibration; it could be defined on the overall test distribution \inserted{as $\E[\vert c^*_f(\Ph)-\Ph\vert^\alpha]$} \citep{kumar2019neurips} but we define it for the test \inserted{dataset}:
\begin{align}\label{eq:ce}
\CE^{(\alpha)}=\frac{1}{n}\sum_{i=1}^n \vert c^*_f(\ph_i)-\ph_i\vert^\alpha 
\end{align}
where $\alpha=1$ corresponds to absolute error (MAE) and $\alpha=2$ to squared error (MSE). 
Figure~\ref{fig:true_calibration_map}  shows an example of a true calibration map, where each red line shows the violation of calibration corresponding to a particular data point, and the average length of red lines equals to CE.

\subsection{Reliability Diagrams and ECE}
\label{sec:RD_ECE}

There are multiple ways to estimate calibration error. One of the most popular ways is using reliability diagrams~\citep{murphy1977reliability}.
The reliability diagram is a bar plot, where each bar contains a certain region of probabilities (\inserted{a bin}) and the bar height corresponds to the average label ($\yb_k$) in the \inserted{$k$-th} bin (Figure~\ref{fig:reliability_diagram_classic}). Each red line \inserted{in Figure~\ref{fig:reliability_diagram_classic}} shows the difference between the average label $\yb_k$ and the average prediction $\pb_k$ in the \inserted{$k$-th} bin.
The vector $\vB=(B_1,\dots,B_{b+1})$ provides the bin boundaries $0=B_1<B_2<\ldots<B_b<B_{b+1}=1+\epsilon$, resulting in bins $[B_1,B_2),\dots,[B_b,B_{b+1})$, where $\epsilon$ is an infinitesimal to ensure that $1\in [B_b,B_{b+1})$.
Thus, $\yb_k=\frac{1}{n_k}\sum_{i:\ph_i\in[B_k,B_{k+1})} y_i$ and $\pb_k=\frac{1}{n_k}\sum_{i:\ph_i\in[B_k,B_{k+1})} \ph_i$ where $n_k=\vert\{i: \ph_i\in[B_k,B_{k+1})\}\vert$ is the size of bin $k$. 
The bins can be either equal size (each bin has the same \inserted{number} of instances), or equal width (each bin covers the equal region in the probability space).

Based on the reliability diagrams (Figure~\ref{fig:reliability_diagram_classic}), the estimated calibration error (ECE) \citep{naeini2015} is a weighted average between the mean accuracy and the mean probability in each bin:
\begin{align} \label{eq:ece}
\ECE^{(\alpha)}_\vB=\frac{1}{n}\sum_{k=1}^b n_k\cdot\vert\yb_k-\pb_k\vert^\alpha.
\end{align}

The binning-based ECE is known to be biased \citep{Broecker2011EstimatingRA, Ferro2012ABD} with 
\modified{$\E[\ECE^{(\alpha)}_\vB]\neq\E[\CE^{(\alpha)}]$}, hence in our experiments we use debiasing as proposed by \citet{kumar2019neurips}.

\subsection{Calibration Evaluation for Multi-Class Classification}
\label{sec:multiclass}
In contrast to binary classification, there are multiple different definitions of calibration for multi-class tasks: 
\modified{
    \begin{itemize}
        \item a binary classifier is calibrated if all predicted probabilities to be positive are calibrated: $\Pr[Y=1\vert f(X)=\ph]=\ph$ \inserted{for all $\ph\in[0,1]$}; 
        \item a multi-class classifier is class-k-calibrated if all the predicted probabilities of class $k$ are calibrated: $\Pr[Y=k\vert f_k(X)=\ph]=\ph$ \inserted{for all $\ph\in[0,1]$} \citep{Kull2019BeyondTS, kumar2019neurips,Nixon2019MeasuringCI}; 
        \item a multi-class classifier is confidence calibrated if $\Pr[Y=\argmax f(X)\vert\max f(X)=\ph]=\ph$ \inserted{for all $\ph\in[0,1]$} \citep{Kull2019BeyondTS, Guo2017}. 
    \end{itemize}
}

However, in all of the above scenarios, we need to evaluate if the predicted and actual probabilities of an event are equal among all instances with shared predictions.
By redefining $Y=1$ and $Y=0$ to denote whether or not the event happened and $\Ph=f(X)$ to denote the estimated probability of that event, we have essentially reduced all three evaluation tasks to the first task of evaluating calibration in binary classification.
The shared definition of calibration then becomes: $\Pr[Y=1\vert f(X)=\ph]=\ph$ or equivalently, $\E[Y\vert f(X)=\ph]=\ph$,
This explains also why ECE has been applied to all those 3 scenarios.

\subsection{Post-hoc Calibration}
\label{sec:post_hoc}

\emph{Post-hoc calibration} is the task where the goal is to use a validation set to obtain an estimate $\ch$ of the true calibration map $c^*_f$ for a given uncalibrated classifier $f$.
Post-hoc calibration methods view the task basically as binary regression: given the predictions $\ph_1,\dots,\ph_n\in[0,1]$ and the corresponding true binary labels $y_1,\dots,y_n\in\{0,1\}$, find a `regression' model $\ch:[0,1]\to[0,1]$ that best predicts the labels from the predictions, evaluated typically by cross-entropy or mean squared error which in this context are respectively known as the log-loss and the Brier score - two members of the family of strictly proper losses \citep{brier1950verification}. 
\emph{Why are proper losses a good way of evaluating progress towards estimating the true calibration map?}
A common justification is that these losses have the virtue that they are minimised by the perfectly calibrated model $c^*_f$ \citep{kumar2019neurips}, that is:
$\argmin_{\ch(\ph)}\E[l(\ch(\ph),Y)\vert\Ph=\ph]=c^*_f(\ph)$ for any $\ph\in[0,1]$ and any strictly proper loss $l$.
However, this justification refers to the optimum only. Our following Theorem~\ref{thm:calmapfitting} makes even a stronger claim that a reduction of the expected loss $l$ leads to the same-sized improvement in how well $\ch(\ph)$ approximates $c^*_f(\ph)$, measured by any Bregman divergence $d:[0,1]\to[0,1]$ (here $d$ quantifies similarity between two binary categorical probability distributions, and it is a strictly proper loss when the label is its second argument, \modified{see details and proofs of the theorems in Appendix~\ref{sec:proof}}):
\begin{theorem} \label{thm:calmapfitting}
Let $d:[0,1]\times[0,1]\to\sR$ be any Bregman divergence and $\ch_1,\ch_2:[0,1]\to[0,1]$ be two estimated calibration maps. Then 
\begin{align*}
    \E\Bigl[d(\ch_1(\ph),Y)\vert\Ph=\ph\Bigr]-\E\Bigl[d(\ch_2(\ph),Y)\vert\Ph=\ph\Bigr]
    \\=d\Bigl(\ch_1(\ph),c^*_f(\ph)\Bigr)- d\Bigl(\ch_2(\ph),c^*_f(\ph)\Bigr).
\end{align*}
\end{theorem}


The above theorem involves expectations conditioned on $\ph$ which are typically impossible to estimate for any particular $\ph$ in isolation, because there is just one or very few instances with exactly the same predicted probability $\ph$.
Therefore, most post-hoc calibration methods minimize the empirical loss $\sum_{i=1}^{n}d(\ch(\ph_i),y_i)$ for $\ch$ in some sub-family $\cC$ within all possible calibration maps, using inductive biases such as assuming $c^*_f$ is monotonic (\emph{isotonic calibration}~\citep{isotonic}), or belongs to some parametric family, e.g. logistic functions (\emph{Platt scaling}~\citep{platt1999probabilistic}).

\section{The Fit-on-Test Paradigm} \label{sec:fit_on_the_test}

\subsection{Evaluation of Calibration Always Involves Estimation} \label{subsec:eval_involves_estimation}

The goal of evaluating calibration is to measure how far a classifier $f$ is from being perfectly calibrated, based on a given test set. 
Ideally, we would like to know for each test instance how far the prediction $\ph_i$ is from the corresponding perfectly calibrated probability $c^*_f(\ph_i)$.
The fundamental problem is that we can never directly observe $c^*_f(\ph_i)$, even on the test data.
Therefore, evaluation of calibration always involves some form of estimation, and one cannot measure the true calibration error precisely.

The standard ECE measure gets around this problem by introducing bins. 
The idea is that if there are sufficiently many instances in the bin $[B_k,B_{k+1})$, and the bin is narrow enough so that the corresponding perfectly calibrated probabilities $c^*_f(\ph_i)=\E[Y\mid\Ph=\ph_i]$ do not vary much within the bin, then one can estimate the calibration error in the bin as follows:
$$\frac{1}{n_k}\sum_{i:\ph_i\in[B_k,B_{k+1})}\vert\E[Y\mid\Ph=\ph_i]-\ph_i\vert^\alpha\approx\vert\yb_k-\pb_k\vert^\alpha$$
that is by the difference between the proportion of positives and the average prediction within the bin. 
If the bins are too narrow, then there are not sufficiently many instances in them, resulting in high variance of calibration error estimation. 
If the bins are too wide, then the corresponding perfectly calibrated probabilities $c^*_f(\ph_i)$ vary too much inside the bin, resulting in potential bias in the estimation.

\subsection{Fit-on-Test Estimation of Calibration Error} \label{subsec:fot_estimation}

Seeing the challenges of choosing a good binning for ECE and the existing attempts of improving over ECE~\citep{sweep}, we looked for alternatives in estimating the calibration error.
A classical and intuitive estimation method is \emph{plug-in estimation}, where the estimate is calculated using the same formula as the population statistic it is estimating.
We propose to use this for the true calibration error defined earlier as Eq.(\ref{eq:ce}), getting the plug-in estimator: 
\begin{align} \label{eq:plugin}
\widehat{\CE}^{(\alpha)}=\frac{1}{n}\sum_{i=1}^n \vert \ch(\ph_i)-\ph_i\vert^\alpha
\end{align}
where $\ch(\cdot)$ is some estimator of the function $c^*_f(\cdot)$.
The intuition is that in order to estimate the true calibration error we would first estimate the true calibration map. 
After that we can use the average discrepancy between the predictions and the corresponding estimated calibration map values as our estimate of the true calibration error.

Our task now is to find a way to estimate the true calibration map $c^*_f(\cdot)$.
Here it is very important to note that this estimation needs to be performed only using the given test set.  
This is because the goal of evaluating calibration is to do so based on a given test set.

Here we can turn to the existing literature on post-hoc calibration. Indeed, the goal of post-hoc calibration is also to estimate the true calibration map, except that the estimation is performed there on the validation set.
All we need to do is to take a post-hoc calibration method and apply it instead on test data. 
By this we can get an estimated calibration map $\ch(\cdot)$ which can be used within Eq.(\ref{eq:plugin}) to approximate calibration error.
As estimating a calibration map is essentially fitting a function, we refer to such plug-in estimation as the \emph{fit-on-test estimation of calibration error}. 
To summarise, {\bf any post-hoc calibration method can be applied on the test data and used within the plug-in estimator to turn it into a fit-on-test estimator of calibration error.} 
\insertnew{Note that this does not mean that all methods would be equally useful as plug-in estimators; more discussion about the limitations of a fit-on-test estimator is in Sections~\ref{subsec:fot_discussion} and~\ref{sec:discussion}.} 

\begin{figure}
     \begin{center}
        \includegraphics[width=\textwidth]{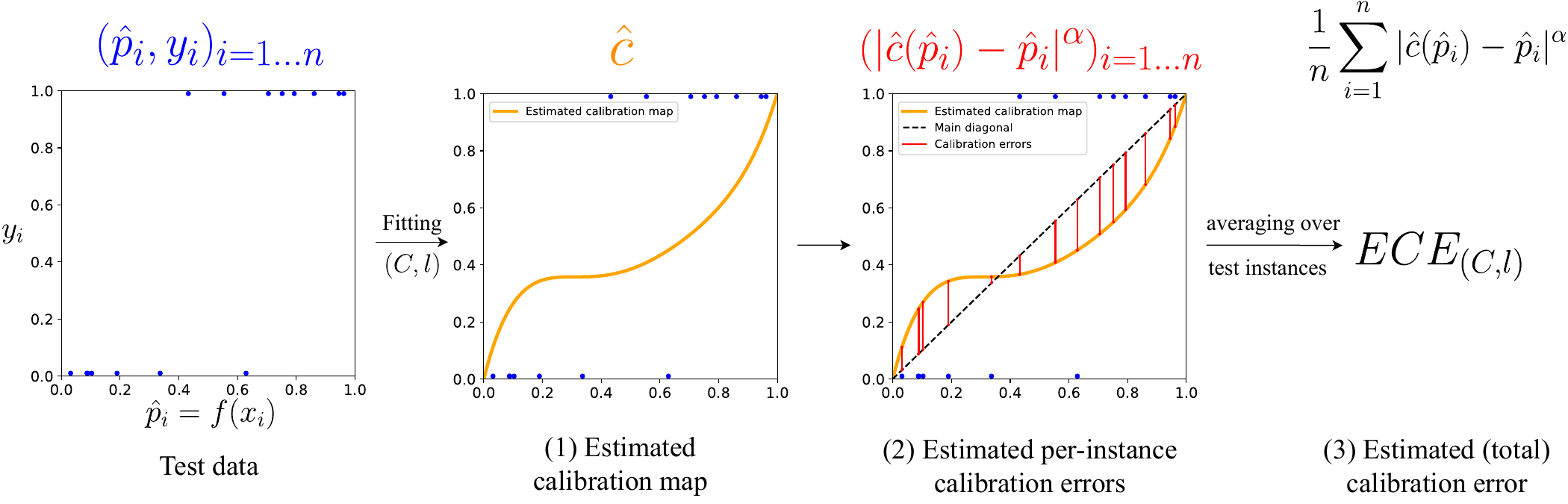}

    \caption{Fit-on-test estimation of calibration error: (1) a calibration map $\ch$ is obtained fitting a family of calibration maps $\cC$ by minimising the loss $l$ on the test data; (2) instance-wise calibration errors are estimated as distances of predictions from calibrated predictions; (3) overall calibration error is estimated as the average of instance-wise errors. The plots are illustrative, not based on real data.
    }
    \label{fig:fot_overview}
    \end{center}
\end{figure}

However, nothing prevents us from going beyond the set of existing post-hoc calibration methods.
Whenever we have some family $\cC$ of potential calibration map functions and some strictly proper loss $l$, we can define the corresponding fit-on-test calibration evaluation measure by first performing fitting on the test data:
\begin{align*}
\ch_{\text{fit-}(\cC,l)\text{-on-test}}=\argmin_{c\in\cC} \frac{1}{n}\sum_{i=1}^n l(c(\ph_i),y_i)
\end{align*}
and then using it within the plug-in estimator:
\begin{align*}
\ECE^{(\alpha)}_{\text{fit-}(\cC,l)\text{-on-test}}=\frac{1}{n}\sum_{i=1}^n \vert\ch_{\text{fit-}(\cC,l)\text{-on-test}}(\ph_i)-\ph_i\vert^{\alpha}.
\end{align*}
Fit-on-test estimation of calibration error has been visualised in Figure~\ref{fig:fot_overview} (illustration only, not on real data).

\subsection{Discussion}
\label{subsec:fot_discussion}

As the idea of using plug-in estimation is almost trivial, one might wonder why it has not been introduced before.
We guess this is partly due to the following potential concerns:
\begin{enumerate}
    \item Due to inevitable overfitting (or the generalisation gap) in any fitting process, we are bound to get our estimated $\ch$ closer to the observed labels than $c^*_f$ is. This bias can harm our capability of estimating the true calibration error $\vert c^*_f(\ph)-\ph\vert$;
    \item By choosing a particular family $\cC$ of functions to be used during the fitting process, we would potentially misjudge the calibration error in the cases where $c^*_f$ is not in this family. 
\end{enumerate}
It seems impossible to fully solve both problems at the same time: a more restrictive set of functions helps against overfitting and alleviates the first problem, but increases the second problem; 
a bigger set of functions helps against the second problem, but increases overfitting.
However, our experiments demonstrate that good tradeoffs are possible, using flexible families but still with relatively few parameters.

The classical binning-based ECE might seem to sidestep this problem and instead of estimating $c^*_f$ at all given points, it performs the comparison of bin averages $\pb$ and $\yb$.
Perhaps surprisingly though, it can be proved (see the next subsection) that the binning-based ECE can also be seen as a fit-on-test estimator of calibration error for a particular family of functions $\cC$ with the Brier score (MSE) as the loss function. Therefore, the above concerns are valid for the standard binning-based ECE as well.

\subsection{Classical binned ECE is also a fit-on-test estimator} \label{subsec:ece_is_fot}

Next we prove that the standard binning-based ECE measure as defined by Eq.(\ref{eq:ece}) is a fit-on-test estimator with a certain calibration map family $\cC$ that we will present in a moment.
Before this, we first introduce a bigger family $\cC_{\cB,\cH,\cA}^{(b)}=\{c_{(\vB,\vH,\vA)}\}$ of piecewise linear functions with $b$ pieces (or bins), parametrised by the following $3$ vectors:
\begin{itemize}
    \item  $\vB\in [0,1]^{b+1}$ - the boundaries of the $b$ pieces (or bins) with the constraint $0=B_1<B_2<\ldots<B_b<B_{b+1}=1+\epsilon$;
    \item $\vH\in\sR^{b}$ - values of the function at the boundaries $B_1,\dots,B_b$;
    \item $\vA\in\sR^{b}$ - slopes of linear functions within the bins.
\end{itemize}
The values of these functions can be calculated as follows: 
$$c_{(\vB,\vH,\vA)}(\hat{p})=\sum_{k=1}^b I[B_k\,{\leq}\,\hat{p}\,{<}\,B_{k+1}]\cdot (H_k+A_k(\hat{p}-B_k))$$ 
where $I[\cdot]$ is the indicator function.
Note that the resulting functions $c_{(\vB,\vH,\vA)}$ can be non-continuous because the right side of the bin $[B_k,B_{k+1})$ ends near the value $H_k+A_k(B_{k+1}-B_k)$ and nothing is preventing this value from being different than $H_{k+1}$ which is the left side of the bin $[B_{k+1},B_{k+2})$. 

\begin{figure}
     \begin{center}
     
     \includegraphics[width=\textwidth]{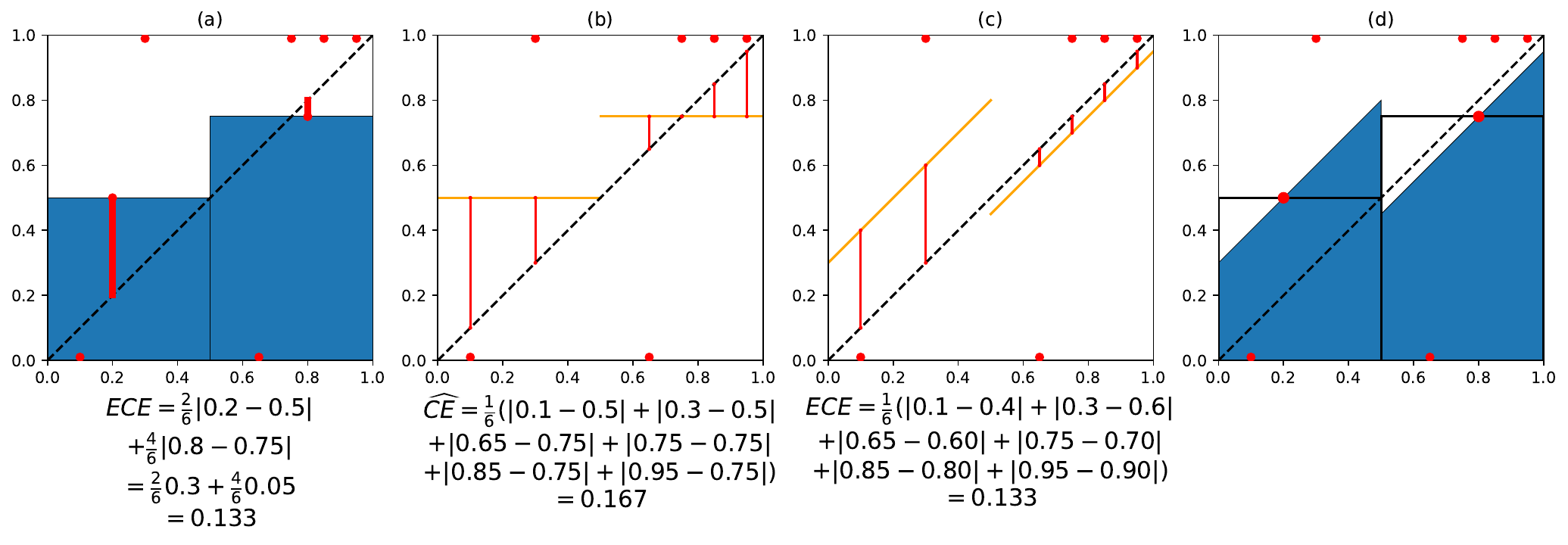}

    \caption{A synthetic example about how ECE can be viewed as a fit-on-test estimator of calibration error. (a) A reliability diagram of a test set with 6 instances (4 positives with predicted probabilities 0.3, 0.75, 0.85, 0.95 and 2 negatives with predicted probabilities 0.1, 0.65), yielding $ECE=0.133$; (b) a piecewise constant fit-on-test estimator yields $\widehat{CE}=0.167$; (c) a piecewise slope-1 fit-on-test estimator provably yields the same $ECE=0.133$ as the original in (a) while visualizing per-instance calibration errors also; (d) our proposed \emph{reliability diagram with diagonal filling} combines elements from (a) and (c).
    }
    \label{fig:ece_fitting}
    \end{center}
\end{figure}

It turns out that the classical binning-based ECE is a fit-on-test estimator of calibration error with respect to a particular subfamily of $\cC_{\cB,\cH,\cA}^{(b)}$ that we will describe next.
As ECE is calculated from the reliability diagrams that are piecewise constant, one might guess that this subfamily would contain all the piecewise constant (slope $0$) functions with a particular fixed binning $\vB$. 
However, this is not true.
To see this, consider a synthetic example in Figure~\ref{fig:ece_fitting}(a) with 6 instances (2 negatives and 4 positives) shown as red dots, and 2 bins $\vB=(0,0.5,1+\varepsilon)$. 
The traditional definition of ECE in Eq.(\ref{eq:ece}) yields $ECE=0.133$ in this example.
Fitting a piecewise constant function with binning $\vB$ by minimizing the Brier score results in the calibration map $\ch$ visualised in Figure~\ref{fig:ece_fitting}(b).
As the Brier score is a proper loss, the optimum is achieved by empirical averages of labels within each bin. 
Hence, the shape of the calibration map $\ch$ in Figure~\ref{fig:ece_fitting}(b) matches with the reliability diagram in Figure~\ref{fig:ece_fitting}(a).
However, if we now use the fit-on-test estimator of calibration error as the average length of vertical red lines in Figure~\ref{fig:ece_fitting}(b), then we get $\widehat{CE}=0.167$, which is different from $ECE=0.133$ with the same binning.

Instead, ECE with the binning $\vB$ is actually a fit-on-test estimator of calibration error with respect to the subfamily $\cC^{(b)}_{(\vB,\cH,\vOne)}$ which contains all piecewise linear functions with slope $1$ (i.e. a 45-degrees ascending slope) in each of the bins.
In other words, $\cC_{(\vB,\cH,\vOne)}$ contains all functions with the fixed binning $\vB$, fixed slopes $\vA=\vOne=(1,1,\dots,1)$ i.e. slope $1$ for each of the $b$ bins, and any heights $\vH\in\sR^b$ for the left-side boundaries of these bins.
Fitting this family for our example results in the calibration map visualised in Figure~\ref{fig:ece_fitting}(c).
Using the fit-on-test estimator of calibration error we now get exactly the standard $ECE=0.133$ as the average length of vertical red lines.
This can be confirmed visually, seeing that each of the vertical red lines in Figure~\ref{fig:ece_fitting}(c) has exactly the same length as the bin-specific red line in the standard reliability diagram of Figure~\ref{fig:ece_fitting}(a).
These lengths are equal because the diagonal and the calibration map have both slope equal to $1$.
The following theorem confirms this by proving that the standard ECE with binning $\vB$ can be seen as a fit-on-test estimator of calibration error.
\begin{theorem}
\label{thm:slope1}
Consider a predictive model with predictions $\ph_1,\dots,\ph_n\in[0,1]$ on a test set with actual labels $y_1,\dots,y_n$ and a binning $\vB$ with $b\geq 1$ bins and boundaries $0=B_1<\dots<B_{b+1}=1+\epsilon$.
Then for any $\alpha>0$, the measure $\ECE^{(\alpha)}_{\vB}$ as defined by Eq.(\ref{eq:ece}) is equal to the fit-on-test estimator of calibration error using the family $\cC_{(\vB,\cH,\vOne)}$ and fitting the Brier score:
\modified{$$\ECE^{(\alpha)}_{\vB}=\frac{1}{n}\sum_{i=1}^n \vert \ch(\ph_i)-\ph_i\vert ^{\alpha}$$}
\modified{$$\text{\it where}\quad\ch=\argmin_{c\in\cC_{(\vB,\cH,\vOne)}} \frac{1}{n}\sum_{i=1}^n (c(\ph_i)-y_i)^2.$$}
Furthermore, $\ch(\pb_k)=\yb_k$ for $k=1,\dots,b$, where $\pb_k$ and $\yb_k$ are the average $\ph_i$ and $y_i$ in the bin $[B_k,B_{k+1})$.
\end{theorem}
\begin{proof} 
See the Supplementary Material.
\end{proof}

\subsection{Fit-on-test reliability diagrams} \label{subsec:fot_reliability}

Reliability diagrams are a common way of evaluating calibration of classifiers visually.
Next we discuss the shortcomings of existing reliability diagrams and propose enhancements to them.

\begin{figure}
    \begin{center}
    \begin{subfigure}[b]{0.8\columnwidth}
        \includegraphics[width=\textwidth]{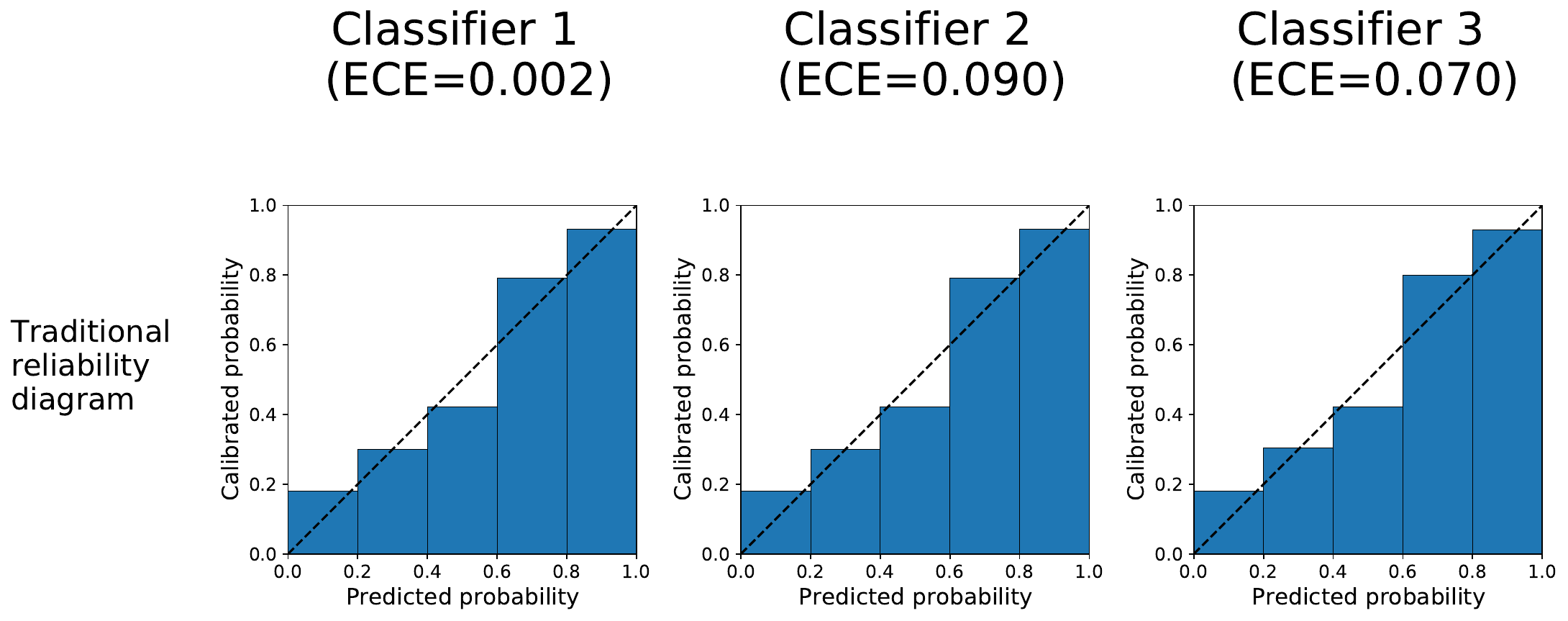}
        \label{fig:reliability_diagram_flat}
    \end{subfigure}%

    \begin{subfigure}[b]{0.8\columnwidth}
        \includegraphics[width=\textwidth]{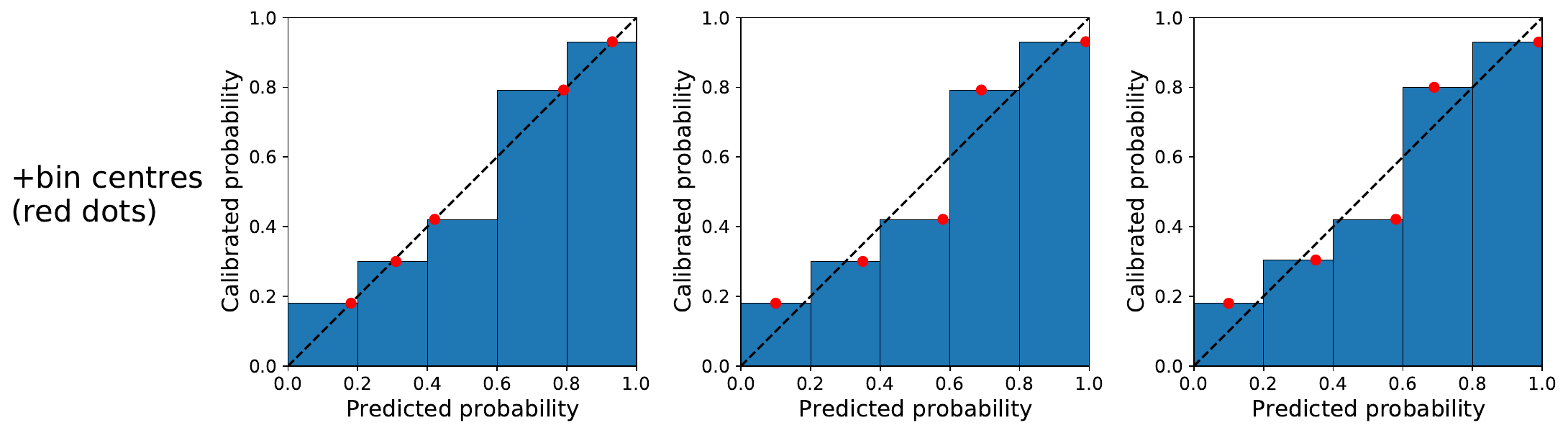}
        \label{fig:reliability_diagram_flat_red}
    \end{subfigure}%

    \begin{subfigure}[b]{0.8\columnwidth}
        \includegraphics[width=\textwidth]{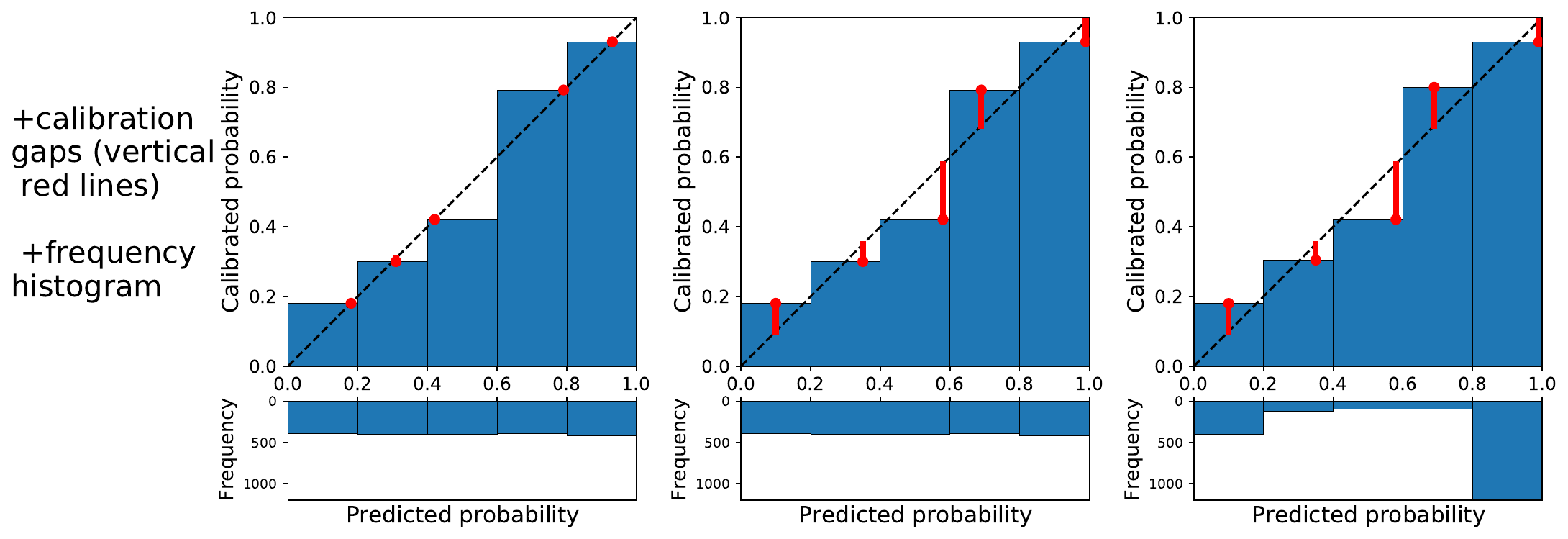}
        \label{fig:reliability_diagram_flat_red_hist}
    \end{subfigure}%

    \begin{subfigure}[b]{0.8\columnwidth}
        \includegraphics[width=\textwidth]{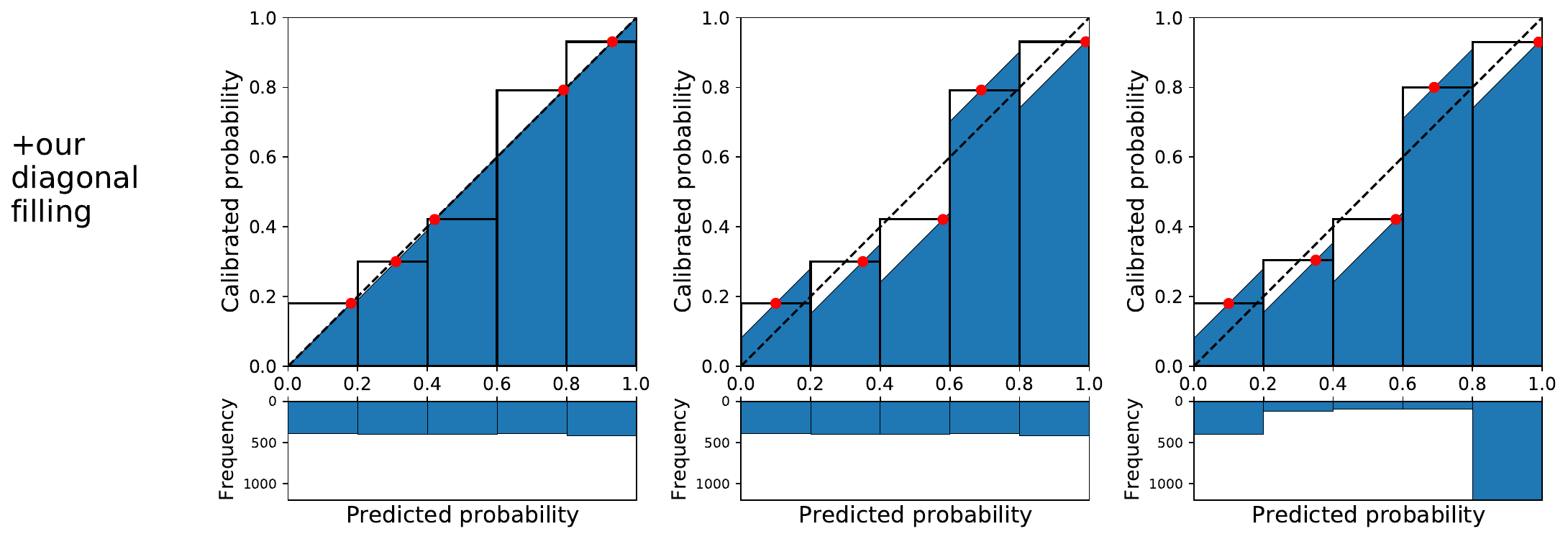}
        \label{fig:reliability_diagram_diag_red_hist}
    \end{subfigure}%
    
    \caption{Reliability diagrams of 3 classifiers (columns) created with different visualisation methods (rows); the concrete classification task is irrelevant.
    All classifiers have different ECE, but the plain reliability diagrams (top row) are identical. Differences are gradually revealed by adding bin centres indicating the average predicted probabilites (red dots, in the second row) and frequency histograms (in the third row). The last row shows our proposed reliability diagrams with diagonal filling, where ECE is better visualised because it is equal to the instance-wise average distance from the blue boundary to the main diagonal of perfect calibration.
    }
    \label{fig:flat_vs_slope}
    \end{center}
\end{figure}

The simplest classical binning-based reliability diagrams just present the bar plot showing the bins and the proportions of positives in these bins.
As we demonstrate in Figure~\ref{fig:flat_vs_slope} (top row), it can happen that 3 classifiers with very different ECE values of $0.002$, $0.090$ and $0.070$ have an identical reliability diagram.
This is because the proportions $\yb_k$ of positives in the bins $k=1,...,b$ are respectively the same for the 3 classifiers.
However, ECE is different for these classifiers due to differences in the average predictions $\pb_k$ in the bins.
A common way to address this problem is to visually indicate the average prediction within each bin~\citep{song2021classifier}. 
The second row of Figure~\ref{fig:flat_vs_slope} does so by showing bin centres $(\pb_k,\yb_k)$ with red dots.
This reveals that the first classifier is actually nearly calibrated according to this binning, because the red dots are almost at the diagonal of perfect calibration, hence very low ECE of $0.002$. 
However, the second and third classifiers still have identical visualisation, while the values of ECE are different.
This is caused by different numbers $n_k$ of instances within the bins, thus resulting in different weights for the terms in the formula Eq.(\ref{eq:ece}) for ECE.
Therefore, it is important to complement reliability diagrams with frequency histograms~\citep{song2021classifier}, as we have done in the third row of Figure~\ref{fig:flat_vs_slope}.

In the last row of Figure~\ref{fig:flat_vs_slope} we propose new reliability diagrams that we call \emph{reliability diagrams with diagonal filling}.
The original bar plot is kept there with a black line.
The blue colour does not fill the bars to the horizontal top as usual, but instead to a diagonal line with slope $1$ that crosses the top of the bar at the red bin centre $(\pb_k,\yb_k)$. 
As we know from Theorem~\ref{thm:slope1}, the top boundary of the blue filling represents the calibration map resulting from fitting the family $\cC_{(\vB,\cH,\vOne)}$.
The theorem also states that the average distance between this calibration map and the main diagonal of perfect calibration is equal to the standard ECE.
In this way, the difference between classifiers 1 and 2 becomes clearly evident from the figure.
Classifier 1 is almost perfectly calibrated because the estimated calibration map nearly matches the main diagonal, whereas classifier 2 is quite far from being calibrated.
More precisely, the area between the diagonal filling and the main diagonal is exactly equal to ECE, assuming that the bins have equal width and an equal number of instances in them.
However, if the bins have different numbers of instances (e.g. as for the 3rd classifier), then the areas between the diagonal filling and the main diagonal need to be weighted accordingly, as in the third row of the figure.

Our reliability diagrams with diagonal filling are just one example of getting reliability diagrams using fit-on-test calibration maps.
More generally, one could use any post-hoc calibration method on the test data to obtain an estimated calibration map, and then fill the area under this curve.
We call such visualisations as \emph{fit-on-test reliability diagrams}.
With any such diagram, the fit-on-test estimator of calibration error can be measured as the average distance across all instances between the reliability diagram and the main diagonal. 
Figure~\ref{fig:1_part2_pl_cv_reldiag} shows examples of fit-on-test reliability diagrams, but the calibration map families PL and PL3 used there will yet be introduced in Section~\ref{sec:PL_RD}.

\subsection{Cross-Validated Number of Bins for ECE}
\label{sec:cv_n_bins}

\begin{figure}
    \begin{center}
    \begin{subfigure}[b]{0.33\columnwidth}
        \caption{}
        \includegraphics[width=\textwidth]{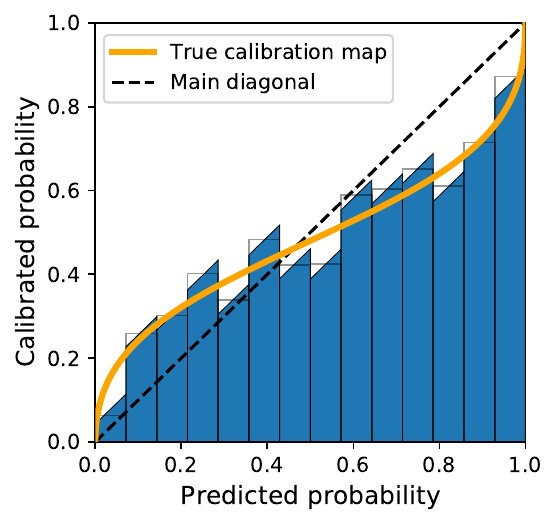}
        \label{fig:reliability_diagram_cv}
    \end{subfigure}%
     ~ %
    \begin{subfigure}[b]{0.33\columnwidth}
        \caption{}
        \includegraphics[width=\textwidth]{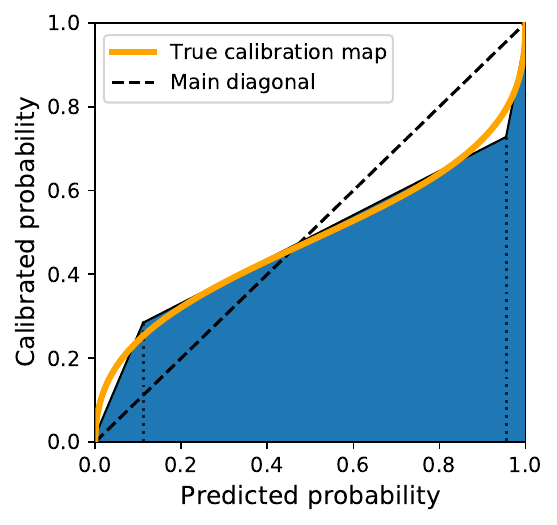}
        \label{fig:reliability_diagram_pl}
    \end{subfigure}%
     ~ %
    \begin{subfigure}[b]{0.33\columnwidth}
        \caption{}
        \includegraphics[width=\textwidth]{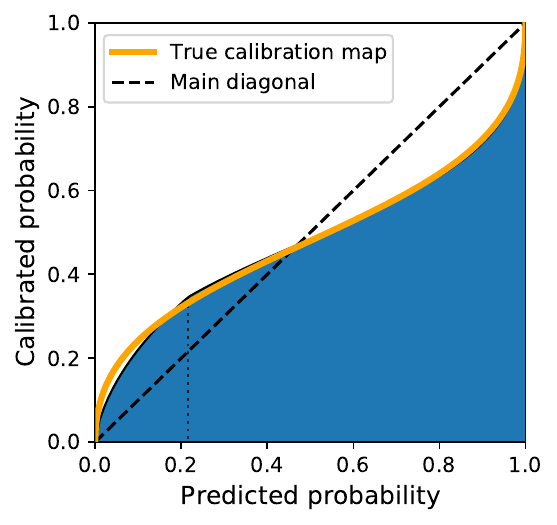}
        \label{fig:reliability_diagram_pl3}
    \end{subfigure}

    \caption{Different reliability diagrams with the number of bins or pieces optimised using cross-validation, on the same data as in Figure~\ref{fig:1_calmap_reldiag}: (a) a reliability diagram with diagonal filling using 14 bins; (b) piecewise linear reliability diagram with 3 pieces; (c) piecewise linear in logit-logit space reliability diagram with 2 pieces. 
    }
    \label{fig:1_part2_pl_cv_reldiag}
    \end{center}
\end{figure}

As discussed at the beginning of this Section~\ref{sec:fit_on_the_test}, choice of the number of bins strongly influences how well the standard binning-based $\ECE^{(\alpha)}_{\vB}$ is estimating the true calibration error.
Viewing ECE as fitting the family $\cC^{(b)}_{(\vB,\cH,\vOne)}$ on the test data by minimising the Brier score, we can see the choice of the number of bins $b$ as a hyper-parameter optimisation task.
We can now come up with novel methods for choosing the number of bins for ECE. 
For example, we can split the test set randomly into two folds: on one fold we perform fitting with different numbers of bins, and on the other fold evaluate which number of bins provides the best fit according to the Brier score. 
After that, the final reliability diagram could be drawn with this selected number of bins, and ECE calculated based on this diagram. Instead of such fixed split into two folds, any other hyperparameter optimisation technique can be used.
We propose to use cross-validation (CV), a typical hyper-parameter tuning method, to select the number of bins which provides the best fit. In the example of Figure~\ref{fig:1_calmap_reldiag}, the optimum was achieved by 14 bins as shown in Figure~\ref{fig:reliability_diagram_cv}.
Note that we are fitting $\ch(\ph_i)$ to $y_i$ for $i=1,\dots,n$ and thus CV is improving the fit between the estimated calibration map (which is the top of diagonal filling of the reliability diagram) and the binary labels.
As a result, the fit between the estimated calibration function $\ch$ and the true calibration curve $c^*_f$ also improves in expectation, as implied by Theorem~\ref{thm:calmapfitting}. 
A better fit of $\ch$ and $c^*_f$ implies a `more reliable' reliability diagram, in the sense that on average, the top of the filling is on average closer to the true calibration function.
While cross-validation is a standard tool for hyperparameter tuning, it has been missed for ECE because it has not been seen as fitting before.

In the implementation of CV, inspired by \citet{TIKKA20082604}, we prefer a lower number of bins whenever the relative difference in loss is less than 0.1 percent, further improving performance (\inserted{for details} see Appendix~\ref{subsec:cv}).

\section{Calibration Map Families PL and PL3}
\label{sec:PL_RD}

The family $\cC^{(b)}_{(\vB,\cH,\vOne)}$ of functions used by the binning-based ECE has several weaknesses: (1) it contains non-continuous functions while `jumps' are unlikely to be present in the true calibration function; (2) it only contains segments with slope $1$, making it hard to fit the true calibration function in regions with a different slope. 

Therefore, we are instead looking for a family satisfying the following criteria: (1) contains only continuous functions; (2) has flexibility to fit any curve; (3) contains the identity function; (4) has few parameters not to overfit heavily.

We first revisit the families used by the existing post-hoc calibration methods. Some methods use families with a constant fixed number of parameters, such as Platt scaling \citep{platt1999probabilistic} (2 parameters) and beta calibration \citep{pmlr-v54-kull17a} (3 parameters), see also Table~\ref{tab:fot_comparison}.
A small fixed number of parameters is clearly not sufficient to have the flexibility to fit any curve.
The non-parametric methods like isotonic calibration~\citep{isotonic} have a tendency to be overconfident and thus overfit the data \citep{allikivi2019nonparametric}.
Therefore, we are looking for methods that have a flexible number of parameters, so that a suitable number could be selected according to the dataset, for example by cross-validation.
Piecewise linear functions with slope 1 satisfy this requirement, but these in turn are not continuous.
This motivates our following proposal of using the family of continuous piecewise linear calibration maps, with unconstrained slopes.

\begin{table}
\centering
\caption{A selection of fit-on-test estimators of calibration error in binary classification, resulting from different choices of the calibration map family $\cC$ and proper loss function $l$.}
\label{tab:fot_comparison}
\begin{adjustbox}{width=1\textwidth}
\begin{tabular}{l|ccl}
\toprule
calibration map family $\cC$        & number of parameters & proper loss $l$                 & $ECE_{(\cC,l)}$ \\
\midrule
piecewise slope 1: $\cC_{(\vB,\cH,\vOne)}$      & $b$ (number of bins)                  & Brier score      & {\bf ECE (classical)} \\
logistic                                                & $2$                  & log-loss          & ECE-Platt       \\
beta calibration                                        & $3$                  & log-loss          & ECE-Beta        \\
isotonic (monotonically increasing)                     & non-parametric     & Brier or log-loss & ECE-Iso         \\
piecewise linear : $\cC_{\cB,\cH,\text{cont}}^{(b)}$   & $2b$                & Brier or log-loss & ECE-PL          \\
piecewise linear in logit-logit space                   & $2b$                 & Brier or log-loss & ECE-PL3         \\
\bottomrule
\end{tabular}
\end{adjustbox}
\end{table}

\subsection{PL - Piecewise Linear Calibration Maps}

We propose to use the subfamily $\cC_{\cB,\cH,\text{cont}}^{(b)}$ of continuous functions from the piecewise linear function family $\cC_{\cB,\cH,\cA}^{(b)}$.
The family $\cC_{\cB,\cH,\text{cont}}^{(b)}$ is only parametrised by the bin boundaries $\vB$ and by the values $\vH$ of the function at the boundaries, whereas the slopes can be calculated from $\vB$ and $\vH$ with $A_k=\frac{H_{k+1}-H_{k}}{B_{k+1}-B_{k}}$ to ensure that the line at the right end of bin $k$ coincides with the line at the left end of bin $k+1$.
Note that the bin boundaries $\vB$ are now also parameters to be fitted, together with the values $\vH$.
As $B_1=0$ and $B_{b+1}=1+\epsilon$ are fixed, we are fitting $2b$ parameters: $b-1$ bin boundaries and $b+1$ values in $\vH$.
The number of bins $b$ is optimised through cross-validation, similarly to Section~\ref{sec:cv_n_bins}.

We call the corresponding fit-$(\cC_{\cB,\cH,\text{cont}}^{(b)},l)$-on-test method as PL: \emph{the piecewise linear method for evaluating calibration}.
In particular, we can now draw new kind of \emph{piecewise linear reliability diagrams} (Figure~\ref{fig:reliability_diagram_pl}) which provide a better fit to the true calibration function than the binning-based methods, as demonstrated in our experiments (Section~\ref{sec:experiments}). For a visual comparison, check \modified{Figure~\ref{fig:reliability_diagram_cv} and Figure~\ref{fig:reliability_diagram_pl}}, as these 
figures have been made with the same data. More comparative examples can be found in Appendix~\ref{subsec:comp_of_rel_diag}.
Therefore, \inserted{the piecewise linear reliability diagram} can be used similarly to the \inserted{classical} ECE reliability diagram to check \inserted{visually} how well the model is calibrated.
We can then also use this estimated calibration map for fit-on-test estimation of calibration error, getting the measure which we call the \emph{piecewise linear} ECE or in short ECE-PL. 
According to the fit-on-test estimation method, ECE-PL measures the instance-wise average distance from the piecewise linear function to the main diagonal: $\ECE_{\PL}=\frac{1}{n}\sum_{i=1}^n \vert\ch(\ph_i)-\ph_i\vert^{\alpha}$ where 
$\ch=\argmin_{c\in\cC_{\cB,\cH,\text{cont}}^{(b)}} \frac{1}{n}\sum_{i=1}^n (c(\ph_i)-y_i)^2$.

\paragraph{Implementation Details}
%
Although the continuous piecewise linear functions are mathematically very well known, we found only one existing public implementation~\citep{pwlf}, based on least squares fitting with differential evolution. We included this in our experiments with the name $PL_{DE}$  \inserted{(results in Appendix~\ref{sec:results_supp})}. However, we also created ourselves a neural network based implemention \inserted{depicted in Figure~\ref{fig:pl_nn_architecture}}, 
which allowed us to add cross-entropy fitting. 
Full details about the architecture are given in \inserted{Appendix}~\ref{sec:implementation_supp}, but here is a short overview. 

We have a single input ($\ph$) and a single output ($\ch$) connected through two layers: the binning layer and the interpolation layer.
The binning layer has $b+1$ gating units corresponding to the bin boundaries, each outputting whether $\ph$ is to the left or to the right of the boundary ($L_i=\hat{p}-B_i$ and $R_i=B_i-\hat{p}$). The binning layer is parametrised by $b$ real values ($\mathbf{Z}=z_1, z_2, \dots, z_b$) which are passed through the softmax \modified{($\sigma$) to obtain the widths of the bins and through cumulative sum ($B_i = \sum_{k=1}^{i} \sigma_k(\mathbf{Z})$) to obtain bin boundaries.} These parameters are initialised such that all bins contain the same number of training instances.
The interpolation layer has $b+1$ parameters which are each passed through the logistic function \inserted{($\phi$)} to obtain the calibration map values $H_1,\dots,H_{b+1}$ at the bin boundaries. These are initialised such that the represented calibration map is the identity function. The $b$ units correspond to the bins and the bin to which $\ph$ belongs produces the linearly interpolated output. 
Based on these values, the piecewise linear function value is calculated as a sum of two neighbouring nodes ($H_k$, $H_{k+1}$) as follows: $$\ch = \sum_{k=1}^{b} g(L_{k+1}, R_k)\left(H_k\frac{L_{k+1}}{L_{k+1} + R_k} + H_{k+1}\frac{R_k}{L_{k+1} + R_k}\right)$$
where the gating function $g$ is defined as $g(L_{k+1}, R_k) = I[(L_{k+1}>0)\&(R_k>0)]$.

\inserted{
    \begin{figure}
     \centering
    \includegraphics[width=0.8\textwidth]{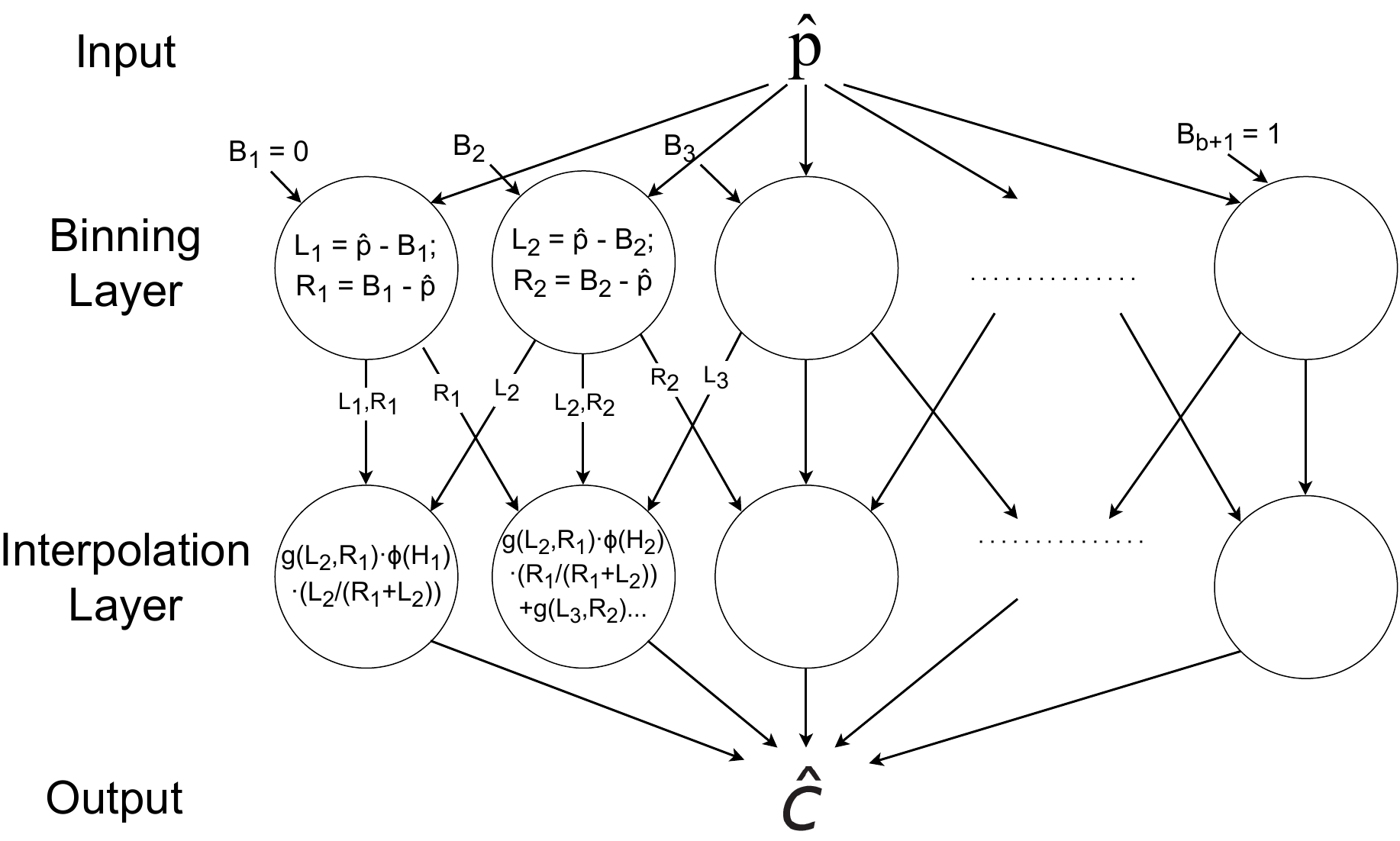}
    \caption{Architecture of the piecewise linear function implementation as a neural network. \inserted{The bin boundaries $B_i$ are parametrised by logits ($\mathbf{Z}=z_1, z_2, \dots, z_b$). These are passed through the softmax with cumulative sum ($B_i = \sum_{k=1}^{i} \sigma_k(\mathbf{Z})$). The $H_i$ values are activated by logistic function $\phi$. The $g$ stands for a gate, which is an indicator function, that outputs $1$ if input is in the bin or $0$ otherwise.}}
    \label{fig:pl_nn_architecture}
    \end{figure}
}

We use 10-fold-cross-validation to select the number of segments in the piecewise linear function to best approximate the true calibration map, similar to Section~\ref{sec:cv_n_bins}. 
The same way as in hyperparameter optimization for Dirichlet calibration \citep{Kull2019BeyondTS}, the predictions on test data are obtained as an average output from all the 10 models with the chosen number of segments but trained from different folds, i.e. we are not refitting a single model on all 10 folds. 

\subsection{PL3 - Piecewise Linear in Logit-Logit Space}

The piecewise linear method can be used for calibration evaluation universally for any kinds of models, but it also makes sense to seek for dedicated families for special cases, such as for neural networks. 
Next we propose the family PL3 specifically for evaluating calibration in neural networks, taking inspiration from temperature scaling \citep{Guo2017} and beta calibration \citep{pmlr-v54-kull17a}.

\begin{figure}
    \begin{center}
    \begin{subfigure}[b]{0.91\columnwidth}
        \caption{}
        \includegraphics[width=\textwidth]{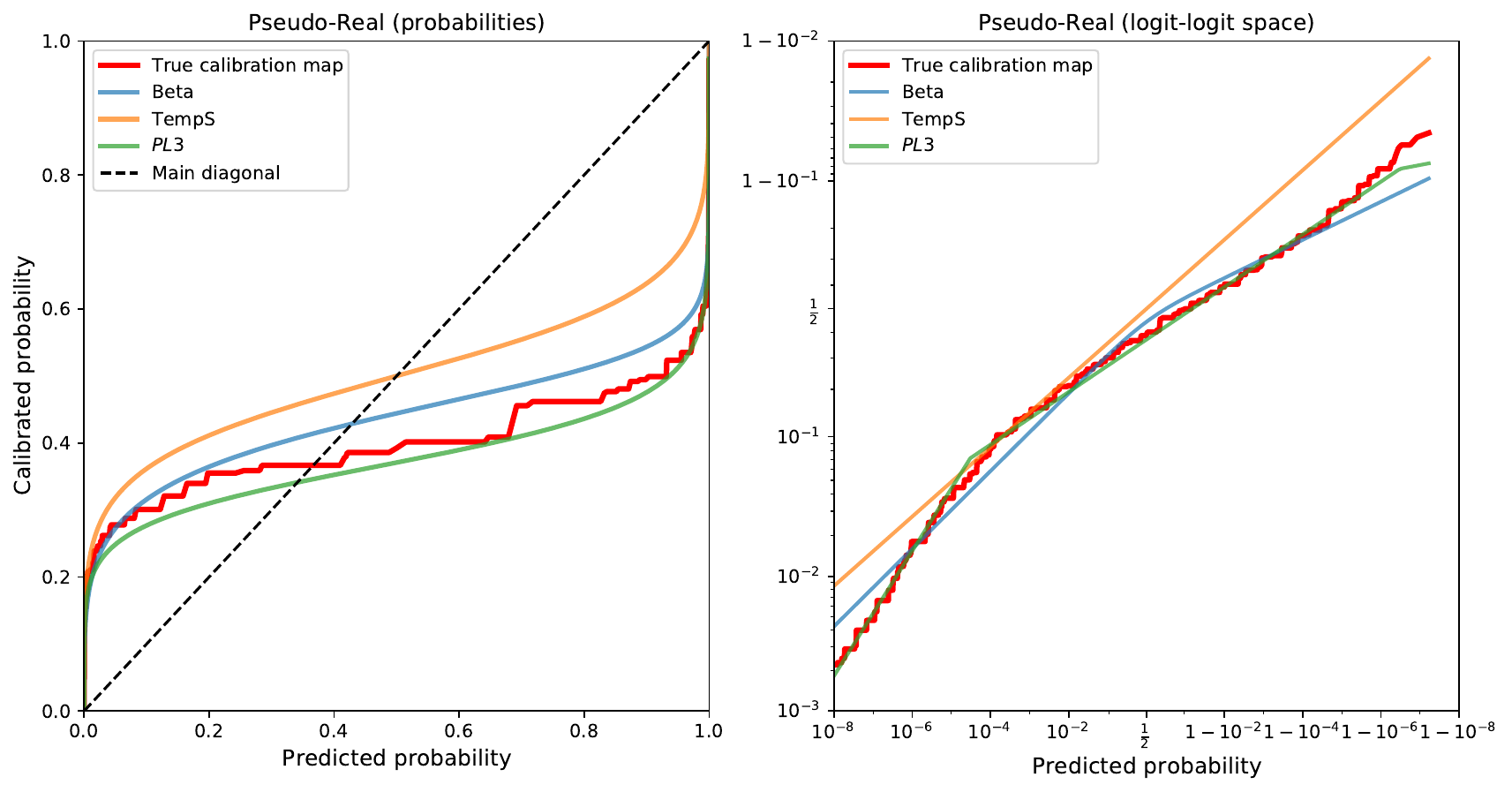}
        \label{fig:pl3_motivation_1}
    \end{subfigure}%

    \begin{subfigure}[b]{0.91\columnwidth}
        \caption{}
        \includegraphics[width=\textwidth]{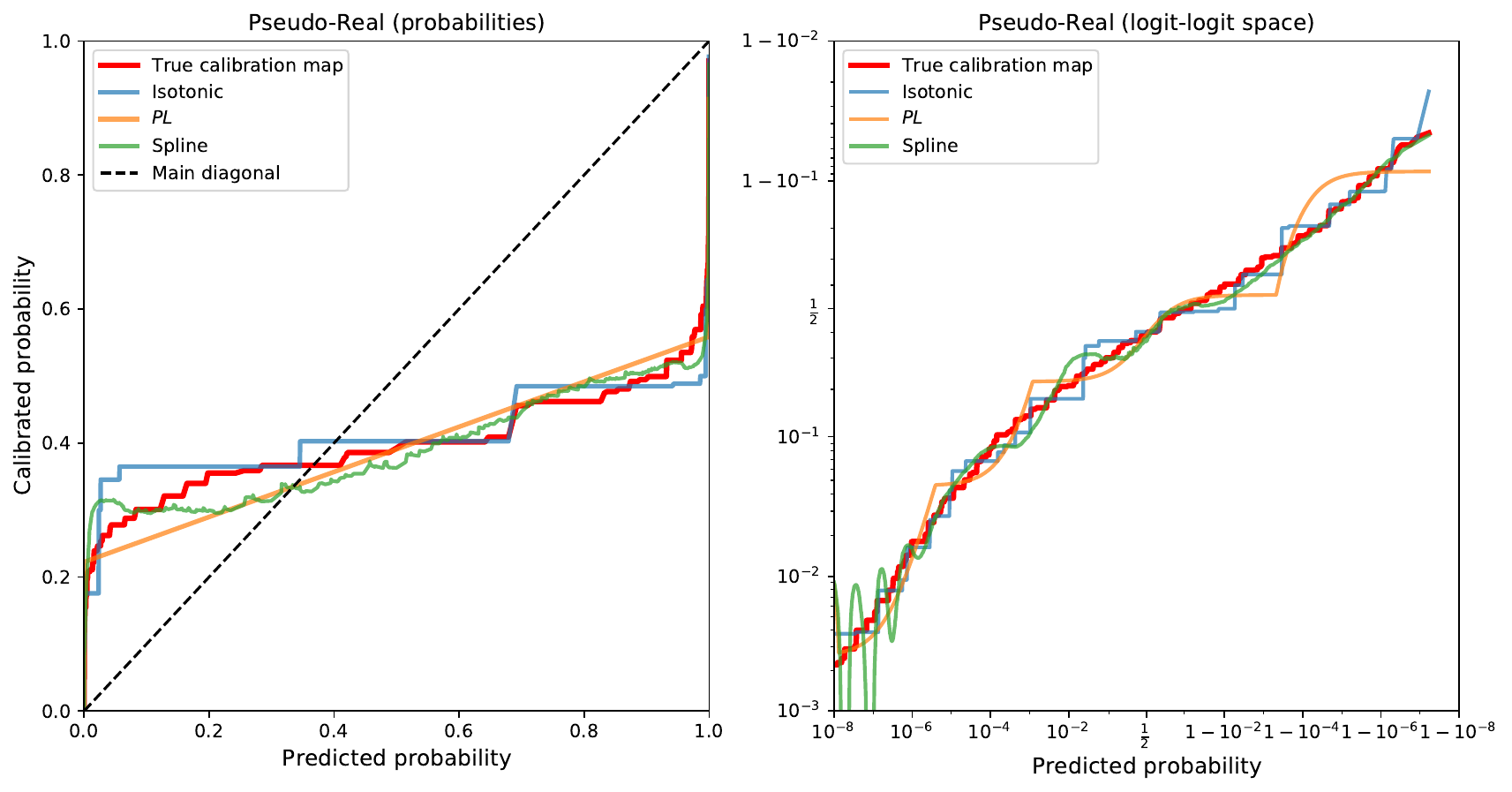}
        \label{fig:pl3_motivation_2}
    \end{subfigure}

    \caption{Motivation for PL3. Comparison of results in the probability space (left column) and in the logit-logit space (right column). \inserted{Methods are split into two groups for clarity. In essence, subfigures (a) and (b) are the same, only display different functions. Model ResNet110~\citep{resnet} on "cats vs rest" task of CIFAR-5m.} 
    }
    \label{fig:pl3_motivation}
    \end{center}
\end{figure}

Temperature scaling is fitting a family of functions $\ch(\ph)=\sigma(\mathbf{z}/t)$ with a single temperature parameter $t$, where the softmax $\sigma$ is applied on logits $\mathbf{z}$ that the uncalibrated model would have directly converted into probabilities with $\ph=\sigma(\mathbf{z})$ \citep{Guo2017}. 
In the binary classification case with a single output, $\sigma(z)=1/(1+e^{-z})$ is the logistic function, the inverse function of the logit $\sigma^{-1}(p)=\ln(p/(1-p))$.
Importantly, if plotted in the logit-logit scale, binary temperature scaling fits a straight line (Figure~\ref{fig:pl3_motivation}), since $\sigma^{-1}(\ch(\ph))=\sigma^{-1}(\sigma(\mathbf{z}/t)=z/t=1/t\cdot \sigma^{-1}(\sigma(\mathbf{z}))=\sigma^{-1}(\ph)$. 
Further, Figure~\ref{fig:pl3_motivation} compares various methods in the probability space and logit-logit space, giving extra information on how well the methods fit the calibration map in the low and high-probability regions.

Interestingly, another post-hoc calibration method known as beta calibration \citep{pmlr-v54-kull17a} fits calibration maps which in the logit-logit space are approximately piecewise linear with two segments, as shown \inserted{in Figure~\ref{fig:pl3_motivation} (top right subfigure, blue line)}. The proof for this fact is given in Appendix~\ref{subsec:calmap_in_ll_scale}. This motivates our calibration map family PL3 of Piecewise Linear functions in the Logit-Logit space (PLLL=PL3), \inserted{which corresponds to temperature scaling when using one piece, approximates beta calibration when using two pieces, and can take more complicated shapes when using more linear pieces in the logit-logit space.}
Figure~\ref{fig:reliability_diagram_pl3} shows an example of a piecewise linear in the logit-logit space reliability diagram.
\inserted{Further details are provided in Appendix~\ref{subsec:calmap_in_ll_scale}.}

\paragraph{Implementation Details}  
To implement PL3, we use the same neural architecture as for PL, with the following modifications: (1) the expected input is in the logit space; (2) the bin boundaries are converted into the logit space; (3) the logistic is removed from the parameters feeding the interpolation layer and instead the logistic is applied on the final output. Further details are given in \inserted{Appendix}~\ref{sec:implementation_supp}.

\section{Assessment of Calibrators and Evaluators}
 \label{sec:eval_methods_of_cal}

Before proceeding to the experiments, let us discuss the methods for assessing the quality of calibrators and evaluators of calibration.

\subsection{Assessment of Post-hoc Calibrators}
 \label{subsec:eval_methods_of_posthoccal}
Post-hoc calibrators should be evaluated based on the effectiveness of calibration, but this effectiveness could be interpreted in two ways.
\modified{Firstly}, we could evaluate how well-calibrated the outputs of the calibrator are by calculating the calibration error after calibration (CEAC):
\begin{align*}
    CEAC&=\CE(\ch\circ f)=\E[d(\Ch,c^*_{\ch\circ f}(\Ch))] 
\end{align*}
\modified{Secondly},
we could evaluate how well \inserted{the calibrator} approximates the true calibration map by calculating the calibration map estimation error (CMEE):
\begin{align*}
    CMEE&=\E[d(\ch(\Ph),c^*_f(\Ph))]
\end{align*}
where $d$ is any Bregman divergence and $\Ch=\ch(\Ph)=(\ch\circ f)(X)=\ch(f(X))$.
We prove that low calibration map estimation error is a stronger requirement than low calibration error after calibration, because $CMEE$ is an upper bound for $CEAC$:
\begin{theorem}
\label{thm:cmee}
\begin{equation*}
    CMEE=CEAC+\E[d(c^*_{\ch\circ f}(\Ch),c^*_f(\Ph))].
\end{equation*}
\end{theorem}
Intuitively, the difference between $CMEE$ and $CEAC$ is due to ties introduced by $\ch$ where $\ch(\ph)=\ch(\ph')$ while $c^*_f(\ph)\neq c^*_f(\ph')$ for some $\ph\neq\ph'$.
As low $CMEE$ is a stronger requirement, we prefer this measure in the experiments. We use the notation $d(\ch,c^*)$ as a more memorable synonym for $CMEE$.

\subsection{Assessment of Calibration Evaluators}
\label{subsec:eval_of_cal_evaluators}

Calibration evaluators that follow the fit-on-test paradigm, estimate $\ch$ on the test data.
The resulting $\ch$ can be viewed as a reliability diagram, or used to estimate the calibration error with $\ECE^{(\alpha)}_{\text{fit-}(\cC,l)\text{-on-test}}$. 

We see 3 main application scenarios:
\begin{enumerate}
\item
the reliability diagram is needed if the reliability of individual predictions has to be known separately; for example, this is needed if there is a downstream decision-making process performed by a human or AI;
\item 
the estimated calibration error is needed if the general level of trust needs to be known, not specifically for each output;
\item
and the ranking of the estimated calibration errors is needed if performing selection of the best calibrated model.
\end{enumerate}

\modified{In the experiments, we evaluate each calibration evaluation method $\mathcal{M}$ against these three objectives, measuring: (1) the quality of reliability diagrams with $d(\ch_\mathcal{M},c^*)$; (2) the quality of calibration error estimates with $\vert ECE_\mathcal{M}-CE\vert$; and (3) the quality of ranking performance using Spearman's correlation of the ranking produced by the calibration evaluator against the ranking of true calibration errors, which we refer to as $rankcorr(ECE_\mathcal{M},CE)$.}
All these methods assume good estimates of the true calibration map, which can be obtained either on synthetic data or if there is access to magnitudes of more data than the calibration evaluator used.

\inserted{
     Note that in the experiments we mostly use the absolute difference $d(\ch,c^*)=\vert\ch-c^*\vert$ as the distance measure between the estimated and true calibration maps, following the tradition of how ECE is calculated from the reliability diagrams. As the absolute difference is not a Bregman divergence, the Appendix also considers the squared difference $d(\ch,c^*)=\vert\ch-c^*\vert^2$ which is a Bregman divergence.
     }
\inserted{
    Importantly, a method $\mathcal{M}$ that produces the best calibration map estimate with the lowest $\vert\ch_\mathcal{M}-c^*\vert$ might not be the same as method $\mathcal{M}'$ that produces the best calibration error estimate $\vert ECE_{\mathcal{M}'}-CE\vert$. For example, this occurs when comparing the methods in Figures~\ref{fig:reliability_diagram_cv} and \inserted{\ref{fig:reliability_diagram_pl}}. In Figure~\ref{fig:reliability_diagram_pl}, the piecewise linear method (PL) achieves $\vert ECE_{PL}-CE\vert=0.0045$ and $\vert\ch_{PL}-c^*\vert=0.0204$, while in Figure~\ref{fig:reliability_diagram_cv}, equal-width binning with cross-validated number of bins ($EWCV$) achieves $\vert ECE_{EWCV}-CE\vert=0.0024$ and $\vert\ch_{EWCV}-c^*\vert=0.0398$. Thus, even though PL has in this example better $\vert\ch-c^*\vert$, it still loses to $EWCV$ with respect to $\vert ECE-CE\vert$. Intuitively, $EWCV$ has bigger errors in the reliability diagram, but during the calculation of ECE these errors happen to cancel out more than in the case of $PL$.
}

\section{Experiments and Results}
\label{sec:experiments}

\subsection{Pseudo-real Experiments and results}
\inserted{
    The goal of our experimental studies is to (1) evaluate our proposed PL and PL3 calibration map families as calibration methods; and (2) find the best calibration map family for calibration evaluation (thanks to the fit-on-test paradigm).
}

\inserted{
     A key problem for research in calibration evaluation methods and post-hoc calibration methods is that  proper evaluation requires access to the true calibration map. This, however, is unknown. One workaround would be to use synthetically created data, where the true calibration map is known. However, with synthetic data the shape of the true calibration map might not be realistic.
}
Our solution to this is to use the pseudo-real dataset CIFAR-5m~\citep{cif5m} with 5 million synthetic images, created such that the models trained on CIFAR-10~\citep{cif} have very similar performance on CIFAR-5m, and vice versa.
Thus, it is likely that the true calibration maps are also very realistic.
\inserted{
    Thanks to the vast size of the CIFAR-5m dataset we can estimate the true calibration map very precisely. In our experiments, we used isotonic calibration on 1 million hold-out datapoints to estimate the true calibration map (other options than isotonic were considered in Appendix~\ref{sec:results_supp}, minor differences in the results).
}
The main part of the experiments concentrates on CIFAR-5m, while \inserted{Appendix}~\ref{sec:results_supp} shows more results on synthetic and real datasets, where the results are comparable to CIFAR-5m with minor differences, supporting the overall conclusions. 
On CIFAR-5m, we concentrated on three 1-vs-rest calibration tasks (car, cat, dog) and confidence calibration.
\inserted{
Only three 1-vs-rest tasks were used due to computational limitations.
}

%
%
%
ResNet110~\citep{resnet}, WideNet32~\citep{wide} and DenseNet40~\citep{dense} models were trained on 45k datapoints from CIFAR-5m, additional 5k datapoints were used to calibrate the outputs of models with multi-class calibration methods: \modified{temperature scaling (TempS), vector scaling (VecS), and matrix scaling with off-diagonal and intercept regularisation (MSODIR); Dirichlet calibration with ODIR regularisation (dirODIR), Spline with natural method (Spline); Order-invariant version of intra-order preserving functions (IOP); binary calibration methods: Platt, isotonic, beta calibration (beta), scaling-binning (ScaleBin), ECE-based binning ($ES_{sweep}$) methods with equal-size binning (ES), piecewise linear methods (PL and PL3).}
\inserted{
    For PL we used our neural network model trained with cross-entropy loss (the results with optimising for Brier score and with the alternative optimisation method based on differential evolution are included in Appendix~\ref{sec:results_supp}).
}


\begin{table}
\caption{Comparison of post-hoc calibrators with respect to $\vert\ch_\mathcal{M}-c^*\vert$ \inserted{$(\times10^{3})$} in estimating the true calibration map of DenseNet40, ResNet110 and WideNet32 classifiers on CIFAR-5m. 
Calibrators are trained on 5000 instances, results are averaged across the 3 classifiers. 
}
    \label{table:cal-method-table}
    \centering
    \begin{adjustbox}{width=\textwidth}

\begin{tabular}{l|llllllllll|lll}
\toprule
 &       Platt &         beta &    isotonic &   ScaleBin &        TempS &         VecS &      dirODIR &       MSODIR &      Spline &          IOP & $ES_{sweep}$ &               $PL3$ &        $PL$ \\
\midrule
cars vs rest &   $3.6_{2}$ &    $4.3_{7}$ &   $4.0_{5}$ &  $4.1_{6}$ &   $7.3_{12}$ &    $3.8_{4}$ &    $3.8_{3}$ &    $4.3_{8}$ &  $6.7_{10}$ &   $8.2_{13}$ &   $7.1_{11}$ &  $\mathbf{2.9_{1}}$ &   $4.9_{9}$ \\
cats vs rest &   $6.8_{2}$ &  $15.0_{12}$ &   $9.0_{4}$ &  $9.5_{5}$ &  $14.0_{11}$ &   $10.4_{6}$ &   $13.3_{9}$ &   $12.9_{8}$ &   $8.8_{3}$ &  $13.5_{10}$ &  $19.3_{13}$ &  $\mathbf{5.3_{1}}$ &  $12.3_{7}$ \\
dogs vs rest &   $4.5_{2}$ &  $14.4_{12}$ &   $6.8_{4}$ &  $6.7_{3}$ &  $11.7_{11}$ &    $8.6_{5}$ &  $11.6_{10}$ &   $11.5_{9}$ &   $8.7_{6}$ &   $11.0_{8}$ &  $25.5_{13}$ &  $\mathbf{3.5_{1}}$ &  $10.6_{7}$ \\
confidence   &  $13.1_{4}$ &   $17.4_{7}$ &  $11.6_{3}$ &  $8.7_{2}$ &   $22.7_{8}$ &  $34.2_{11}$ &  $38.1_{13}$ &  $37.5_{12}$ &  $15.4_{6}$ &  $24.8_{10}$ &   $23.9_{9}$ &  $\mathbf{8.5_{1}}$ &  $15.4_{5}$ \\
\bottomrule
\end{tabular}
\end{adjustbox}
\end{table}

\paragraph{Results} We show the results in absolute differences (i.e. $\alpha=1$) as in most earlier works (
\inserted{Appendix}~\ref{sec:results_supp} has the quadratic also, i.e. $\alpha=2$). 
\inserted{
    Table~\ref{table:cal-method-table} assesses PL and PL3 as post-hoc calibration methods. 
    The calibration map family corresponding to ECE fit with the sweeping method \citep{sweep} is also included as $ES_{sweep}$. 
    The calibration methods are evaluated by how well they approximate the true calibration map: $\vert\ch_\mathcal{M}-c^*\vert=\sum_{i=1}^n\vert\ch_\mathcal{M}(\ph_i)-c^*(\ph_i)\vert$ is measured on unseen 1 million data points against the ground truth, where $\ph_i$ are the outputs of the classifier, $\ch_\mathcal{M}(\ph_i)$ are the post-hoc calibrated predictions, and $c^*(\ph_i)$ is the `true' calibration map.
}
PL3 is the best method in all cases, showing the usefulness of the logit-logit space when calibrating neural models which are far from \inserted{being} calibrated. 
Note that the errors of PL3 are in all cases more than 20\% smaller than the errors of the second-performing method Platt.
\inserted{Appendix}~\ref{sec:results_supp} includes more variations of these methods, the KDE method and results for each architecture separately (minor differences).

\begin{table}
\caption{Comparison of calibration evaluators in estimating the reliability diagram with $\vert\ch_\mathcal{M}-c^*\vert$ \inserted{$(\times10^{3})$} on CIFAR-5m. \inserted{In contrast to Table~\ref{table:cal-method-table}, the evaluators are compared on predictions that have been previously post-hoc calibrated.} 
}
\label{table:chat_dist_c_combo}
\centering
\begin{adjustbox}{width=\textwidth}
\begin{tabular}{ll|llllllllll}
\toprule
     &  &    $ES_{15}$ & $ES_{sweep}$ &    $ES_{CV}$ &   $PL3$ &       $PL$ &   Platt &         beta &     isotonic & Spline & IOP  \\
{} & {} &              &              &              &              &                           &              &              &   & &           \\

\midrule
     & All &    $9.47_{9}$ &   $8.62_{7}$ &   $8.52_{6}$ &   $7.27_{2}$ &  $\mathbf{6.27_{1}}$ &   $8.39_{5}$ &    $7.6_{3}$ &     $9.2_{8}$ &    $9.55_{10}$ &   $7.81_{4}$ \\
     \hline
Initial model & ResNet110 &    $9.43_{8}$ &   $9.04_{6}$ &   $9.22_{7}$ &   $7.33_{2}$ &   $\mathbf{6.7_{1}}$ &    $8.9_{5}$ &   $7.97_{3}$ &    $9.66_{9}$ &   $10.15_{10}$ &   $8.64_{4}$ \\
     & DenseNet40 &   $9.98_{10}$ &   $8.59_{7}$ &   $8.37_{6}$ &   $7.08_{3}$ &  $\mathbf{5.98_{1}}$ &   $7.93_{5}$ &   $7.35_{4}$ &    $9.15_{8}$ &     $9.42_{9}$ &   $6.92_{2}$ \\
     & WideNet32 &     $9.0_{9}$ &   $8.22_{6}$ &   $7.97_{5}$ &   $7.39_{2}$ &  $\mathbf{6.14_{1}}$ &   $8.33_{7}$ &   $7.49_{3}$ &     $8.8_{8}$ &    $9.08_{10}$ &   $7.86_{4}$ \\
     \hline
Data size & 1000 &  $13.34_{10}$ &  $11.44_{6}$ &  $11.49_{7}$ &  $10.23_{4}$ &  $\mathbf{8.02_{1}}$ &  $10.88_{5}$ &   $9.63_{3}$ &   $12.99_{9}$ &    $12.08_{8}$ &   $8.17_{2}$ \\
     & 3000 &    $8.46_{8}$ &   $7.64_{6}$ &   $8.19_{7}$ &   $6.64_{2}$ &  $\mathbf{6.22_{1}}$ &   $7.44_{5}$ &    $6.8_{3}$ &    $8.65_{9}$ &    $9.24_{10}$ &   $7.43_{4}$ \\
     & 10000 &    $6.61_{6}$ &   $6.77_{7}$ &   $5.88_{3}$ &   $4.93_{2}$ &  $\mathbf{4.59_{1}}$ &   $6.85_{8}$ &   $6.38_{5}$ &    $5.97_{4}$ &     $7.34_{9}$ &  $7.81_{10}$ \\
     \hline
Task & cars vs rest &    $4.07_{7}$ &   $4.61_{9}$ &   $3.82_{4}$ &   $3.58_{3}$ &  $\mathbf{2.74_{1}}$ &   $4.21_{8}$ &   $3.13_{2}$ &     $3.9_{6}$ &    $4.84_{10}$ &   $3.82_{5}$ \\
     & cats vs rest &  $10.45_{10}$ &  $10.15_{9}$ &   $9.24_{5}$ &   $7.44_{2}$ &  $\mathbf{6.64_{1}}$ &   $9.82_{7}$ &   $9.02_{4}$ &     $9.6_{6}$ &    $10.11_{8}$ &   $8.61_{3}$ \\
     & dogs vs rest &    $9.17_{9}$ &   $8.75_{7}$ &   $8.67_{6}$ &   $7.33_{3}$ &   $\mathbf{5.8_{1}}$ &   $7.81_{5}$ &   $7.41_{4}$ &    $8.98_{8}$ &     $9.3_{10}$ &   $6.63_{2}$ \\
     & confidence &   $14.19_{9}$ &  $10.96_{4}$ &  $12.35_{7}$ &  $10.72_{2}$ &  $\mathbf{9.91_{1}}$ &  $11.71_{5}$ &  $10.84_{3}$ &  $14.34_{10}$ &    $13.95_{8}$ &  $12.16_{6}$ \\
\bottomrule

\end{tabular}
\end{adjustbox}
\end{table}

Next we compare the calibration evaluators against each of the 3 objectives listed in 
\inserted{Section~\ref{subsec:eval_of_cal_evaluators}}
. We perform the comparison in tasks where the models are already quite close to being calibrated, because this is typical when evaluators are used for finding out which post-hoc calibrator is performing best. 
We compare evaluators in the task of evaluating 6 post-hoc calibrators: we \inserted{measure} 
how precisely the evaluators $\mathcal{M}$ estimate the reliability diagrams (Table~\ref{table:chat_dist_c_combo} showing $\vert \ch_\mathcal{M}-c^*\vert$), the total true calibration errors (Table~\ref{table:cifar5m-ECE-ranking}   
showing $\vert ECE_\mathcal{M}-CE\vert$), and how well the estimated ranking of 6 calibrators agrees with the true ranking based on true calibration errors (Table~\ref{table:cifar5m-ECE-ranking} showing $rankcorrel(ECE_\mathcal{M},CE)$).
The 6 calibrators were chosen as the best methods from Table~\ref{table:cal-method-table}: beta, vector scaling, Platt, PL3, scaling-binning, isotonic. 
The evaluators are compared on different test set sizes 1k, 3k, 10k with 5 different random seeds for each size.

\begin{table}
\caption{Comparison of calibration evaluators in estimating the true calibration error \inserted{with} 
$\vert ECE_\mathcal{M}-CE\vert$ \inserted{$(\times10^{3})$}
and the ranking of 6 calibrators \inserted{with} 
$rankcorrel(ECE)$
on CIFAR-5m.
%
}
    \label{table:cifar5m-ECE-ranking}
    \centering
    \begin{adjustbox}{width=\textwidth}
\begin{tabular}{ll|llllllllll}
\toprule
           &  &            $ES_{15}$ & $ES_{sweep}$ &    $ES_{CV}$ &                $PL3$ &    $PL$ &      Platt &         beta &              isotonic & Spline & IOP \\
Metric & Data &                      &              &              &                      &              &              &              &              &         &              \\

\midrule
$\vert ECE_\mathcal{M}-CE\vert$ & one-vs-rest &  $\mathbf{2.3_{1}}$ &   $3.24_{5}$ &   $3.66_{8}$ &   $2.59_{2}$ &     $2.87_{3}$ &   $3.63_{7}$ &            $2.9_{4}$ &             $3.8_{9}$ &    $4.16_{10}$ &     $3.48_{6}$ \\
         & confidence &          $4.54_{2}$ &   $5.14_{4}$ &   $6.37_{7}$ &   $4.73_{3}$ &     $5.24_{5}$ &   $5.83_{6}$ &  $\mathbf{4.39_{1}}$ &             $7.2_{9}$ &    $7.39_{10}$ &     $6.96_{8}$ \\
         \hline
$rankcorrel(ECE)$ & one-vs-rest &         $0.457_{3}$ &  $0.195_{7}$ &  $0.443_{4}$ &   $0.36_{6}$ &    $0.513_{2}$ &   $0.01_{9}$ &          $0.128_{8}$ &  $\mathbf{0.552_{1}}$ &    $0.398_{5}$ &  $-0.043_{10}$ \\
         & confidence &         $0.624_{2}$ &  $0.434_{6}$ &  $0.563_{3}$ &  $0.406_{7}$ &     $0.51_{5}$ &  $0.018_{9}$ &          $0.511_{4}$ &  $\mathbf{0.657_{1}}$ &    $0.392_{8}$ &  $-0.095_{10}$ \\         
\bottomrule

\end{tabular}
\end{adjustbox}
\end{table}


The first objective is to assess the reliability diagrams using \modified{$\vert\ch_\mathcal{M}-c^*\vert=\sum_{i=1}^n\vert\ch_\mathcal{M}(\ph_i)-c^*(\ph_i)\vert$} \inserted{measured on 1 million unseen data points},  
\inserted{
    where $\ph_i$ are now the already post-hoc calibrated outputs of the classifier (calibrated with the 6 best methods from Table~\ref{table:cal-method-table} trained on 5k data points), $\ch_\mathcal{M}(\ph_i)$ are the results of fit-on-test calibration applied on top of the post-hoc calibrated predictions (trained on another separate data set of size either 1k, 3k, or 10k), and $c^*(\ph_i)$ is the `true' calibration map of the post-hoc calibrated predictions.
}



The results in Table~\ref{table:chat_dist_c_combo} show that PL is the best on average, as well as after disaggregating according to the classifier's architecture, test set size, or the task. PL3 is mostly second and the evaluator using the beta calibration map family is mostly third. \inserted{The order invariant version of IOP shows also promising results.}
The performance of beta calibration varies with the size of the test dataset. This is expected, because this method has only 3 parameters which is good for small test sets but worse for bigger test sets. \inserted{Similarly, the data set size affects $IOP$ too, again because of the small number of parameters.} 
The methods based on equal-size binning are performing worse, with $ES_{CV}$ ranking the highest on average, closely followed by $ES_{sweep}$, and the classical $ES_{15}$ with 15 bins lagging behind.
\inserted{Further, the Spline method is among one of the worst performing methods.}
\inserted{
    From Table~\ref{table:cal-method-table} and Table~\ref{table:chat_dist_c_combo} we can conclude that when predictions are far from being calibrated then PL3 is best for approximating the true calibration map, and PL is best when predictions are nearly calibrated.
}
Appendix~\ref{sec:results_supp} discusses the results showing different aggregations, and reports the optimal numbers of bins for ES and PL methods.

The second objective is to estimate the numeric value of the total true calibration error, and here the rows $\vert ECE_\mathcal{M}-CE\vert$ of Table~\ref{table:cifar5m-ECE-ranking} show the benefits of $ES_{15}$, beta calibration and PL3 (with some differences across tasks). 
This demonstrates that while the tilted-top reliability diagrams of $ES_{15}$ are not precise, their debiased average distance from the diagonal closely agrees with the average distance of the true reliabilities (true calibration map) from the diagonal. 
While PL and PL3 perform reasonably well, there is a big potential for further improvements, because debiasing remains as future work for these methods.
The ranking of 6 calibrators is best done by the isotonic fit-on-test evaluator, achieving over 55\% correlation to the true ranking for one-vs-rest tasks and over 65\% for the confidence task. 

\section{Discussion}
\label{sec:discussion}

\insertnew{As we explained in our work, evaluating the calibration error using the fit-on-test approach is done by estimating the calibration map by fitting a calibration map family on test data, and then using the plugin-estimator of calibration error. 
The most popular method for the calibration error evaluation is a binning ECE, which we proved to also be fit-on-test. The name fit-on-test might sound controversial but has been chosen deliberately, to highlight the essence of issues in current methods for assessing calibration.
The main concern with fitting on the test data is getting a good fit, like for any other fitting task. Thus, one needs a suitable family of functions and a suitable fitting procedure so the data is not overfitted or underfitted. A poor fit leads to poor performance of the method and unreliable results.
However, since the true calibration map and the true calibration error are not available, we do not even know in practice whether there is overfitting or underfitting during fit-on-test evaluation. One way to tackle the problem is to use synthetic or pseudo-real data that would look very similar to real data. This way, we would have an estimate of how well each fit-on-test method could work on real data.
We tested experimentally some deep neural network architectures on different datasets and saw some performance fluctuations between methods across different subtasks. 
However, there are bigger differences across different evaluation tasks. 
Even though the piecewise linear method performs very well in calibration map prediction, it is still not better than $ES_{15}$ when predicting the calibration error or ranking. This might be because of the bias introduced, and thus, there might be a need for debiasing.
Note that we did not dive into problems of data shift and out-of-distribution prediction, which are certainly affecting model uncertainty calibration as well in practice.}

\section{Conclusion and Future Work} %
\label{sec:conclusion}
We suggest to view \inserted{evaluation of} calibration according to the fit-on-test paradigm, promoting the use of post-hoc calibration methods for calibration evaluation. 
This view enables reliability diagrams that are closer to the true calibration maps, more exact estimates of the total calibration error, and ranking of calibrators in the order which better corresponds to their true quality.
Following fit-on-test, we have proposed cross-validation to tune the number of bins in ECE, and demonstrated the benefits of piecewise linearity in the original as well as in the logit-logit space, inspired by temperature scaling and beta calibration.
\insertnew{Having said that, the limitation of this approach is, as the name states, fitting on the test set. Thus, it is essential to be careful not to overfit the evaluation method and check which methods would best work on given data.}

Future work involves the development of debiasing methods for $ECE_{PL}$ and $ECE_{PL3}$ and analysing further the benefits of different calibration map families in different scenarios, including dataset shift.



\backmatter

\bmhead{Acknowledgments}

This work was supported by the Estonian Research Council grant PRG1604 and by the European Social Fund via IT Academy programme.


\section*{Statements and Declarations}

\bmhead{Funding}
This work was supported by the Estonian Research Council grant PRG1604 and by the European Social Fund via IT Academy programme.\textbf{}

\bmhead{Competing Interests}
\inserted{The authors have no competing interests to declare that are relevant to the content of this article.} 

\bmhead{Ethics approval} Not applicable.

\bmhead{Consent to participate} The authors agree to participate in the conference.

\bmhead{Consent for publication} The authors permit the publication of the article and its materials.

\bmhead{Availability of data and materials}
All the data used for the work is publicly available, and the generation of new data and the work results are included in the source code.

\bmhead{Code Availability}
The source code is available at \url{https://github.com/markus93/fit-on-test}.

\bmhead{Authors' contributions}
All the authors worked together on developing and analyzing methods and concepts. Mathematical theorems and proofs were written by Meelis Kull. Data preparation and practical side were done by Markus Kängsepp and Kaspar Valk. Writing of the article was done jointly by all of the authors.

\bibliography{main}

\newpage

\begin{appendices}


\tableofcontents

\section{Source Code}
\label{sec:source_code}

The source code is available at \url{https://github.com/markus93/fit-on-test}.
It contains everything needed to run the experiments and to get the results displayed in the article. 
The source code contains a yml-file for generating a Conda environment with the needed packages and versions.

\section{Proofs}
\label{sec:proof}

\subsection{Definitions Related to Theorems~\ref{thm:calmapfitting} and \ref{thm:cmee}}

The main paper includes 3 theorems. Theorems~\ref{thm:calmapfitting} and \ref{thm:cmee} include Bregman divergences as a way to measure dissimilarity between two probability distributions. We use Bregman divergences in the context of binary class probability estimation, in which case a probability distribution can be represented by a single real number in the range $[0,1]$, representing the probability for the positive class. Thus, in our case, Bregman divergences are functions that take in two positive class probabilities and output a real number representing the divergence of the distributions represented by these probabilities, $d:[0,1]\times[0,1]\to\sR$. Note that the order of the two arguments is such that the first is the `prediction' and the second is the `ground truth' (we have seen both orders in the literature, but have found this order more natural). The formal definition is as follows:

\begin{definition}
A function $d:[0,1]\times[0,1]\to\sR$ is called a Bregman divergence, if there exists a continuously-differentiable and strictly convex function $\phi:[0,1]\to\sR$, such that for every $p,q\in[0,1]$:
$$d(p,q)=\phi(q)-\phi(p)-(q-p)\phi'(p)$$
where $\phi'$ is the derivative of $\phi$.
\end{definition}

Theorems~\ref{thm:calmapfitting} and \ref{thm:cmee} involve random variables. As described in the main paper, we have $X$ as a randomly drawn instance (i.e. $X$ is a vector of its feature values) and $Y$ is its label. In Theorem~\ref{thm:calmapfitting} we consider a particular fixed probabilistic classifier $f:\mathcal{X}\to\sR$ and thus, its output can be viewed as a random variable also, $\Ph=f(X)$. The true calibration map of $f$ can be calculated as $c^*_f(\ph)=\E\Bigl[Y\Big\vert f(X)=\ph\Bigr]=\E\Bigl[Y\Big\vert \Ph=\ph\Bigr]$ for any $\ph\in[0,1]$. 

In Theorem~\ref{thm:cmee} we consider a particular calibrator $\ch:[0,1]\to[0,1]$, and the true calibration map of the approximately calibrated model $\ch\circ f$, which according to the definition of $c^*$ can be written out as $c^*_{\ch\circ f}(c)=\E\Bigl[Y\Big\vert (\ch\circ f)(X)=c\Bigr]=\E\Bigl[Y\Big\vert \ch(f(X))=c\Bigr]$ for any $c\in[0,1]$.

\subsection{Proof of Theorem~\ref{thm:calmapfitting}}

\begin{customthm}{1} 
Let $d:[0,1]\times[0,1]\to\sR$ be any Bregman divergence and $\ch_1,\ch_2:[0,1]\to[0,1]$ be two estimated calibration maps. Then 
$$\E\Bigl[d(\ch_1(\ph),Y)\Big\vert \Ph=\ph\Bigr]-\E\Bigl[d(\ch_2(\ph),Y)\Big\vert \Ph=\ph\Bigr]=d\Bigl(\ch_1(\ph),c^*_f(\ph)\Bigr)- d\Bigl(\ch_2(\ph),c^*_f(\ph)\Bigr).$$
\end{customthm}
\begin{proof}
It is sufficient to prove that the value of the following expression does not depend on $\ch_i$ where $i=1$ or $i=2$ and thus is the same for $\ch_1$ and $\ch_2$:
$$\E\Bigl[d(\ch_i(\ph),Y)\Big\vert \Ph=\ph\Bigr]-d\Bigl(\ch_i(\ph),c^*_f(\ph)\Bigr).$$
Let $\phi$ be a convex function that gives rise to $d$, then according to the definition of the Bregman divergence we can rewrite the above expression as follows:
\begin{multline*}
    \E\Bigl[\phi(Y)-\phi(\ch_i(\ph))-(Y-\ch_i(\ph))\phi'(\ch_i(\ph))\Big\vert \Ph=\ph\Bigr]\\
    -\Bigl(\phi(c^*_f(\ph))-\phi(\ch_i(\ph))-(c^*_f(\ph)-\ch_i(\ph))\phi'(\ch_i(\ph))\Bigr).
\end{multline*}
As $\E\Bigl[\phi(Y)\Big\vert \Ph=\ph\Bigr]$ and $\phi(c^*_f(\ph))$ do not depend on $\ch_i$ and as the terms $\phi(\ch_i(\ph))$ and $\ch_i(\ph)\phi'(\ch_i(\ph))$ both cancel out, we are left to prove that the value of the remaining expression does not depend on $\ch_i$:
$$\E\Bigl[-Y\phi'(\ch_i(\ph))\Big\vert \Ph=\ph\Bigr]+c^*_f(\ph)\phi'(\ch_i(\ph)).$$
Noting that $\phi'(\ch_i(\ph))$ does not depend on $\Ph$, it can be taken out from the expectation, and thus the expression can be written as:
$$\Bigl(-\E[Y\vert \Ph=\ph]+c^*_f(\ph)\Bigr)\phi'(\ch_i(\ph)).$$
However, this is equal to zero, since by the definition of the true calibration map $c^*_f$ we have:
$$c^*_f(\ph)=\E[Y\vert \Ph=\ph].$$
\end{proof}

\subsection{Proof of Theorem~\ref{thm:slope1}}
\begin{customthm}{2}
Consider a predictive model with predictions $\ph_1,\dots,\ph_n\in[0,1]$ on a test set with actual labels $y_1,\dots,y_n$ and a binning $\vB$ with $b\geq 1$ bins and boundaries $0=B_1<\dots<B_{b+1}=1+\epsilon$.
Then for any $\alpha>0$, the measure $\ECE^{(\alpha)}_{\vB}$ is equal to:
$$\ECE^{(\alpha)}_{\vB}=\frac{1}{n}\sum_{i=1}^n \vert \ch(\ph_i)-\ph_i\vert ^{\alpha}$$
$$\text{\it where}\quad\ch=\argmin_{c\in\cC_{(\vB,\cH,\vOne)}} \frac{1}{n}\sum_{i=1}^n (c(\ph_i)-y_i)^2$$
Furthermore, $\ch(\pb_k)=\yb_k$ for $k=1,\dots,b$, where $\pb_k$ and $\yb_k$ are the average $\ph_i$ and $y_i$ in the bin $[B_k,B_{k+1})$.
\end{customthm}
\begin{proof}
Our first goal is to prove that $\ch(\pb_k)=\yb_k$ for $k=1,\dots,b$.
To find the values of parameters at the optimum $\ch$, we study the stated minimization task and consider any $c\in\cC_{(\vB,\cH,\vOne)}$ with any values of parameters $(H_1,\dots,H_b)$.
Let us rewrite the quantity to be minimized, grouping the instances by bins and using the definition of $c(\cdot)$:
\begin{align*}
&\frac{1}{n}\sum_{i=1}^n (c(\ph_i)-y_i)^2\\
&=\frac{1}{n}\sum_{k=1}^b \sum_{\substack{i\\ \ph_i\in[B_k,B_{k+1})}} (c(\ph_i)-y_i)^2 \\
&=\frac{1}{n}\sum_{k=1}^b \sum_{\substack{i\\ \ph_i\in[B_k,B_{k+1})}} ((H_k+1(\ph_i-B_k))-y_i)^2 \\
\end{align*}
Equating the derivatives of this expression with respect to each $H_k$ to zero, we get:
\begin{align*}
\frac{1}{n}\sum_{\substack{i\\ \ph_i\in[B_k,B_{k+1})}} 2(H_k+\ph_i-B_k-y_i)=0 \\
\frac{2}{n}n_k (H_k-B_k)=\frac{2}{n}\sum_{\substack{i\\ \ph_i\in[B_k,B_{k+1})}} (y_i-\ph_i) \\
H_k-B_k=\frac{1}{n_k}\sum_{\substack{i\\ \ph_i\in[B_k,B_{k+1})}} (y_i-\ph_i) \\
H_k-B_k=\yb_k-\pb_k
\end{align*}
for each $k$.
Therefore, $H_k-B_k=\yb_k-\pb_k$ holds for the optimum at $\ch$ and we get
\begin{align*}
\ch(\pb_k)=H_k+1(\pb_k-B_k)
&=\pb_k+(H_k-B_k) \\
&=\pb_k+(\yb_k-\pb_k)=\yb_k
\end{align*}
More generally, for any $\ph\in[B_k,B_{k+1})$ we get $\ch(\ph)=H_k+1(\ph-B_k)=\ph+(H_k-B_k)=\ph+(\yb_k-\pb_k)$.
This implies that the term $\vert \pb_k-\yb_k\vert $ in the definition of $\ECE^{(\alpha)}_{\vB}$ is equal to $\vert \ch(\ph)-\ph\vert $ for any $\ph\in[B_k,B_{k+1})$.
Using this, we can rewrite $\ECE^{(\alpha)}_{\vB}$ as follows:
\begin{align*}
\ECE^{(\alpha)}_\vB
&=\frac{1}{n}\sum_{k=1}^b n_k\cdot\vert \pb_k-\yb_k\vert ^\alpha \\
&=\frac{1}{n}\sum_{k=1}^b \sum_{\substack{i\\ \ph_i\in[B_k,B_{k+1})}}\vert \pb_k-\yb_k\vert ^\alpha \\
&=\frac{1}{n}\sum_{k=1}^b \sum_{\substack{i\\ \ph_i\in[B_k,B_{k+1})}}\vert \ch(\ph_i)-\ph_i\vert ^\alpha \\
&=\frac{1}{n}\sum_{i=1}^n \vert \ch(\ph_i)-\ph_i\vert ^\alpha \\
\end{align*}
\end{proof}

\subsection{Proof of Theorem~\ref{thm:cmee}}

Theorem~\ref{thm:cmee} applies for any positive class probability estimator $f:\mathcal{X}\to[0,1]$, any calibrator $\ch:[0,1]\to[0,1]$, and any Bregman divergence $d$. 
First we remind of the definitions of $CMEE$ and $CEAC$ given in the main paper:
\begin{align*}
\text{calibration error after calibration:$\quad$}
& CEAC=\CE(\ch\circ f)=\E[d(\Ch,c^*_{\ch\circ f}(\Ch))] \\
\text{calibration map estimation error:$\quad$}
& CMEE=\E[d(\ch(\Ph),c^*_f(\Ph))]
\end{align*}
where $\Ch=\ch(\Ph)=(\ch\circ f)(X)=\ch(f(X))$.

\begin{customthm}{3} 
$CMEE=CEAC+\E[d(c^*_{\ch\circ f}(\Ch),c^*_f(\Ph))]$.
\end{customthm}
\begin{proof}
We have to prove that $CEAC+\E[d(c^*_{\ch\circ f}(\Ch),c^*_f(\Ph))]-CMEE=0$, that is:
$$\E[d(\Ch,c^*_{\ch\circ f}(\Ch))]+
\E[d(c^*_{\ch\circ f}(\Ch),c^*_f(\Ph))]
-\E[d(\Ch,c^*_f(\Ph))]=0.$$
We will prove the variant of this equality where all expectations are replaced by conditional expectations conditioned on $\Ch$, from which the original equality follows due to the law of total expectation, i.e. $\E[\E[V\vert W]]=\E[V]$ for any random variables $V$ and $W$. That is, it is sufficient to prove that:
$$\E\Bigl[d(\Ch,c^*_{\ch\circ f}(\Ch))\Big\vert \Ch\Bigr]+
\E\Bigl[d(c^*_{\ch\circ f}(\Ch),c^*_f(\Ph))\Big\vert \Ch\Bigr]
-\E\Bigl[d(\Ch,c^*_f(\Ph))\Big\vert \Ch\Bigr]=0.$$
Let $\phi$ be a convex function that gives rise to $d$, then according to the definition of the Bregman divergence we can rewrite the above equality as follows:
\begin{align*}
\E\Bigl[\phi(c^*_{\ch\circ f}(\Ch))-\phi(\Ch)-(c^*_{\ch\circ f}(\Ch)-\Ch)\phi'(\Ch)\Big\vert \Ch\Bigr] \\
+\E\Bigl[\phi(c^*_f(\Ph))-\phi(c^*_{\ch\circ f}(\Ch))-(c^*_f(\Ph)-c^*_{\ch\circ f}(\Ch))\phi'(c^*_{\ch\circ f}(\Ch))\Big\vert \Ch\Bigr]\\
-\E\Bigl[\phi(c^*_f(\Ph))-\phi(\Ch)-(c^*_f(\Ph)-\Ch)\phi'(\Ch)\Big\vert \Ch\Bigr]=0.
\end{align*}
The first two terms in each conditional expectation cancel out among the three conditional expectations, leaving us with only the last terms. Taking into account that $c^*_f(\Ph)$ is the only term which is not a constant under conditioning with $\Ch$, the above is equivalent to:
\begin{align*}
-(c^*_{\ch\circ f}(\Ch)-\Ch)\phi'(\Ch) \\
-(\E[c^*_f(\Ph)\vert \Ch]-c^*_{\ch\circ f}(\Ch))\phi'(c^*_{\ch\circ f}(\Ch))\\
+(\E[c^*_f(\Ph)\vert \Ch]-\Ch)\phi'(\Ch)=0.
\end{align*}
As $\Ch\phi'(\Ch)$ cancels out, we can reorganise the remaining terms as follows:
\begin{align*}
\Bigl(\E[c^*_f(\Ph)\vert \Ch]-c^*_{\ch\circ f}(\Ch)\Bigr)\Bigl(\phi'(\Ch)-\phi'(c^*_{\ch\circ f}(\Ch)\Bigr)=0.
\end{align*}
It now suffices to prove that $\E[c^*_f(\Ph)\vert \Ch]=c^*_{\ch\circ f}(\Ch)$ for every value of $\Ch$.
According to the definition of the true calibration maps $c^*_f(\Ph)$ and $c^*_{\ch\circ f}(\Ch)$, this equality can be rewritten as:
$$\E\Bigl[\E[Y\vert \Ph]\Big\vert \Ch\Bigr]=\E[Y\vert \Ch].$$
Since $\Ph$ functionally determines $\Ch$ through $\Ch=\ch(\Ph)$, the above equality follows directly from the law of total expectation applied on conditional expectations, i.e. $\E[\E[V\vert \mathcal{G}_2]\vert \mathcal{G}_1]=\E[V\vert \mathcal{G}_1]$ for any random variable $V$ and $\sigma$-algebras $\mathcal{G}_1\subseteq\mathcal{G}_2$.
\end{proof}

\section{Implementation Details} 
\label{sec:implementation_supp}
This section provides the implementation details for all the methods used in the experiments.
\subsection{Piecewise Linear Fitting Using a Neural Network}
\label{subsec:pwl_fitting_details}
\subsubsection{PL Details}
This section gives a more precise overview of implementation details of the piecewise linear method. Information about the overall architecture is available in the main part of the article.
Furthermore, the exact implementation is available in the source code in the file "piecewise\_linear.py". 

The model is initialised such that it represents the identity calibration map. It is trained up to 1500 epochs. 
Early stopping patience is set to 20, which means that if the training loss has not gone smaller for 20 epochs, then the fitting is stopped. 
The model is optimized using Adam~\citep{kingma2014adam} optimiser with the learning rate of $0.01$. 
The batch size is $min(n_{data}/4, 512)$, e.g. for 1000 data points the batch size is 250, and for 10000 data points it is 512. Cross-validation is used to pick the number of nodes for the binning layer. The number of nodes minus 1 gives the number of bins.
All numbers of bins from 1 bin to 16 bins are considered for the model. The only exception is when there are 1000 instances, where the maximum number of bins considered is 6.
The model is trained using the MSE or cross-entropy loss.  Cross-entropy loss (CE) was showcased in the main article, due to better performance. In the following tables of results, $PL_{NN}^{CE}$ stands for training with the CE loss.

The $PL$ method fitting for a single model takes under 15 seconds, and with 10-fold cross-validation it takes up to 150 seconds depending on the number of nodes the model has. In total, finding the best number of bins (1 to 16 bins) with 10-fold CV takes up to 25 minutes depending on the data size and complexity of the fitted function.
Mostly the model performs much faster, but it gets slower as more bins are used or the data sizes get bigger.
Further speedups can be obtained by reducing the number of folds and the different numbers of bins considered in hyperparameter optimisation.
The scripts were run in a high performance computing center using CPU processing power (Intel(R) Xeon(R) CPU E5-2660 v2 @ 2.20GHz) with up to 6GB of RAM.

\subsubsection{PL3 Details}
\label{subsubsec:PL3_details}
This section gives a further implementation details of the piecewise linear method in the logit-logit space. Similarly to PL, information about the overall architecture is available in the main part of the article and the exact implementation is available in source code in "piecewise\_linear.py".

The model training part is exactly the same as for the PL method.
 
The PL3 method fitting tends to take more time comparing to PL method. Fitting a single model takes under 30 seconds, and with 10-fold cross-validation it takes up to 300 seconds. In total, finding the best number of bins with 10-fold CV takes up to 80 minutes.
Large speedup can be obtained by reducing the number of folds and the different numbers of bins considered in hyperparameter optimisation.

\subsection{Details of Cross-Validation}
\label{subsec:cv}
In our experiments, cross-validation is used to find the best number of bins for the $ES_{CV}$, $PL$ and $PL3$ methods. We have seen that the results improve by using a simple complexity-reducing regularisation trick, according to which we prefer a lower number of bins instead of a higher number of bins whenever the relative difference in the cross-validated loss estimate is less than 0.1 percent. Furthermore, the same way as in hyperparameter optimization for Dirichlet calibration~\citep{Kull2019BeyondTS}, the predictions on test data are obtained as an average output from all
the 10 models with the chosen number of segments but trained from different folds, i.e. we are not refitting a single model on all 10 folds.

Table~\ref{table:chat_dist_c_trick} depicts the difference in results for estimating $\vert\ch_\mathcal{M}-c^*\vert$ with the classical CV and with the CV using the complexity-reducing regularisation. In Table~\ref{table:ECE_abs_trick} the same is depicted for estimating $\vert ECE_\mathcal{M}-CE\vert$. It can be seen that the complexity reducing regularisation makes results better for $\vert\ch_\mathcal{M}-c^*\vert$ estimation in most cases. The only exception is $PL3^{CE}$, where the regularisation makes the results a tiny bit worse. On the other hand, the regularisation results on $\vert ECE_\mathcal{M}-CE\vert$ are opposite, and the overall results get worse. Nevertheless, the regularisation was left in as it showed promising results on smaller data set. 
\subsection{Details About the Binning Methods}
The binning methods were implemented using NumPy and Scikit-Learn packages. The binning methods follow the approaches previously established, however cross-validation and unit-slope calibration maps are added. We have also implemented the sweep method to choose the highest number of bins just before the calibration map gets non-monotonic. All of the binning methods enable using both equal size and equal width binning. The implementation is available in the source code in the file "binnings.py".

\subsection{Implementation Details of Other Methods}
Other methods used for comparisons are taken from publicly available packages as follows:
isotonic calibration and Platt scaling are available from the pycalib package \citep{pycalib}; beta calibration from the betacal package \citep{pmlr-v54-kull17a}; KCE from the pycalibration package \citep{Widmann2019CalibrationTI} and $PL_{DE}$ from the pwlf package \citep{pwlf}. For KCE we used the unbiased version with RBFKernel. \inserted{Both, splines~\citep{spline_gupta21} and intra-order preserving (IOP) functions~\citep{IOP_rahimi20} are implemented using official implementation provided in the original articles.}
\inserted{
The best number of splines for spline methods was found using CV similarly as for piecewise linear methods (Subsection~\ref{subsec:cv}). For spline, three different methods were used: natural, parabolic and cubic. For IOP functions, the configurations given with original paper were used and two different versions: order-invariant ($IOP_{OI}$) and order-preserving ($IOP_{OP}$). Unfortunately, the third version, diagonal version, did not work.
}
For KDE we used the implementation provided in the original article \citep{Zhang2020MixnMatchEA}. 
We used point-wise estimates for KDE as they seemed to offer better results than the integral based estimates proposed in the original article. 

The best number of bins for $PL_{DE}$ was found using CV similarly as for piecewise linear methods (Subsection~\ref{subsec:cv}). The pwlf package also supports fitting piecewise quadratic curves (degree 2) in addition to the piecewise linear curves (degree 1). The degree 1 had better results than degree 2 according to the results in Table~\ref{table:chat_dist_c_CE_MSE}. For degree of 1, seven bins were chosen as the maximum limit for CV, as from there on the model fitting got very slow. For the degree of 2, five bins were chosen as the maximum limit for CV. On average, fitting $PL_{DE}$ with 10-fold CV took about 30 minutes.
The licenses of packages have been checked to be freely usable for our work.


\subsection{Debiasing ECE}
Debiasing was applied for the binning-based ECE methods. Reminding the notation, $\yb_k=\frac{1}{n_k}\sum_{\ph_i\in[B_k,B_{k+1})} y_i$ is the average label in $k$-th bin,
$\pb_k=\frac{1}{n_k}\sum_{\ph_i\in[B_k,B_{k+1})} \ph_i$ is the average prediction in $k$-th bin,
$n_k=\vert \{i\mid \ph_i\in[B_k,B_{k+1})\}\vert $ is the size of bin $k$, 
$\ECE^{(1)}$ is defined as $\ECE^{(1)}_\vB=\frac{1}{n}\sum_{k=1}^b n_k\cdot\vert \pb_k-\yb_k\vert ^1$. 

\citet{kumar2019neurips} proposed to debias $\ECE^{(1)}$ by defining $\yb_k$ in each bin as a sample from a random variable defined by a Gaussian distribution $N(\yb_k, \frac{\yb_k(1-\yb_k)}{n_k})$. Bias can then be estimated by drawing repeated samples from the same random variable. The final debiased estimate of $\CE^{(1)}$ can then be achieved by subtracting the approximated bias from the $\ECE^{(1)}$ value.

Instead of drawing repeated random samples, we propose to use integration and the probability density function of the same random variable 
for computationally faster results. 
Instead of drawing samples from $R_k \sim N(\yb_k, \frac{\yb_k(1-\yb_k)}{n_k})$ for each bin to find $\mathbb{E}[n_k\cdot\vert \pb_k-R_k\vert ]$ as proposed by \citet{kumar2019neurips}, one can find it computationally faster by finding in each bin
\[
\int_{-\infty}^{\infty} n_k \cdot \vert \pb_k - x\vert \cdot f_{R_k}(x) \,dx,
\] where $f_{R_k}$ is the probability density function of $R_k$.
We used the simple trapezoidal integration with 10k equally-spaced integration points within the area up to 5 standard deviations away from the mean.



\section{Datasets and Experimental Setup}
\label{sec:data_exp_setup}
We ran experiments on pseudo-real, synthetic and real datasets. More details about these datasets and the experimental setup is in the following sections.
We have checked the licenses of datasets and confirmed that these datasets are freely usable for this work.

\subsection{List of Methods with Shortened Names}

\inserted{
\begin{itemize}
    \item $ES_{15}$ - ECE method with equal-size binning and 15 bins, $EW$ for equal-width binning;
    \item $ES_{sweep}$ - ECE method with equal-size binning and sweep method for choosing the number of bins;
    \item $ES_{CV}$ - ECE method with equal-size binning and cross-validation for choosing the number of bins;
    \item Platt - platt scaling;
    \item beta - Beta calibration;
    \item isotonic - isotonic calibration;
    \item Scaling-Binning - scaling-binning method;
    \item TempS/VecS - temperature/vector scaling;
    \item MSODIR - matrix scaling with off-diagonal and intercept regularisation;
    \item dirODIR/dirL2 - dirichlet calibration with ODIR and L2 regularisation;
    \item 1-Temp - temperature scaling learning in 1-vs-rest fashion for binary calibration;
    \item Spline (natural/cubic/parabolic) - spline calibration with different methods;
    \item $IOP_{OI}$ and $IOP_{OP}$ - intra-order preserving functions with order-invariant and order-preserving variants;
    \item $PL_{NN}^{CE}$ and $PL_{NN}^{MSE}$- piecewise linear method with cross-entropy or MSE loss;
    \item $PL3^{CE}$ - piecewise linear method in logit-logit space;
    \item $PL_{DE}$ - piecewise linear method based on least squares fitting with differential evolution. $PL_{DE}^{2}$ - degree 2 for quadratic curves;
    \item KDE - kernel density estimation;
    \item KCE - kernel calibration error.
\end{itemize}
}

\subsection{Pseudo-Real Experiments}
The pseudo-real dataset is called CIFAR-5m~\citep{cif5m} and contains over 5 million synthetic images similar to CIFAR-10~\citep{cif} with size of 32x32 pixels. These images were created by sampling DDPM model~\citep{ddpm} trained on CIFAR-10 data. The pseudo-real data was used to get close to real data with the advantage of being able to estimate the true calibration map very precisely.

In order to estimate the true calibration map on 1 million hold-out datapoints 3 different calibration methods were used:  (1) isotonic calibration; (2) equal size binning with 100 bins with flat-top (i.e. slope 0) bins; (3) equal size binning with 100 bins with slope-1-tops (i.e. tops of the bins parallel to the main diagonal). Only minor differences across different `true' calibration maps occurred (see Tables \ref{table:chat_dist_c}, \ref{table:chat_dist_c_cgt0} and \ref{table:chat_dist_c_cgt1}).

To set up the experiments, the datasets were calibrated with various 2-class calibration methods (beta, isotonic, Platt, ScalingBinning), multiclass calibrators (TempS, VecS, MSODIR, dirODIR, \inserted{IOP, Spline}), piecewise linear methods $PL3$, $PL_{NN}$ \inserted{with cross-entropy loss}, and the method $ES_{sweep}$ adapted from calibration evaluation to calibration map fitting. Table~\ref{table:cal_methods_all} compares the results. The calibration is done in 1-vs-Rest fashion to achieve results for a 2-class problem. We have 1 confidence calibration task and 3 one-vs-Rest subtasks (car, cat, dog), for 6 calibration methods on 3 models, in total 72 combinations. The 6 calibration methods were chosen as the best methods in Table\inserted{~\ref{table:cal-method-table}} of the main article: beta, vector scaling, Platt, PL3, ScalingBinning, isotonic.

\subsection{Synthetic Experiments}
\label{subsec:syn_experiments}

\begin{figure}
 \centering
\includegraphics[width=\textwidth]{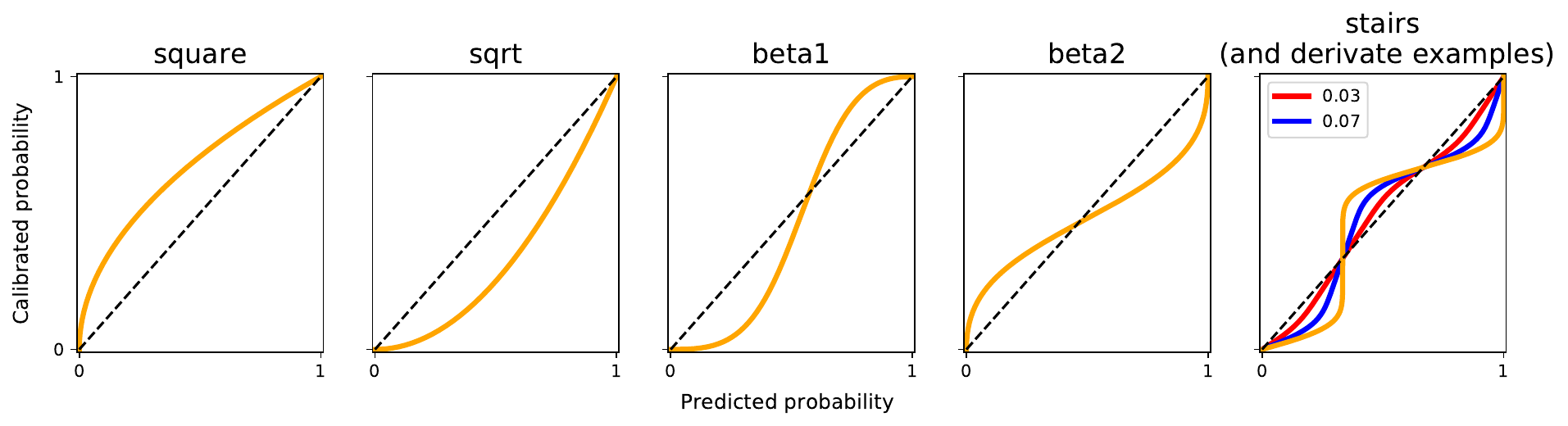}
\caption{Base shapes used for generating synthetic data. The last plot also contains two examples of derivates with expected absolute calibration errors 0.03 and 0.07 for the $\mathsf{Uniform}$ distribution.}
\label{fig:calmap_bases}
\end{figure}

For synthetic data, only the predicted probabilities $\ph$, labels $y$, and corresponding calibrated probabilities $c^*_f(\ph)$ were generated.
Synthetic data were generated based on five different base shapes (Figure \ref{fig:calmap_bases}). 
The 4 first shapes were chosen to mimic likely scenarios of calibration mappings, with combinations of over- and under-confidence for values below and above $0.5$. The 5th shape `stairs' was added as a more challenging shape that crosses the identity function in two places.

From each shape, multiple variants that we refer to as `derivates' were generated by linearly mixing the shape with the identity function in different proportions. Derivates were generated to have datasets with different expected calibration errors. 
To generate a synthetic dataset from a particular derivate, first the calibrated probabilities $c^*_f(\ph)$ were sampled from a base distribution. 
Then the labels $y$ were sampled from the calibrated probabilities. 
Finally, the derivate was used to transform the calibrated probabilities $c^*_f(\ph)$ to their corresponding predicted probabilities $\ph$. 
Therefore, the construction of derivates creates the inverse of the calibration map, and then the calibration map can be obtained from it by inverting the function.
It might seem more intuitive to first sample the predicted probabilities $\ph$ and then use the true calibration map to find the corresponding calibrated probabilities $c^*_f(\ph)$.
However, the more unintuitive approach used here has the following benefit the other approach does not. 
Namely, by first sampling $c^*_f(\ph)$ and the labels, we can generate synthetic datasets that have the exact same set of labels but different predicted probabilities.
This allows to mimic a realistic scenario where we have several models trained on the same dataset and we would like to choose the model with the lowest calibration error. E.g., if we would like to rank different post-hoc calibration methods. This fact has been used to generate Table \ref{table:syn_uniform_spearman}.
The five base shapes were defined by the following functions from $c^*_f(\ph)$ to $\ph$:
\begin{itemize}
    \item $square(x)=x^2$
    \item $sqrt(x)=\sqrt{x}$
    \item $beta1(x)=1 / (1 + 1 / (e^c\cdot x^a / (1-x)^b)),\\ 
\text{ where }a=0.4,b=0.45,c=b\ln{(0.6)}-a\ln(0.4)$
    \item $beta2(x)=1/(1+1/(e^c\cdot x^a/(1-x)^b)),\\ 
\text{ where } a=2, b=2.2, c=b\ln{(0.52)}-a\ln(0.48)$
    \item $stairs(x)=stairs\_helper(x+1/3) - stairs\_helper(1/3),\\ 
\text{ where } stairs\_helper(x)=
        step(step(3x\pi))/(3\pi), \text{ and } step(x)=x-\sin(x)
        $
\end{itemize}

Synthetic data were generated for data sizes 1000, 3000, 10000 with 5 different data seeds. The calibrated predictions $c^*_f(\ph)$ were sampled 
\inserted{from the uniform distribution}. Derivates with expected absolute calibration errors $0.00,0.005, 0.01,\dots,0.10$ were generated for each of the 5 base shapes. In total, we used 3 data sizes, 5 random seeds, 5 base shapes, and 21 derivates per shape.
In Figure \ref{fig:calmap_bases}, two examples of derivates for the "stairs" function are also shown. Note that the derivates have been obtained by linear mixing with the identity function in the horizontal direction because of creating the inverse calibration maps first.

\inserted{For Figure~\ref{fig:1_calmap_reldiag} and Figure~\ref{fig:1_part2_pl_cv_reldiag} in the main article 3000 data points were generated in the way described above. Uniform distribution and the base function $beta2$ were used.}

\subsection{Real Experiments}

The real dataset contains 60k images from CIFAR-10/100~\citep{cif} with size of 32x32 pixels with 10 or 100 classes. Each of the datasets have 5k validation and 10k train instances.
In total, there are 10 model-dataset combinations, the model outputs have been calibrated using 5 different methods (TempS~\citep{Guo2017}, VecS~\citep{Guo2017}, MS\_ODIR~\citep{Guo2017}, dir\_L2~\citep{Kull2019BeyondTS}, dir\_ODIR~\citep{Kull2019BeyondTS}) using the validation set. This gives us 50 datasets of model and calibration method combinations. 
To have a comparison with the synthetic data, we extracted test data subsets of sizes 1000 (10 sets), 3000 (3 sets), 10000 (1 set). The number of sets is also multiplied by 5, as there were 5 different calibration methods. 
In total, we got 500 sets with 1000 instances, 150 sets with 3000 instances, and 50 sets with 10000 instances.

\section{Visualizations of Calibration}
\subsection{Comparisons of Reliability Diagrams}
\label{subsec:comp_of_rel_diag}
\begin{figure}
 \centering
\includegraphics[width=\textwidth]{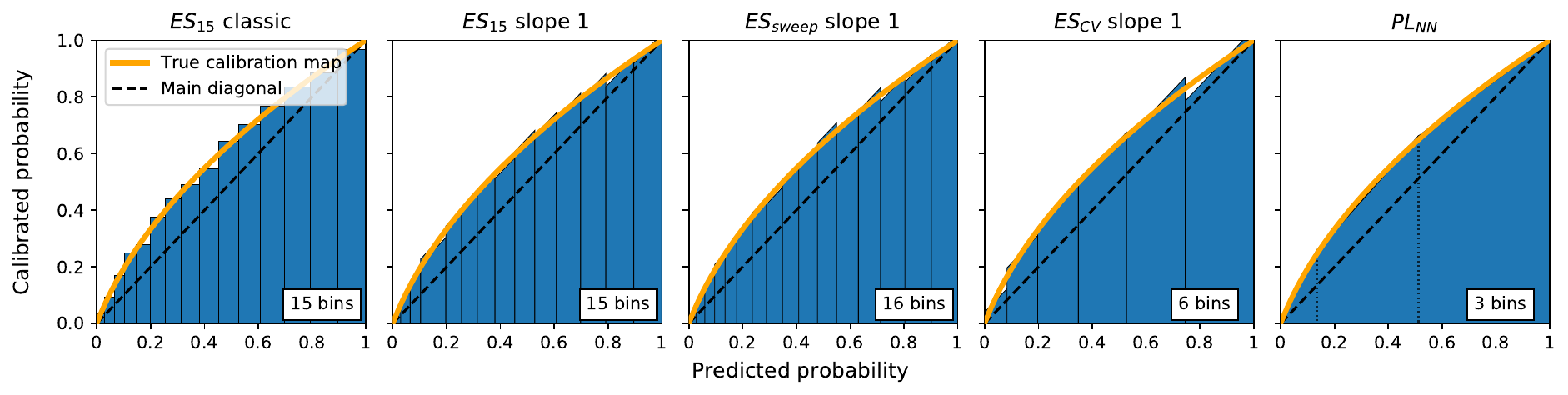}
\caption{Reliability diagrams on synthetic data. 10k datapoints. Derivate with 0.10 expected absolute calibration error for $square$.}
\label{fig:square}
\end{figure}

\begin{figure}
 \centering
\includegraphics[width=\textwidth]{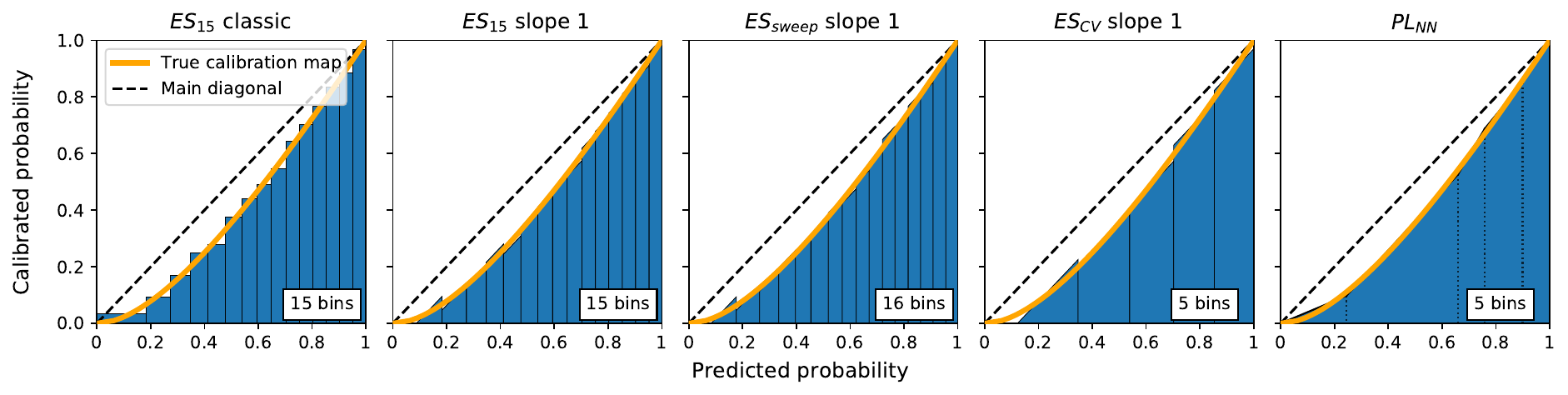}
\caption{Reliability diagrams on synthetic data. 10k datapoints. Derivate with 0.10 expected absolute calibration error for $sqrt$.}
\label{fig:sqrt}
\end{figure}

\begin{figure}
 \centering
\includegraphics[width=\textwidth]{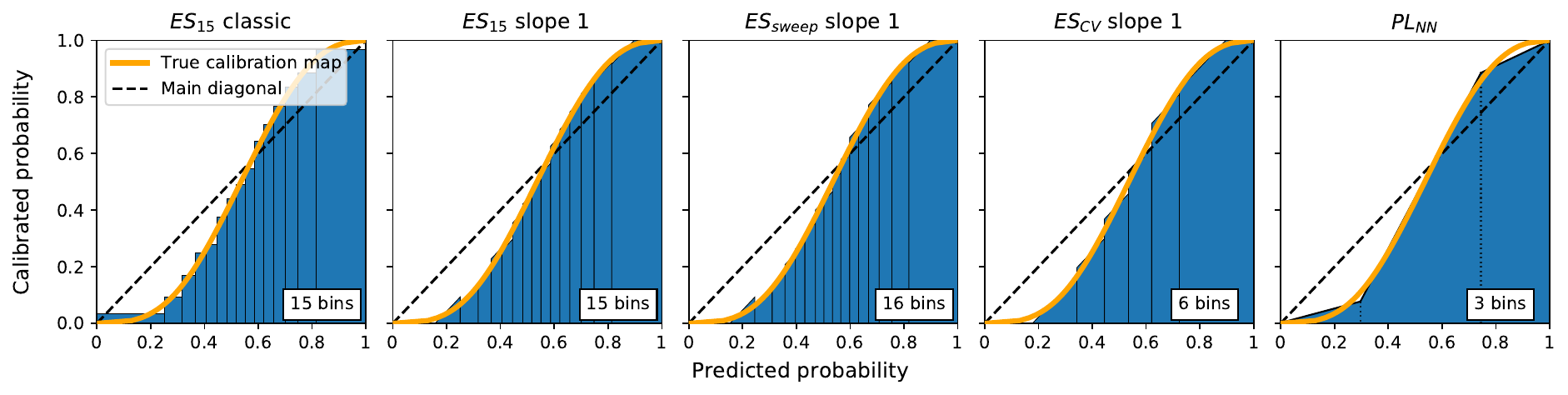}
\caption{Reliability diagrams on synthetic data. 10k datapoints. Derivate with 0.10 expected absolute calibration error for $beta1$.}
\label{fig:beta1}
\end{figure}

\begin{figure}
 \centering
\includegraphics[width=\textwidth]{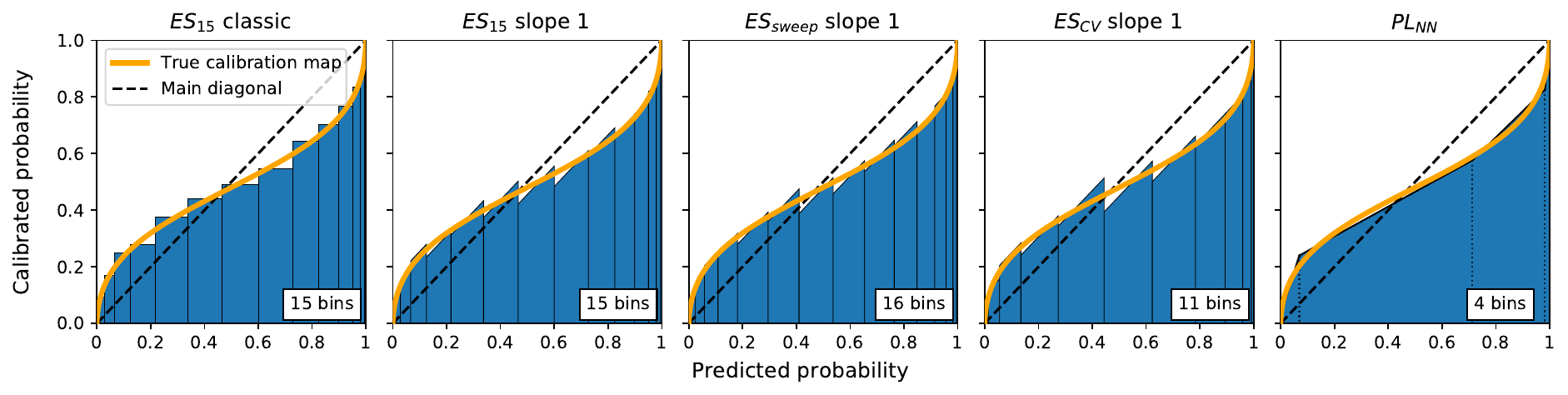}
\caption{Reliability diagrams on synthetic data. 10k datapoints. Derivate with 0.10 expected absolute calibration error for $beta2$.}
\label{fig:beta2}
\end{figure}

\begin{figure}
 \centering
\includegraphics[width=\textwidth]{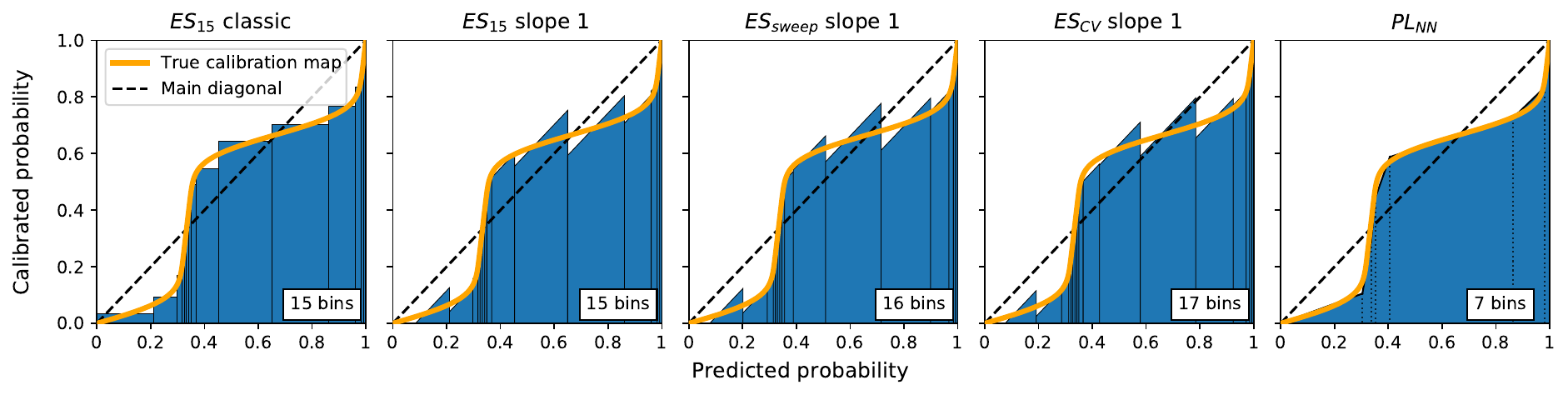}
\caption{Reliability diagrams on synthetic data. 10k datapoints. Derivate with 0.10 expected absolute calibration error for $stairs$.}
\label{fig:stairs}
\end{figure}

Figures \ref{fig:square}-\ref{fig:stairs} depict comparisons between different reliability diagrams obtained on synthetic data with 10000 instances, true calibration error 0.1 and different calibration functions. The middle three diagrams of every figure are depicted with slope 1 (CE estimation by fitting). Therefore, the diagrams for $ES_{sweep}$ might not seem monotonic, but they would be if they were plotted with the classic flat bin roofs. Based on the figures, the piecewise linear method is able to follow the true calibration map the best.

\subsection{Calibration Maps in the Logit-Logit Scale}
\label{subsec:calmap_in_ll_scale}
As proved in the main paper, the temperature scaling method applied in binary classification fits a calibration map which is linear in the logit-logit scale.
Furthermore, here we show that the beta calibration method fits a calibration map which is approximately piecewise linear in the logit-logit scale with 2 pieces.

\begin{figure}[ht!]
\begin{center}
\centerline{\includegraphics[width=0.7\columnwidth]{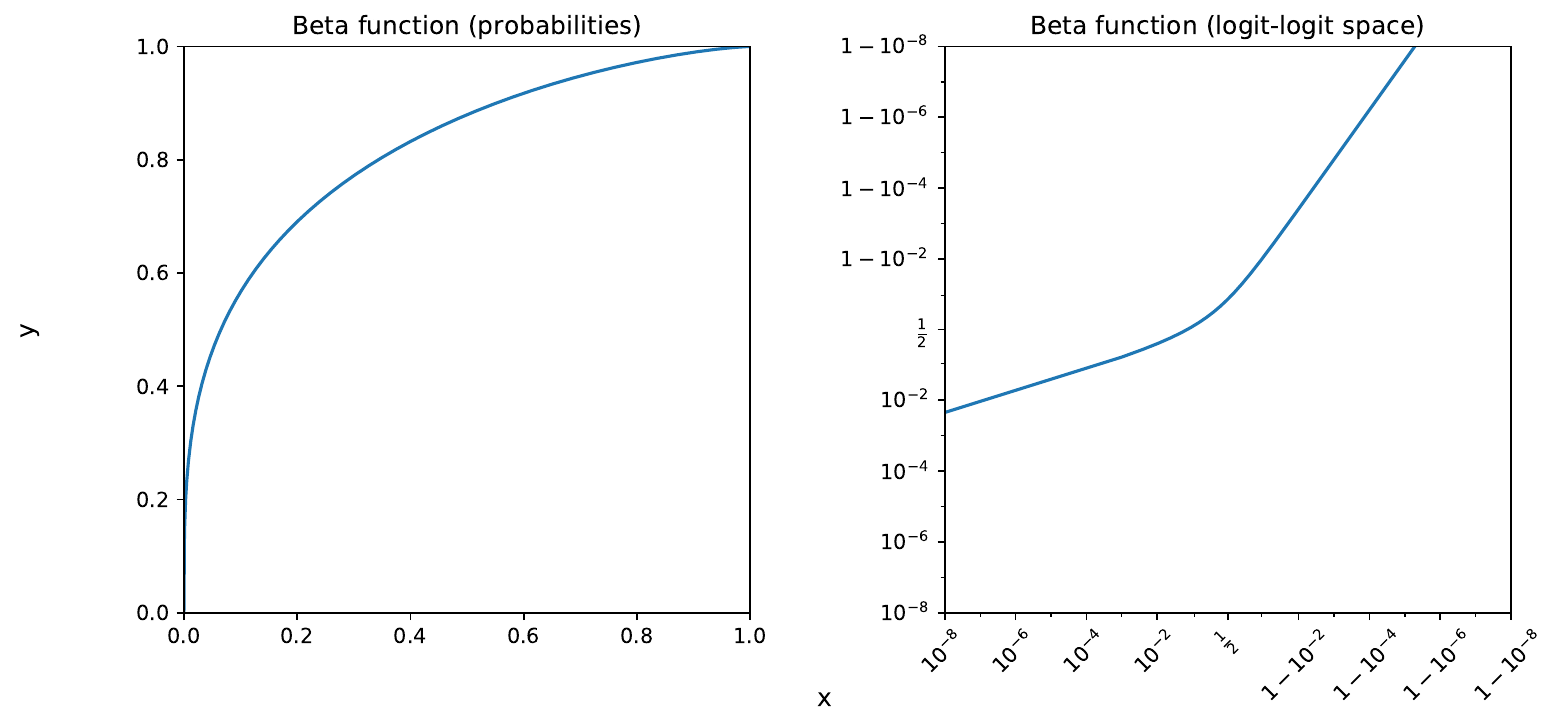}}
\caption{Left: beta calibration map with parameters ($a=0.3; b=1.4; c=0.0$); Right: the same calibration map in the logit-logit space, illustrating that beta calibration approximately fits a piecewise linear calibration map with 2 segments.}
\label{fig:beta_logit}
\end{center}
\vskip -0.1in
\end{figure}

Consider the calibration map family of beta calibration, $\ch(\ph)=\frac{1}{1+1/\left(e^c\frac{\ph^a}{(1-\ph)^b}\right)}$. 
Changing the y-axis to logit scale, we get 
$$\ln\Bigl(\frac{\ch(\ph)}{1-\ch(\ph)}\Bigr)=\ln\Bigl(e^c\frac{\ph^a}{(1-\ph)^b}\Bigr)=c+a\ln\ph-b\ln(1-\ph).$$
We can rewrite it in two ways:
\begin{align*}
\ln\Bigl(\frac{\ch(\ph)}{1-\ch(\ph)}\Bigr)
&=c+a\ln\Bigl(\frac{\ph}{1-\ph}\Bigr)+(a-b)\ln(1-\ph) \\
\ln\Bigl(\frac{\ch(\ph)}{1-\ch(\ph)}\Bigr)
&=c+b\ln\Bigl(\frac{\ph}{1-\ph}\Bigr)+(a-b)\ln\ph.
\end{align*}
For low values of $\ph$ near  $0$, the term $(a-b)\ln(1-\ph)$ in the first way of writing is nearly zero, and thus we get a linear approximation in the logit-logit scale with slope $a$. For high values of $\ph$ near  $1$, the term $(a-b)\ln\ph$ in the second way of writing is nearly zero, and thus we get a linear approximation in the logit-logit scale with slope $b$. This can be clearly seen visually in Figure~\ref{fig:beta_logit}, where the breakpoint between the two pieces is at $\ph=0.5$ and near this point the two linear segments are interpolated non-linearly.

\modified{Figure~\ref{fig:pl3_motivation_1} from the main article} shows a case from the pseudo-real dataset. As seen in the figure on the right, PL3 has been able to approximate most of the true calibration map quite well in the logit-logit space piecewise linearly, with two long linear segments (and a third short segment to the right of $1-10^{-6}$). In contrast, beta calibration has failed to capture these segments because it is bound to have the breakpoint between the two segments at $\ph=0.5$.

\section{Results}
\label{sec:results_supp}

\subsection{Results of Pseudo-Real Experiments}

Here is a brief guide to how we have arranged the tables with the results:
\begin{itemize}
    \item Table~\ref{table:cal_methods_all}. Pseudo-Real: Calibration method comparison;
    \item Tables~\ref{table:chat_dist_c} and~\ref{table:chat_dist_c_sq}. Pseudo-Real: Calibration Maps - absolute and square errors;
    \item Tables~\ref{table:ECE_abs} and~\ref{table:ECE_sq}. Pseudo-Real: ECE - absolute and square errors;
    \item Tables~\ref{table:chat_dist_c_cgt0}, \ref{table:chat_dist_c_cgt1},  \ref{table:cifar5m_ECE_ranking_cgt0} and \ref{table:cifar5m_ECE_ranking_cgt1}. Pseudo-Real: Different ground-truths;
    \item Tables~\ref{table:chat_dist_c_ndata} and~\ref{table:ECE_abs_ndata}. Pseudo-Real: Comparison of different numbers of instances;
    \item Tables~\ref{table:chat_dist_c_CE_MSE} and~\ref{table:ECE_abs_CE_MSE}. Pseudo-Real: CE and MSE comparison, including $PL_{DE}$ with degree 1 and 2;
    \item Tables~\ref{table:chat_dist_c_es_ew} and~\ref{table:ECE_abs_es_ew}. Pseudo-Real: Equal-size vs equal-width;
    \item Tables~\ref{table:chat_dist_c_trick} and~\ref{table:ECE_abs_trick}. Pseudo-Real: CV Regularisation;
    \item Tables~\ref{table:chat_dist_c_std} and~\ref{table:ECE_abs_std}. Pseudo-Real: Standard Deviations;
    \item Tables~\ref{table:syn_uniform_calmap}, \ref{table:syn_uniform_ece}, \ref{table:syn_uniform_ece_sq} and~\ref{table:syn_uniform_spearman}. Synthetic: results;
    \item Table~\ref{table:real_data_biases}. Real: biases.
\end{itemize}

\begin{table}
\caption{$PL$-methods, binning methods and various other calibration methods used for calibration. Evaluated on CIFAR-5m. The table displays thousandths of $\vert\ch_\mathcal{M}-c^*\vert$. The value of $c\inserted{^*}$ was found by estimating the true calibration map on $10^6$ unseen data with isotonic regression. Table is shown as two lines for better readability.}
\label{table:cal_methods_all}
\centering
\begin{adjustbox}{width=1\textwidth}
\begin{tabular}{ll|lllll|lllp{3.5em}l}
\toprule
       & Cal. Fn. &  $ES_{sweep}$ &            $PL3^{CE}$ &          $PL3^{MSE}$ & $PL_{NN}^{CE}$ & $PL_{NN}^{MSE}$ &                Platt &          beta &     isotonic & Scaling-Binning &         TempS        \\
Model & {} &               &                       &                      &                &                 &                      &               &              &                &                    \\
\midrule
ResNet110 & cars vs rest &   $7.89_{18}$ &            $3.88_{3}$ &          $4.49_{11}$ &     $3.73_{2}$ &      $4.36_{9}$ &            $3.9_{4}$ &    $4.27_{7}$ &   $4.34_{8}$ &     $4.13_{6}$ &   $8.75_{19}$  \\
       & cats vs rest &  $23.93_{20}$ &            $6.95_{2}$ &  $\mathbf{6.92_{1}}$ &   $12.61_{11}$ &     $11.31_{8}$ &          $11.26_{7}$ &   $13.2_{12}$ &  $10.92_{6}$ &   $11.66_{10}$ &  $17.45_{19}$  \\
       & dogs vs rest &  $28.48_{20}$ &   $\mathbf{3.28_{1}}$ &           $4.39_{2}$ &   $10.32_{10}$ &    $10.69_{11}$ &           $6.87_{4}$ &  $11.28_{12}$ &   $6.43_{3}$ &     $7.74_{5}$ &  $13.16_{19}$\\
       & confidence &  $21.19_{12}$ &  $\mathbf{10.64_{1}}$ &           $11.1_{3}$ &    $12.69_{7}$ &     $12.62_{5}$ &          $10.75_{2}$ &   $12.68_{6}$ &  $13.61_{9}$ &    $11.56_{4}$ &  $23.92_{13}$  \\
       \hline
DenseNet40 & cars vs rest &   $6.45_{20}$ &            $2.77_{3}$ &           $3.8_{10}$ &    $5.96_{15}$ &     $5.44_{14}$ &           $3.34_{6}$ &   $4.99_{13}$ &   $3.57_{7}$ &      $3.3_{5}$ &    $3.68_{9}$ \\
       & cats vs rest &  $10.43_{11}$ &             $3.6_{2}$ &           $5.72_{4}$ &   $10.52_{12}$ &      $8.29_{8}$ &  $\mathbf{3.32_{1}}$ &  $15.88_{20}$ &   $7.07_{6}$ &    $9.19_{10}$ &  $10.77_{13}$  \\
       & dogs vs rest &  $21.94_{20}$ &            $3.55_{2}$ &           $3.99_{3}$ &    $10.1_{14}$ &    $10.37_{15}$ &  $\mathbf{3.38_{1}}$ &  $14.55_{19}$ &   $6.62_{6}$ &     $6.07_{4}$ &   $9.59_{12}$  \\
       & confidence &   $23.5_{13}$ &   $\mathbf{5.54_{1}}$ &            $5.9_{2}$ &    $12.77_{7}$ &      $7.98_{4}$ &          $13.93_{8}$ &  $18.31_{11}$ &   $9.38_{5}$ &      $6.1_{3}$ &  $21.56_{12}$  \\
       \hline
Wide32 & cars vs rest &   $6.83_{13}$ &   $\mathbf{2.15_{1}}$ &           $4.42_{8}$ &    $4.94_{11}$ &      $4.48_{9}$ &           $3.53_{3}$ &    $3.66_{4}$ &   $3.99_{6}$ &    $4.88_{10}$ &   $9.34_{18}$ \\
       & cats vs rest &  $23.62_{20}$ &            $5.29_{2}$ &  $\mathbf{5.28_{1}}$ &   $13.73_{13}$ &    $11.99_{10}$ &           $5.68_{3}$ &  $15.86_{18}$ &   $9.05_{5}$ &     $7.61_{4}$ &  $13.87_{15}$  \\
       & dogs vs rest &  $26.11_{20}$ &            $3.59_{3}$ &           $3.56_{2}$ &   $11.47_{13}$ &      $10.2_{8}$ &  $\mathbf{3.38_{1}}$ &  $17.29_{19}$ &    $7.4_{6}$ &     $6.19_{5}$ &  $12.49_{17}$  \\
       & confidence &  $26.96_{15}$ &            $9.27_{4}$ &  $\mathbf{8.24_{1}}$ &   $20.67_{10}$ &      $8.73_{3}$ &           $14.6_{7}$ &  $21.23_{11}$ &  $11.81_{6}$ &     $8.33_{2}$ &  $22.53_{13}$ \\
\hline
       & avg rank &        $16.8$ &                 $2.1$ &                $4.0$ &         $10.4$ &           $8.7$ &                $3.9$ &        $12.7$ &        $6.1$ &          $5.7$ &        $14.9$  \\
\bottomrule
\end{tabular}
\end{adjustbox}

\begin{adjustbox}{width=1\textwidth}
\begin{tabular}{ll|lllllp{3.5em}p{3.5em}p{3.5em}p{3.5em}p{3.5em}}
\toprule
       & Cal. Fn. &         1-Temp &          VecS &         dirL2 &       dirODIR &        MSODIR & Spline natural &  Spline cubic & Spline parabolic &                $IOP_{OI}$ &        $IOP_{OP}$ \\
Model & {} &                &                      &               &               &               &      & & & &          \\
\midrule
ResNet110 & cars vs rest &     $\mathbf{3.0_{1}}$ &   $4.84_{13}$ &   $5.53_{14}$ &    $3.94_{5}$ &   $4.53_{12}$ &    $6.46_{17}$ &      $6.45_{16}$ &   $6.45_{15}$ &    $9.7_{20}$ &   $4.45_{10}$ \\
       & cats vs rest &         $16.17_{18}$ &   $11.57_{9}$ &  $13.92_{15}$ &  $13.46_{14}$ &  $13.37_{13}$ &      $9.9_{3}$ &       $9.93_{4}$ &    $10.0_{5}$ &  $15.27_{17}$ &  $15.25_{16}$ \\
       & dogs vs rest &           $9.46_{8}$ &    $9.64_{9}$ &  $13.11_{18}$ &  $12.63_{15}$ &  $12.88_{16}$ &     $8.46_{7}$ &       $8.35_{6}$ &   $11.6_{13}$ &  $12.05_{14}$ &  $12.97_{17}$ \\
       & confidence &       $30.13_{15}$ &  $34.78_{16}$ &   $37.0_{20}$ &  $36.66_{19}$ &  $36.02_{17}$ &   $14.48_{11}$ &     $14.14_{10}$ &   $13.48_{8}$ &  $25.42_{14}$ &  $36.22_{18}$ \\
       \hline
DenseNet40 & cars vs rest &    $\mathbf{2.27_{1}}$ &    $3.05_{4}$ &   $6.28_{16}$ &    $3.67_{8}$ &   $3.99_{11}$ &    $6.33_{19}$ &      $6.33_{18}$ &   $6.32_{17}$ &   $4.72_{12}$ &     $2.6_{2}$ \\
       & cats vs rest &          $12.31_{17}$ &    $8.59_{9}$ &  $15.77_{19}$ &  $12.67_{18}$ &  $11.42_{16}$ &     $6.92_{5}$ &       $7.64_{7}$ &     $5.3_{3}$ &  $10.81_{14}$ &  $11.18_{15}$ \\
       & dogs vs rest &          $6.22_{5}$ &    $7.35_{8}$ &   $13.4_{18}$ &  $11.53_{17}$ &  $10.65_{16}$ &     $7.34_{7}$ &       $7.55_{9}$ &   $9.78_{13}$ &   $9.38_{10}$ &   $9.58_{11}$ \\
       & confidence &       $24.32_{14}$ &  $28.76_{17}$ &  $40.19_{20}$ &  $34.82_{19}$ &  $32.84_{18}$ &    $15.0_{10}$ &      $14.98_{9}$ &   $10.41_{6}$ &  $26.65_{16}$ &  $24.78_{15}$ \\
       \hline
Wide32 & cars vs rest &    $9.63_{19}$ &    $3.42_{2}$ &   $4.96_{12}$ &    $3.68_{5}$ &    $4.42_{7}$ &    $7.16_{16}$ &      $7.16_{15}$ &   $7.15_{14}$ &  $10.04_{20}$ &   $8.39_{17}$ \\
       & cats vs rest &     $22.83_{19}$ &   $11.07_{9}$ &  $15.35_{17}$ &  $13.67_{12}$ &  $13.85_{14}$ &     $9.66_{6}$ &       $9.82_{8}$ &     $9.8_{7}$ &  $14.56_{16}$ &  $12.85_{11}$ \\
       & dogs vs rest &      $4.97_{4}$ &    $8.74_{7}$ &  $11.66_{15}$ &   $10.7_{11}$ &  $10.83_{12}$ &    $10.39_{9}$ &     $10.65_{10}$ &  $12.92_{18}$ &  $11.58_{14}$ &  $12.13_{16}$ \\
       & confidence &      $26.13_{14}$ &  $38.94_{17}$ &  $45.22_{20}$ &  $42.89_{18}$ &   $43.5_{19}$ &    $16.69_{9}$ &      $15.54_{8}$ &   $10.09_{5}$ &  $22.37_{12}$ &  $30.46_{16}$ \\
\hline
       & avg rank &              $11.2$ &        $10.0$ &        $17.0$ &        $13.4$ &        $14.2$ &          $9.9$ &           $10.0$ &        $10.3$ &        $14.9$ &        $13.7$ \\
\bottomrule
\end{tabular}
\end{adjustbox}

\end{table}

\subsubsection{Results of Estimating the Reliability Diagram With $\vert\ch_\mathcal{M}-c^*\vert$ and $\vert\ch_\mathcal{M}-c^*\vert^2$}
The following paragraphs have results of estimating the reliability diagram with $\vert\ch_\mathcal{M}-c^*\vert$ \inserted{and $\vert\ch_\mathcal{M}-c^*\vert^2$} on CIFAR-5m.

Results on CIFAR-5m indicate that the performance for measures vary over different methods. However, beta calibration and piecewise linear methods PL and PL3 with cross-entropy loss are in the lead. Beta calibration has average rank of \modified{4.6}, $PL3^{CE}$ has average rank of \modified{4.2}, \inserted{$PL_{NN}^{MSE}$ has average rank of 3.4} and $PL_{NN}^{CE}$ has average rank of \modified{1.2} (Table~\ref{table:chat_dist_c}). In case of quadratic loss, \modified{$PL_{NN}^{CE}$ and $PL_{NN}^{MSE}$ perform the best, followed by $IOP_{OI}$, $PL3^{CE}$, $ES_{10}$ and beta calibration} (Table~\ref{table:chat_dist_c_sq}). Note that $PL_{NN}^{MSE}$ performs well, however it is clearly outperformed by $PL_{NN}^{CE}$.

The number of data instances (Table~\ref{table:chat_dist_c_ndata}), does not change much the overall ordering. However, the more data instances, the better all models are, at getting closer to true calibration map. Additionally, as stated in the main article, the beta calibration works really well with smaller data sizes - 1000, 3000, similarly $IOP_{OI}$.

Next, the cross-entropy loss seems to be beneficial for both $PL3$ and $PL_{NN}$ methods (Table~\ref{table:chat_dist_c_CE_MSE}). The method $PL_{DE}$ with degree of 1 is performing better than with degree of 2. Overall, $PL_{DE}$ is mostly behind $PL3$ and $PL_{NN}$ with CE loss.

As expected, equal-size (ES) binning methods performed much better than equal-width (EW) binning methods (Table~\ref{table:chat_dist_c_es_ew}).
Furthermore, comparing $ES_{sweep}$ with $ES_{CV}$, CV variant gets better results 8 times out of 12 and thus getting also better average ranking 
\inserted{8.8} compared to average rank 
\inserted{11.3} of sweep (Table~\ref{table:chat_dist_c}).

The standard deviations (Table~\ref{table:chat_dist_c_std}) of calibration measures are really similar, only $PL_{DE}$ and $KDE$ perform worse than other methods. Interestingly, the standard deviations are generally higher for DenseNet40 with confidence dataset.

Different estimated ground-truths (isotonic; equal-size binning with flat tops; with slope 1 tops) of CIFAR-5m data, result in only minor differences (Tables~\ref{table:chat_dist_c}, \ref{table:chat_dist_c_cgt0}, \ref{table:chat_dist_c_cgt1}).

\subsubsection{Results of Estimating the True Calibration Error with $\vert ECE_\mathcal{M}-CE\vert$ and $\vert ECE_\mathcal{M}-CE\vert^2$}
The following paragraphs have results of estimating the true calibration error \modified{with $\vert ECE_\mathcal{M}-CE\vert$ and $\vert ECE_\mathcal{M}-CE\vert^2$} on CIFAR-5m.

$ECE_{15}$ (among all $ES$) and $PL3^{CE}$ are the best at estimating calibration error, with average ranks of 3.8 and 5.2 (Table~\ref{table:ECE_abs}). On the other hand, the best measures for \modified{the quadratic loss are $PL_{NN}^{CE}$ with the average rank of 2.8 and $IOP_{OI}$ with the average rank of 4.2 (Table~\ref{table:ECE_sq}).}

The number of data instances (Table~\ref{table:ECE_abs_ndata}), does not change the ranking \inserted{for most methods}, similarly to estimating the true calibration map. \inserted{In comparison between other methods, $ES_{sweep}$, $beta$, and $IOP_{OI}$ work better with less data, while $ES_{20}$, $ES_{25}$, $ES_{CV}$ work better with more data}. Furthermore, again \inserted{all }the methods \modified{get more accurate with more data}.

The cross-entropy loss is beneficial for $PL3$, but $PL_{NN}$ with MSE loss is better at estimating $\vert ECE_\mathcal{M}-CE\vert$ (Table~\ref{table:ECE_abs_CE_MSE}). Again, $PL_{DE}$ with degree 1 is outperforming degree 2. $PL3$ with CE loss is the best at estimating $\vert ECE_\mathcal{M}-CE\vert$ among piecewise linear methods.

For estimating $\vert ECE_\mathcal{M}-CE\vert$, equal-size binning methods performed much better than equal-width binning methods (Table~\ref{table:ECE_abs_es_ew}), the only exception is $ES_{10}$, where equal-size performs similarly to equal-width binning.
Next, comparing $ES_{sweep}$ with $ES_{CV}$, $ES_{sweep}$ is better at absolute error, however $ES_{CV}$ is better at quadratic error (Tables~\ref{table:ECE_abs} and \ref{table:ECE_sq}).

The standard deviations (Table~\ref{table:ECE_abs_std}) of calibration measures are really similar, only $PL_{DE}$ and $KDE$ perform worse than other methods.

Different estimated ground-truths (isotonic; equal-size binning with flat tops; with slope 1 tops) of CIFAR-5m data, result in only minor differences; however, it does change the top-1 ranking method (Tables \ref{table:cifar5m_ECE_ranking_cgt0}, \ref{table:cifar5m_ECE_ranking_cgt1} and Table~\ref{table:cifar5m-ECE-ranking} from the main article ).

\begin{table}
\caption{Comparison of calibration evaluators in estimating the reliability diagram with $\vert\ch_\mathcal{M}-c^*\vert$ on CIFAR-5m. The initial model predictions have been calibrated by 6 different methods on 5k data points. Corresponding methods (columns) have been then used to estimate the true calibration map on a new unseen test set of size 1k, 3k, 10k. Average of 5 test set data seeds. In total, a single cell value is the average of 6x3x5=90 calibration map estimates. The value of $c\inserted{^*}$ was found by estimating the true calibration map on $10^6$ unseen data with isotonic regression. The table displays thousandths. Table is shown as two lines for better readability.}
\label{table:chat_dist_c}
\centering
\begin{adjustbox}{width=1\textwidth}
\begin{tabular}{ll|lllllll|lllll}
\toprule
       & binning &     $ES_{10}$ &     $ES_{15}$ &     $ES_{20}$ &     $ES_{25}$ &     $ES_{30}$ &  $ES_{sweep}$ &     $ES_{CV}$ &   $PL3^{CE}$ &  $PL3^{MSE}$ &       $PL_{NN}^{CE}$ & $PL_{NN}^{MSE}$ &    $PL_{DE}$ \\
Model & {} &               &               &               &               &               &               &               &              &              &                      &                 &              \\
\midrule
resnet110 & cars vs rest &   $4.38_{14}$ &    $3.78_{6}$ &    $3.71_{5}$ &   $3.98_{11}$ &    $3.82_{7}$ &   $4.64_{16}$ &     $3.9_{9}$ &   $3.52_{3}$ &  $4.57_{15}$ &           $3.18_{2}$ &      $3.64_{4}$ &  $3.92_{10}$ \\
       & cats vs rest &    $8.94_{4}$ &   $10.4_{12}$ &  $10.37_{10}$ &  $10.86_{13}$ &  $10.87_{14}$ &   $11.1_{15}$ &   $10.04_{6}$ &   $7.72_{2}$ &    $9.7_{5}$ &  $\mathbf{7.44_{1}}$ &       $8.3_{3}$ &  $10.17_{9}$ \\
       & dogs vs rest &    $8.28_{7}$ &   $9.24_{12}$ &   $9.24_{13}$ &   $9.87_{15}$ &  $11.04_{19}$ &    $8.53_{8}$ &   $9.07_{11}$ &    $7.0_{2}$ &   $8.93_{9}$ &  $\mathbf{6.37_{1}}$ &      $7.73_{4}$ &  $8.97_{10}$ \\
       & confidence &   $12.69_{9}$ &  $14.29_{13}$ &  $15.77_{17}$ &  $17.61_{18}$ &  $19.11_{19}$ &   $11.89_{8}$ &  $13.86_{11}$ &  $11.09_{4}$ &  $11.63_{6}$ &  $\mathbf{9.79_{1}}$ &     $10.84_{3}$ &  $11.76_{7}$ \\
              \midrule

densenet40 & cars vs rest &    $7.6_{19}$ &   $4.95_{15}$ &   $5.15_{18}$ &   $5.04_{17}$ &   $4.68_{14}$ &   $4.97_{16}$ &    $3.79_{7}$ &   $3.32_{4}$ &  $4.53_{12}$ &  $\mathbf{2.43_{1}}$ &      $3.21_{3}$ &   $4.21_{9}$ \\
       & cats vs rest &  $10.04_{12}$ &  $11.93_{18}$ &   $9.61_{11}$ &  $11.33_{17}$ &  $12.41_{19}$ &  $10.08_{13}$ &    $9.12_{9}$ &   $7.16_{3}$ &   $8.49_{7}$ &  $\mathbf{5.72_{1}}$ &      $6.61_{2}$ &    $7.6_{4}$ \\
       & dogs vs rest &    $7.65_{8}$ &   $8.96_{14}$ &   $9.73_{17}$ &  $10.38_{18}$ &  $10.64_{19}$ &   $8.27_{11}$ &   $8.02_{10}$ &   $6.79_{5}$ &   $7.66_{9}$ &  $\mathbf{5.02_{1}}$ &      $6.17_{3}$ &   $6.94_{6}$ \\
       & confidence &  $12.38_{10}$ &  $14.09_{13}$ &  $17.03_{17}$ &   $18.1_{18}$ &  $19.75_{19}$ &   $11.04_{3}$ &  $12.55_{11}$ &  $11.05_{4}$ &  $12.09_{9}$ &          $10.75_{2}$ &     $11.56_{7}$ &  $11.37_{6}$ \\
              \midrule

wide32 & cars vs rest &    $3.75_{7}$ &    $3.48_{4}$ &    $3.65_{5}$ &    $3.78_{9}$ &   $3.93_{12}$ &   $4.22_{14}$ &    $3.76_{8}$ &  $3.89_{11}$ &  $5.06_{16}$ &  $\mathbf{2.62_{1}}$ &      $3.22_{3}$ &   $3.66_{6}$ \\
       & cats vs rest &   $9.94_{16}$ &   $9.01_{11}$ &  $10.09_{17}$ &  $10.81_{19}$ &  $10.51_{18}$ &   $9.26_{14}$ &    $8.57_{7}$ &   $7.43_{3}$ &  $9.29_{15}$ &  $\mathbf{6.75_{1}}$ &      $7.41_{2}$ &   $8.8_{10}$ \\
       & dogs vs rest &    $8.42_{7}$ &   $9.31_{12}$ &  $10.11_{17}$ &  $11.56_{19}$ &  $11.29_{18}$ &   $9.46_{15}$ &   $8.91_{10}$ &   $8.19_{6}$ &  $9.41_{14}$ &  $\mathbf{6.01_{1}}$ &      $6.89_{2}$ &   $8.53_{8}$ \\
       & confidence &   $11.45_{9}$ &   $14.2_{15}$ &  $15.43_{17}$ &  $17.51_{18}$ &  $19.01_{19}$ &    $9.95_{3}$ &   $10.63_{6}$ &  $10.03_{4}$ &  $10.64_{7}$ &   $\mathbf{9.2_{1}}$ &     $10.05_{5}$ &   $9.76_{2}$ \\
\hline
       & avg rank &        $10.2$ &        $12.1$ &        $13.7$ &        $16.0$ &        $16.4$ &        $11.3$ &         $8.8$ &        $4.2$ &       $10.3$ &                $1.2$ &           $3.4$ &        $7.2$ \\
\bottomrule
\end{tabular}
\end{adjustbox}
\begin{adjustbox}{width=1\textwidth}
\begin{tabular}{ll|llllp{3.5em}p{3.5em}p{3.5em}ll}
\toprule
       &          &         Platt &                 beta &      isotonic &            KDE & Spline natural &  Spline cubic & Spline parabolic &            $IOP_{OI}$ &    $IOP_{OP}$ \\
Model & {} &               &                      &               &                &                &               &                  &                       &               \\
\midrule
resnet110 & cars vs rest &   $4.32_{13}$ &  $\mathbf{3.07_{1}}$ &    $3.85_{8}$ &  $101.32_{21}$ &    $4.72_{18}$ &   $4.71_{17}$ &      $4.74_{19}$ &           $4.08_{12}$ &  $60.72_{20}$ \\
       & cats vs rest &  $11.55_{19}$ &          $10.4_{11}$ &   $10.11_{7}$ &   $37.36_{20}$ &   $11.24_{16}$ &  $11.41_{18}$ &      $11.3_{17}$ &           $10.11_{8}$ &  $78.25_{21}$ \\
       & dogs vs rest &    $8.16_{6}$ &           $8.05_{5}$ &   $9.68_{14}$ &   $27.18_{20}$ &    $10.0_{16}$ &  $10.35_{18}$ &      $10.3_{17}$ &            $7.13_{3}$ &  $78.65_{21}$ \\
       & confidence &   $11.57_{5}$ &          $10.36_{2}$ &  $15.01_{16}$ &   $23.39_{20}$ &   $14.66_{14}$ &  $14.84_{15}$ &     $14.24_{12}$ &          $13.23_{10}$ &   $80.5_{21}$ \\
              \midrule

densenet40 & cars vs rest &    $3.69_{6}$ &           $3.13_{2}$ &     $3.9_{8}$ &  $125.76_{21}$ &    $4.52_{11}$ &    $4.5_{10}$ &      $4.58_{13}$ &            $3.56_{5}$ &  $53.75_{20}$ \\
       & cats vs rest &    $8.88_{8}$ &           $8.31_{6}$ &   $9.57_{10}$ &   $74.05_{21}$ &    $10.6_{14}$ &  $11.05_{16}$ &      $10.9_{15}$ &            $7.71_{5}$ &  $68.08_{20}$ \\
       & dogs vs rest &    $7.15_{7}$ &           $6.66_{4}$ &   $8.45_{12}$ &   $63.42_{20}$ &    $8.69_{13}$ &   $9.52_{16}$ &      $9.03_{15}$ &            $5.86_{2}$ &   $66.8_{21}$ \\
       & confidence &   $11.99_{8}$ &          $11.31_{5}$ &  $14.67_{15}$ &   $21.93_{20}$ &   $13.87_{12}$ &   $15.2_{16}$ &     $14.11_{14}$ &  $\mathbf{10.56_{1}}$ &  $73.02_{21}$ \\
              \midrule

wide32 & cars vs rest &   $4.63_{15}$ &            $3.2_{2}$ &   $3.95_{13}$ &  $170.14_{21}$ &    $5.29_{18}$ &   $5.17_{17}$ &      $5.35_{19}$ &           $3.82_{10}$ &  $54.52_{20}$ \\
       & cats vs rest &   $9.03_{12}$ &           $8.36_{5}$ &   $9.14_{13}$ &   $42.39_{20}$ &      $8.5_{6}$ &    $8.73_{9}$ &       $8.63_{8}$ &            $8.02_{4}$ &  $73.09_{21}$ \\
       & dogs vs rest &    $8.11_{5}$ &           $7.53_{4}$ &     $8.8_{9}$ &   $41.51_{20}$ &    $9.19_{11}$ &    $9.4_{13}$ &      $9.72_{16}$ &             $6.9_{3}$ &  $74.03_{21}$ \\
       & confidence &  $11.56_{10}$ &          $10.85_{8}$ &  $13.33_{12}$ &   $26.38_{20}$ &   $13.33_{13}$ &  $14.41_{16}$ &     $13.58_{14}$ &          $12.69_{11}$ &  $82.83_{21}$ \\
\hline
       & avg rank &         $9.5$ &                $4.6$ &        $11.4$ &         $20.3$ &         $13.5$ &        $15.1$ &           $14.9$ &                 $6.2$ &        $20.7$ \\
\bottomrule
\end{tabular}
\end{adjustbox}
\end{table}
\begin{table}
\caption{$PL$-methods, binning methods and various other calibration methods used for calibration. Evaluated on CIFAR-5m. The table displays thousandths of $\vert\ch_\mathcal{M}-c^*\vert$. The value of $c\inserted{^*}$ was found by estimating the true calibration map on $10^6$ unseen data with isotonic regression. Table is shown as two lines for better readability.}
\label{table:chat_dist_c_sq}
\centering

\begin{adjustbox}{width=1\textwidth}
\begin{tabular}{ll|lllllll|lllll}
\toprule
       & binning &    $ES_{10}$ &    $ES_{15}$ &    $ES_{20}$ &    $ES_{25}$ &    $ES_{30}$ & $ES_{sweep}$ &    $ES_{CV}$ &           $PL3^{CE}$ &  $PL3^{MSE}$ &       $PL_{NN}^{CE}$ & $PL_{NN}^{MSE}$ &    $PL_{DE}$ \\
Model & {} &              &              &              &              &              &              &              &                      &              &                      &                 &              \\
\midrule
resnet110 & cars vs rest &   $0.18_{9}$ &   $0.15_{2}$ &   $0.16_{5}$ &   $0.17_{6}$ &   $0.17_{7}$ &  $0.22_{13}$ &   $0.18_{8}$ &  $\mathbf{0.14_{1}}$ &   $0.2_{12}$ &           $0.16_{4}$ &      $0.16_{3}$ &  $0.19_{10}$ \\
       & cats vs rest &   $0.37_{6}$ &   $0.52_{9}$ &    $0.5_{8}$ &  $0.58_{14}$ &  $0.57_{13}$ &  $0.54_{11}$ &    $0.5_{7}$ &           $0.33_{3}$ &   $0.34_{4}$ &  $\mathbf{0.29_{1}}$ &      $0.31_{2}$ &  $0.52_{10}$ \\
       & dogs vs rest &   $0.32_{4}$ &   $0.4_{10}$ &  $0.43_{11}$ &  $0.51_{13}$ &  $0.58_{15}$ &   $0.34_{7}$ &  $0.47_{12}$ &           $0.29_{3}$ &   $0.38_{9}$ &           $0.21_{2}$ &      $0.34_{6}$ &  $0.52_{14}$ \\
       & confidence &   $0.45_{7}$ &  $0.61_{11}$ &  $0.73_{14}$ &  $0.99_{18}$ &  $1.13_{19}$ &   $0.48_{9}$ &  $0.66_{12}$ &           $0.38_{5}$ &   $0.36_{4}$ &  $\mathbf{0.27_{1}}$ &       $0.3_{2}$ &  $0.53_{10}$ \\
       \hline
densenet40 & cars vs rest &  $0.32_{14}$ &   $0.16_{2}$ &   $0.19_{8}$ &   $0.17_{4}$ &   $0.2_{10}$ &  $0.31_{13}$ &   $0.19_{7}$ &           $0.18_{6}$ &  $0.25_{11}$ &   $\mathbf{0.1_{1}}$ &      $0.16_{3}$ &  $0.41_{15}$ \\
       & cats vs rest &   $0.34_{6}$ &  $0.48_{13}$ &   $0.34_{7}$ &  $0.49_{14}$ &  $0.58_{15}$ &   $0.4_{11}$ &   $0.34_{8}$ &           $0.24_{3}$ &   $0.29_{5}$ &  $\mathbf{0.17_{1}}$ &       $0.2_{2}$ &  $0.37_{10}$ \\
       & dogs vs rest &   $0.23_{4}$ &  $0.34_{11}$ &   $0.4_{13}$ &  $0.48_{16}$ &  $0.52_{18}$ &   $0.28_{6}$ &  $0.36_{12}$ &           $0.29_{7}$ &   $0.32_{9}$ &           $0.15_{2}$ &      $0.18_{3}$ &  $0.33_{10}$ \\
       & confidence &  $0.52_{10}$ &  $0.64_{12}$ &  $0.91_{16}$ &   $1.0_{18}$ &  $1.19_{19}$ &   $0.44_{4}$ &  $0.55_{11}$ &           $0.49_{7}$ &    $0.5_{8}$ &           $0.42_{3}$ &      $0.42_{2}$ &   $0.51_{9}$ \\
       \hline
wide32 & cars vs rest &   $0.14_{4}$ &   $0.12_{2}$ &   $0.16_{7}$ &   $0.15_{6}$ &   $0.18_{8}$ &  $0.22_{12}$ &   $0.15_{5}$ &          $0.22_{13}$ &  $0.29_{14}$ &   $\mathbf{0.1_{1}}$ &      $0.12_{3}$ &   $0.2_{10}$ \\
       & cats vs rest &  $0.39_{11}$ &   $0.35_{7}$ &  $0.41_{15}$ &  $0.49_{18}$ &  $0.48_{17}$ &  $0.39_{13}$ &   $0.34_{6}$ &           $0.27_{3}$ &   $0.33_{5}$ &  $\mathbf{0.24_{1}}$ &      $0.25_{2}$ &  $0.43_{16}$ \\
       & dogs vs rest &   $0.33_{5}$ &   $0.36_{7}$ &   $0.39_{8}$ &  $0.57_{17}$ &  $0.48_{15}$ &  $0.39_{10}$ &  $0.42_{11}$ &           $0.39_{9}$ &  $0.44_{13}$ &  $\mathbf{0.21_{1}}$ &      $0.26_{3}$ &  $0.75_{19}$ \\
       & confidence &  $0.37_{10}$ &  $0.56_{14}$ &  $0.71_{17}$ &  $0.89_{18}$ &  $1.03_{19}$ &    $0.3_{3}$ &  $0.38_{11}$ &           $0.31_{6}$ &   $0.31_{5}$ &  $\mathbf{0.26_{1}}$ &      $0.27_{2}$ &   $0.31_{4}$ \\
\hline
       & avg rank &        $7.5$ &        $8.3$ &       $10.8$ &       $13.5$ &       $14.6$ &        $9.3$ &        $9.2$ &                $5.5$ &        $8.2$ &                $1.6$ &           $2.8$ &       $11.4$ \\
\bottomrule
\end{tabular}
\end{adjustbox}
\begin{adjustbox}{width=1\textwidth}
\begin{tabular}{ll|llllp{3.5em}p{3.5em}p{3.5em}ll}
\toprule
       &          &        Platt &         beta &     isotonic &            KDE & Spline natural & Spline cubic & Spline parabolic &           $IOP_{OI}$ &    $IOP_{OP}$ \\
Model & {} &              &              &              &                &                &              &                  &                      &               \\
\midrule
resnet110 & cars vs rest &  $0.31_{15}$ &   $0.2_{11}$ &  $0.33_{16}$ &    $90.6_{21}$ &    $0.53_{19}$ &  $0.53_{17}$ &      $0.53_{18}$ &          $0.23_{14}$ &  $12.49_{20}$ \\
       & cats vs rest &  $0.56_{12}$ &  $0.62_{15}$ &  $0.64_{16}$ &   $17.57_{20}$ &    $0.76_{18}$ &  $0.78_{19}$ &      $0.76_{17}$ &           $0.35_{5}$ &  $18.45_{21}$ \\
       & dogs vs rest &   $0.32_{5}$ &   $0.37_{8}$ &  $0.66_{18}$ &   $11.51_{20}$ &    $0.59_{16}$ &  $0.67_{19}$ &      $0.61_{17}$ &  $\mathbf{0.21_{1}}$ &  $19.05_{21}$ \\
       & confidence &   $0.38_{6}$ &   $0.32_{3}$ &  $0.85_{17}$ &    $3.27_{20}$ &    $0.75_{16}$ &  $0.74_{15}$ &      $0.68_{13}$ &           $0.47_{8}$ &  $18.62_{21}$ \\
       \hline
densenet40 & cars vs rest &    $0.2_{9}$ &  $0.25_{12}$ &  $0.41_{16}$ &  $115.55_{21}$ &    $0.57_{17}$ &  $0.57_{19}$ &      $0.57_{18}$ &           $0.17_{5}$ &  $10.67_{20}$ \\
       & cats vs rest &  $0.43_{12}$ &   $0.36_{9}$ &  $0.59_{16}$ &   $43.51_{21}$ &    $0.82_{17}$ &  $0.86_{19}$ &      $0.82_{18}$ &           $0.25_{4}$ &  $15.31_{20}$ \\
       & dogs vs rest &    $0.3_{8}$ &   $0.27_{5}$ &  $0.53_{19}$ &   $38.26_{21}$ &    $0.43_{14}$ &  $0.51_{17}$ &      $0.44_{15}$ &  $\mathbf{0.15_{1}}$ &  $15.02_{20}$ \\
       & confidence &   $0.47_{6}$ &   $0.46_{5}$ &  $0.94_{17}$ &     $3.1_{20}$ &    $0.69_{13}$ &   $0.8_{15}$ &       $0.7_{14}$ &   $\mathbf{0.3_{1}}$ &  $15.88_{21}$ \\
       \hline
wide32 & cars vs rest &  $0.31_{15}$ &  $0.22_{11}$ &  $0.38_{16}$ &  $157.82_{21}$ &    $1.05_{17}$ &  $1.05_{19}$ &      $1.05_{18}$ &           $0.18_{9}$ &  $11.26_{20}$ \\
       & cats vs rest &   $0.35_{8}$ &  $0.39_{12}$ &  $0.56_{19}$ &   $21.13_{21}$ &     $0.38_{9}$ &   $0.4_{14}$ &      $0.39_{10}$ &           $0.28_{4}$ &   $17.2_{20}$ \\
       & dogs vs rest &   $0.35_{6}$ &   $0.32_{4}$ &  $0.59_{18}$ &   $22.04_{21}$ &    $0.44_{12}$ &  $0.51_{16}$ &      $0.46_{14}$ &           $0.22_{2}$ &  $17.84_{20}$ \\
       & confidence &   $0.34_{8}$ &   $0.33_{7}$ &  $0.67_{16}$ &    $5.03_{20}$ &    $0.55_{13}$ &  $0.61_{15}$ &      $0.55_{12}$ &           $0.36_{9}$ &  $19.41_{21}$ \\
\hline
       & avg rank &        $9.2$ &        $8.5$ &       $17.0$ &         $20.6$ &         $15.1$ &       $17.0$ &           $15.3$ &                $5.2$ &        $20.4$ \\
\bottomrule
\end{tabular}
\end{adjustbox}

\end{table}

\begin{table}
\caption{Comparison of calibration evaluators in estimating the true calibration error \inserted{with }$\vert ECE_\mathcal{M}-CE\vert$ on CIFAR-5m. Evaluated on CIFAR-5m. The initial model predictions have been calibrated by 6 different methods on 5k data points. Corresponding methods (columns) have been then used to estimate the true calibration map on a new unseen test set of size 1k, 3k, 10k. Average of 5 test set data seeds. In total, a single cell value is the average of 6x3x5=90 calibration map estimates. The value of $c$ was found by estimating the true calibration map on $10^6$ unseen data with isotonic regression. The table displays thousandths. Table is shown as two lines for better readability.}
\label{table:ECE_abs}
\centering

\begin{adjustbox}{width=1\textwidth}
\begin{tabular}{ll|lllllll|lllll}
\toprule
       & binning &            $ES_{10}$ &            $ES_{15}$ &    $ES_{20}$ &            $ES_{25}$ &            $ES_{30}$ & $ES_{sweep}$ &    $ES_{CV}$ &           $PL3^{CE}$ &  $PL3^{MSE}$ & $PL_{NN}^{CE}$ &      $PL_{NN}^{MSE}$ &    $PL_{DE}$ \\
Model & {} &                      &                      &              &                      &                      &              &              &                      &              &                &                      &              \\
\midrule
resnet110 & cars vs rest &          $1.99_{14}$ &          $1.85_{10}$ &   $1.85_{9}$ &           $1.61_{4}$ &            $1.7_{5}$ &  $1.94_{13}$ &  $2.84_{19}$ &  $\mathbf{1.36_{1}}$ &  $2.15_{16}$ &    $2.03_{15}$ &           $1.71_{6}$ &    $1.8_{7}$ \\
       & cats vs rest &           $2.32_{2}$ &           $2.35_{3}$ &   $2.37_{4}$ &           $2.41_{5}$ &  $\mathbf{2.26_{1}}$ &  $4.01_{13}$ &   $3.16_{7}$ &           $2.58_{6}$ &  $4.08_{14}$ &     $3.6_{10}$ &           $3.17_{8}$ &   $3.29_{9}$ \\
       & dogs vs rest &           $3.29_{5}$ &  $\mathbf{2.83_{1}}$ &   $2.88_{2}$ &           $3.18_{4}$ &          $3.65_{11}$ &   $3.29_{6}$ &  $4.24_{15}$ &            $2.9_{3}$ &   $4.0_{14}$ &     $3.55_{8}$ &          $3.74_{12}$ &  $3.96_{13}$ \\
       & confidence &           $4.16_{3}$ &           $4.45_{8}$ &   $4.29_{5}$ &           $4.39_{6}$ &          $5.18_{13}$ &  $4.85_{12}$ &  $6.68_{16}$ &           $4.02_{2}$ &  $4.56_{10}$ &     $4.39_{7}$ &          $4.72_{11}$ &   $4.29_{4}$ \\
       \hline
densenet40 & cars vs rest &           $1.23_{4}$ &           $1.11_{2}$ &   $1.29_{6}$ &  $\mathbf{1.09_{1}}$ &           $1.11_{3}$ &  $1.52_{10}$ &  $2.31_{18}$ &           $1.38_{7}$ &  $2.62_{19}$ &     $1.49_{9}$ &           $1.45_{8}$ &  $2.14_{17}$ \\
\bottomrule
\end{tabular}
\end{adjustbox}
\begin{adjustbox}{width=1\textwidth}
\begin{tabular}{ll|llllp{3.5em}p{3.5em}p{3.5em}ll}
\toprule
       &          &        Platt &                 beta &     isotonic &            KDE & Spline natural & Spline cubic & Spline parabolic &   $IOP_{OI}$ &    $IOP_{OP}$ \\
Model & {} &              &                      &              &                &                &              &                  &              &               \\
\midrule
resnet110 & cars vs rest &  $2.34_{17}$ &            $1.4_{2}$ &   $1.52_{3}$ &   $98.37_{21}$ &    $1.87_{11}$ &   $1.85_{8}$ &      $1.92_{12}$ &  $2.49_{18}$ &  $57.26_{20}$ \\
       & cats vs rest &   $5.5_{18}$ &          $3.69_{11}$ &  $3.97_{12}$ &   $29.28_{20}$ &    $5.29_{16}$ &  $5.11_{15}$ &      $5.38_{17}$ &  $5.78_{19}$ &  $73.24_{21}$ \\
       & dogs vs rest &  $3.62_{10}$ &           $3.31_{7}$ &   $5.0_{18}$ &   $20.24_{20}$ &    $4.71_{16}$ &  $4.88_{17}$ &      $5.07_{19}$ &   $3.62_{9}$ &  $75.03_{21}$ \\
       & confidence &   $4.48_{9}$ &  $\mathbf{3.03_{1}}$ &  $6.03_{14}$ &   $13.91_{20}$ &    $6.69_{17}$ &  $6.78_{18}$ &      $6.17_{15}$ &  $7.46_{19}$ &  $75.26_{21}$ \\
       \hline
densenet40 & cars vs rest &  $1.86_{14}$ &           $1.25_{5}$ &  $1.98_{16}$ &  $123.28_{21}$ &    $1.85_{13}$ &  $1.83_{12}$ &      $1.94_{15}$ &   $1.7_{11}$ &  $51.07_{20}$ \\
       & cats vs rest &  $5.14_{15}$ &          $4.49_{12}$ &  $5.27_{16}$ &   $68.41_{21}$ &    $7.14_{17}$ &  $7.69_{19}$ &      $7.45_{18}$ &  $4.08_{10}$ &  $64.47_{20}$ \\
\bottomrule
\end{tabular}
\end{adjustbox}

\end{table}
\begin{table}
\caption{Comparison of calibration evaluators in estimating the true calibration error \inserted{with }$\vert ECE_\mathcal{M}-CE\vert^2$ on CIFAR-5m. Otherwise like Table~\ref{table:ECE_abs}. Table is shown as two lines for better readability.}
\label{table:ECE_sq}
\centering

\begin{adjustbox}{width=1\textwidth}
\begin{tabular}{ll|lllllll|lllll}
\toprule
       & binning &    $ES_{10}$ &    $ES_{15}$ &    $ES_{20}$ &    $ES_{25}$ &    $ES_{30}$ & $ES_{sweep}$ &    $ES_{CV}$ &           $PL3^{CE}$ &          $PL3^{MSE}$ &       $PL_{NN}^{CE}$ &      $PL_{NN}^{MSE}$ &    $PL_{DE}$ \\
Model & {} &              &              &              &              &              &              &              &                      &                      &                      &                      &              \\
\midrule
resnet110 & cars vs rest &    $0.2_{6}$ &   $0.21_{8}$ &  $0.22_{12}$ &   $0.18_{2}$ &  $0.24_{13}$ &  $0.28_{15}$ &  $0.22_{10}$ &           $0.18_{3}$ &          $0.22_{11}$ &           $0.18_{4}$ &  $\mathbf{0.17_{1}}$ &   $0.22_{9}$ \\
       & cats vs rest &   $0.25_{3}$ &   $0.24_{2}$ &   $0.27_{5}$ &   $0.26_{4}$ &   $0.28_{7}$ &  $0.38_{12}$ &   $0.28_{6}$ &  $\mathbf{0.24_{1}}$ &            $0.3_{9}$ &           $0.29_{8}$ &          $0.31_{10}$ &  $0.41_{14}$ \\
       & dogs vs rest &   $0.21_{5}$ &   $0.19_{3}$ &  $0.28_{10}$ &  $0.29_{12}$ &  $0.29_{11}$ &   $0.23_{8}$ &  $0.29_{13}$ &           $0.22_{7}$ &          $0.31_{14}$ &           $0.17_{2}$ &           $0.27_{9}$ &  $0.46_{15}$ \\
       & confidence &   $0.38_{7}$ &   $0.41_{8}$ &  $0.44_{11}$ &  $0.47_{13}$ &  $0.52_{15}$ &  $0.44_{12}$ &  $0.43_{10}$ &           $0.36_{6}$ &           $0.36_{5}$ &           $0.28_{2}$ &           $0.32_{3}$ &  $0.49_{14}$ \\
              \hline

densenet40 & cars vs rest &   $0.17_{9}$ &   $0.16_{6}$ &   $0.16_{7}$ &   $0.16_{5}$ &   $0.17_{8}$ &  $0.26_{14}$ &  $0.18_{11}$ &          $0.17_{10}$ &          $0.24_{13}$ &  $\mathbf{0.11_{1}}$ &           $0.15_{4}$ &  $0.45_{19}$ \\
       & cats vs rest &   $0.21_{6}$ &   $0.21_{5}$ &   $0.23_{7}$ &  $0.25_{10}$ &   $0.23_{8}$ &  $0.27_{11}$ &    $0.2_{3}$ &           $0.24_{9}$ &           $0.3_{12}$ &  $\mathbf{0.16_{1}}$ &            $0.2_{4}$ &  $0.36_{14}$ \\
\bottomrule
\end{tabular}
\end{adjustbox}
\begin{adjustbox}{width=1\textwidth}
\begin{tabular}{ll|llllp{3.5em}p{3.5em}p{3.5em}ll}
\toprule
       &          &        Platt &                 beta &     isotonic &            KDE & Spline natural & Spline cubic & Spline parabolic &           $IOP_{OI}$ &    $IOP_{OP}$ \\
Model & {} &              &                      &              &                &                &              &                  &                      &               \\
\midrule
resnet110 & cars vs rest &  $0.24_{14}$ &           $0.21_{7}$ &  $0.32_{16}$ &    $90.6_{21}$ &    $0.34_{18}$ &  $0.34_{19}$ &      $0.34_{17}$ &            $0.2_{5}$ &  $12.29_{20}$ \\
       & cats vs rest &  $0.47_{15}$ &          $0.33_{11}$ &  $0.59_{16}$ &   $17.58_{20}$ &    $0.81_{17}$ &  $0.82_{19}$ &      $0.82_{18}$ &          $0.41_{13}$ &  $19.14_{21}$ \\
       & dogs vs rest &    $0.2_{4}$ &           $0.21_{6}$ &  $0.63_{19}$ &   $11.38_{20}$ &    $0.52_{16}$ &  $0.58_{18}$ &      $0.53_{17}$ &  $\mathbf{0.16_{1}}$ &  $19.38_{21}$ \\
       & confidence &   $0.32_{4}$ &  $\mathbf{0.24_{1}}$ &  $0.64_{19}$ &    $3.08_{20}$ &    $0.56_{17}$ &  $0.59_{18}$ &      $0.52_{16}$ &           $0.42_{9}$ &  $19.01_{21}$ \\
       \hline
densenet40 & cars vs rest &   $0.14_{3}$ &          $0.22_{12}$ &  $0.42_{18}$ &  $115.57_{21}$ &    $0.31_{15}$ &  $0.31_{16}$ &      $0.31_{17}$ &           $0.14_{2}$ &  $10.57_{20}$ \\
       & cats vs rest &  $0.42_{15}$ &          $0.32_{13}$ &  $0.54_{16}$ &   $43.47_{21}$ &    $0.99_{17}$ &  $1.04_{19}$ &       $1.0_{18}$ &            $0.2_{2}$ &  $15.43_{20}$ \\
\bottomrule
\end{tabular}
\end{adjustbox}

\end{table}

\begin{table}
\caption{Comparison of calibration evaluators in estimating the reliability diagram with $\vert\ch_\mathcal{M}-c^*\vert$ on CIFAR-5m. The value of $c\inserted{^*}$ was found by estimating the true calibration map on $10^6$ unseen data with equal size binning with 100 bins with slope 1 instead of isotonic regression. Otherwise like Table~\ref{table:chat_dist_c}. Table is shown as two lines for better readability.}
\label{table:chat_dist_c_cgt0}
\centering

\begin{adjustbox}{width=1\textwidth}
\begin{tabular}{ll|lllllll|lllll}
\toprule
       & binning &     $ES_{10}$ &     $ES_{15}$ &     $ES_{20}$ &     $ES_{25}$ &     $ES_{30}$ &  $ES_{sweep}$ &     $ES_{CV}$ &   $PL3^{CE}$ &   $PL3^{MSE}$ &       $PL_{NN}^{CE}$ & $PL_{NN}^{MSE}$ &     $PL_{DE}$ \\
Model & {} &               &               &               &               &               &               &               &              &               &                      &                 &               \\
\midrule
resnet110 & cars vs rest &   $4.42_{14}$ &    $3.76_{6}$ &    $3.67_{4}$ &    $3.98_{8}$ &    $3.75_{5}$ &   $4.63_{15}$ &    $3.99_{9}$ &   $3.66_{3}$ &    $4.7_{16}$ &            $3.4_{2}$ &      $3.81_{7}$ &   $4.08_{10}$ \\
       & cats vs rest &    $8.78_{4}$ &   $10.26_{8}$ &  $10.32_{10}$ &  $10.74_{13}$ &  $10.77_{14}$ &  $11.03_{15}$ &   $10.13_{6}$ &   $7.98_{2}$ &    $9.75_{5}$ &  $\mathbf{7.75_{1}}$ &       $8.6_{3}$ &    $10.3_{9}$ \\
       & dogs vs rest &    $8.13_{7}$ &   $9.13_{12}$ &   $9.14_{13}$ &    $9.7_{14}$ &  $10.95_{19}$ &    $8.49_{8}$ &   $9.02_{11}$ &   $7.01_{3}$ &    $8.92_{9}$ &  $\mathbf{6.39_{1}}$ &      $7.67_{4}$ &   $8.95_{10}$ \\
       & confidence &    $12.4_{9}$ &  $14.02_{13}$ &  $15.42_{17}$ &  $17.35_{18}$ &   $18.9_{19}$ &   $11.61_{7}$ &   $13.6_{11}$ &  $10.84_{4}$ &   $11.41_{6}$ &  $\mathbf{9.54_{1}}$ &     $10.54_{3}$ &   $11.61_{8}$ \\
\hline
densenet40 & cars vs rest &   $7.28_{19}$ &    $4.77_{9}$ &    $5.0_{14}$ &    $5.0_{13}$ &    $4.8_{10}$ &   $5.16_{15}$ &    $4.42_{7}$ &   $3.86_{4}$ &   $4.89_{12}$ &  $\mathbf{3.23_{1}}$ &      $3.82_{3}$ &    $4.73_{8}$ \\
       & cats vs rest &    $8.73_{7}$ &  $10.88_{14}$ &   $9.43_{11}$ &  $10.36_{13}$ &  $11.58_{18}$ &   $9.91_{12}$ &    $8.92_{8}$ &   $7.67_{3}$ &    $9.09_{9}$ &  $\mathbf{6.69_{1}}$ &      $7.31_{2}$ &    $8.38_{5}$ \\
\bottomrule
\end{tabular}
\end{adjustbox}
\begin{adjustbox}{width=1\textwidth}
\begin{tabular}{ll|llllp{3.5em}p{3.5em}p{3.5em}ll}
\toprule
       &          &         Platt &                 beta &      isotonic &            KDE & Spline natural &  Spline cubic & Spline parabolic &            $IOP_{OI}$ &    $IOP_{OP}$ \\
Model & {} &               &                      &               &                &                &               &                  &                       &               \\
\midrule
resnet110 & cars vs rest &   $4.24_{13}$ &  $\mathbf{3.23_{1}}$ &   $4.15_{11}$ &  $101.42_{21}$ &    $4.89_{17}$ &   $4.89_{18}$ &       $4.9_{19}$ &           $4.18_{12}$ &  $60.53_{20}$ \\
       & cats vs rest &  $11.46_{17}$ &         $10.42_{11}$ &  $10.52_{12}$ &   $37.34_{20}$ &   $11.42_{16}$ &  $11.56_{19}$ &     $11.54_{18}$ &           $10.15_{7}$ &  $78.04_{21}$ \\
       & dogs vs rest &    $7.93_{6}$ &            $7.9_{5}$ &   $9.98_{15}$ &   $27.16_{20}$ &   $10.24_{16}$ &  $10.59_{18}$ &     $10.55_{17}$ &            $6.87_{2}$ &  $78.32_{21}$ \\
       & confidence &   $11.24_{5}$ &          $10.15_{2}$ &  $15.09_{16}$ &   $23.08_{20}$ &   $14.42_{14}$ &  $14.57_{15}$ &      $14.0_{12}$ &          $12.82_{10}$ &  $80.23_{21}$ \\
\hline
densenet40 & cars vs rest &    $3.93_{6}$ &           $3.78_{2}$ &   $4.87_{11}$ &  $126.62_{21}$ &    $5.45_{17}$ &   $5.43_{16}$ &      $5.52_{18}$ &            $3.91_{5}$ &  $53.47_{20}$ \\
       & cats vs rest &   $9.09_{10}$ &           $8.66_{6}$ &  $11.04_{15}$ &   $74.37_{21}$ &   $11.32_{16}$ &  $11.76_{19}$ &     $11.53_{17}$ &            $7.74_{4}$ &   $67.6_{20}$ \\
\bottomrule
\end{tabular}
\end{adjustbox}

\end{table}
\begin{table}
\caption{Comparison of calibration evaluators in estimating the reliability diagram with $\vert\ch_\mathcal{M}-c^*\vert$ on CIFAR-5m. The value of $c\inserted{^*}$ was found by estimating the true calibration map on $10^6$ unseen data with equal size binning with 100 bins with flat bin tops instead of isotonic regression. Otherwise like Table~\ref{table:chat_dist_c}. Table is shown as two lines for better readability.}
\label{table:chat_dist_c_cgt1}
\centering

\begin{adjustbox}{width=1\textwidth}
\begin{tabular}{ll|lllllll|lllll}
\toprule
       & binning &     $ES_{10}$ &     $ES_{15}$ &     $ES_{20}$ &     $ES_{25}$ &     $ES_{30}$ &  $ES_{sweep}$ &     $ES_{CV}$ &    $PL3^{CE}$ &   $PL3^{MSE}$ &        $PL_{NN}^{CE}$ & $PL_{NN}^{MSE}$ &     $PL_{DE}$ \\
Model & {} &               &               &               &               &               &               &               &               &               &                       &                 &               \\
\midrule
resnet110 & cars vs rest &   $6.78_{15}$ &    $6.17_{8}$ &    $5.95_{4}$ &   $6.41_{13}$ &    $5.82_{3}$ &   $6.78_{14}$ &    $6.3_{10}$ &    $5.99_{5}$ &   $6.96_{16}$ &            $5.72_{2}$ &      $6.11_{7}$ &   $6.33_{11}$ \\
       & cats vs rest &   $10.46_{4}$ &  $11.65_{10}$ &  $11.73_{11}$ &  $12.31_{15}$ &  $12.19_{13}$ &   $12.2_{14}$ &   $11.43_{8}$ &    $9.46_{2}$ &   $11.33_{6}$ &    $\mathbf{9.2_{1}}$ &      $9.96_{3}$ &   $11.45_{9}$ \\
       & dogs vs rest &   $10.19_{7}$ &  $11.04_{12}$ &  $11.06_{13}$ &  $11.57_{15}$ &  $12.69_{17}$ &   $10.46_{8}$ &  $10.97_{11}$ &    $9.06_{3}$ &  $10.91_{10}$ &    $\mathbf{8.6_{1}}$ &      $9.85_{4}$ &   $10.88_{9}$ \\
       & confidence &   $13.17_{9}$ &  $14.79_{12}$ &  $16.19_{15}$ &  $18.09_{18}$ &  $19.57_{19}$ &   $12.53_{8}$ &  $14.37_{11}$ &   $11.59_{4}$ &   $12.36_{6}$ &  $\mathbf{10.38_{1}}$ &     $11.47_{3}$ &   $12.47_{7}$ \\
\hline
densenet40 & cars vs rest &   $9.69_{19}$ &   $7.45_{14}$ &   $7.39_{13}$ &   $7.53_{15}$ &   $7.22_{11}$ &     $7.1_{9}$ &    $6.64_{7}$ &    $6.35_{6}$ &   $7.26_{12}$ &   $\mathbf{6.02_{1}}$ &      $6.34_{5}$ &   $7.18_{10}$ \\
       & cats vs rest &   $11.21_{7}$ &  $13.24_{15}$ &  $11.49_{11}$ &   $12.3_{12}$ &  $13.43_{16}$ &  $12.33_{13}$ &   $11.25_{8}$ &    $10.3_{4}$ &   $11.33_{9}$ &   $\mathbf{9.53_{1}}$ &      $10.0_{3}$ &   $10.88_{6}$ \\
\bottomrule
\end{tabular}
\end{adjustbox}
\begin{adjustbox}{width=1\textwidth}
\begin{tabular}{ll|llllp{3.5em}p{3.5em}p{3.5em}ll}
\toprule
       &          &         Platt &                 beta &      isotonic &            KDE & Spline natural &  Spline cubic & Spline parabolic &           $IOP_{OI}$ &    $IOP_{OP}$ \\
Model & {} &               &                      &               &                &                &               &                  &                      &               \\
\midrule
resnet110 & cars vs rest &    $6.25_{9}$ &  $\mathbf{5.47_{1}}$ &    $6.08_{6}$ &  $103.08_{21}$ &    $7.36_{17}$ &   $7.36_{18}$ &      $7.38_{19}$ &          $6.38_{12}$ &  $61.42_{20}$ \\
       & cats vs rest &  $13.06_{16}$ &          $11.42_{7}$ &  $11.81_{12}$ &   $38.79_{20}$ &   $14.12_{17}$ &  $14.27_{19}$ &     $14.18_{18}$ &          $10.73_{5}$ &  $78.57_{21}$ \\
       & dogs vs rest &    $9.98_{6}$ &           $9.92_{5}$ &  $11.37_{14}$ &   $28.65_{20}$ &   $12.58_{16}$ &  $12.92_{19}$ &     $12.87_{18}$ &           $9.01_{2}$ &  $79.46_{21}$ \\
       & confidence &   $11.91_{5}$ &          $10.97_{2}$ &  $15.77_{13}$ &   $23.49_{20}$ &    $16.2_{16}$ &  $16.31_{17}$ &     $15.81_{14}$ &         $14.35_{10}$ &  $81.67_{21}$ \\
\hline
densenet40 & cars vs rest &    $6.25_{3}$ &           $6.15_{2}$ &    $7.04_{8}$ &  $127.57_{21}$ &    $8.17_{16}$ &   $8.23_{18}$ &      $8.18_{17}$ &           $6.34_{4}$ &  $54.43_{20}$ \\
       & cats vs rest &  $11.42_{10}$ &          $10.82_{5}$ &  $12.97_{14}$ &   $75.79_{21}$ &    $14.7_{17}$ &  $15.33_{19}$ &     $14.88_{18}$ &           $9.54_{2}$ &  $68.07_{20}$ \\
\bottomrule
\end{tabular}
\end{adjustbox}

\end{table}

\begin{table}
\caption{Comparison of calibration evaluators in estimating the true calibration error \inserted{with }$\vert ECE_\mathcal{M}-CE\vert$ (measured in thousandths) and the ranking of 6 calibrators (Spearman $rankcorrel(ECE)$) on CIFAR-5m. The value of $c$ was found by estimating the true calibration map on $10^6$ unseen data with equal size binning with 100 bins with slope 1 instead of isotonic regression.}
\label{table:cifar5m_ECE_ranking_cgt0}
\centering
\begin{adjustbox}{width=1\textwidth}
\begin{tabular}{ll|llllllllp{4em}l}
\toprule
           &  &            $ES_{15}$ & $ES_{sweep}$ &    $ES_{CV}$ &                $PL3$ &    $PL$ &      Platt &         beta &              isotonic & Spline natural & $IOP_{OI}$ \\
Metric & Data &                      &              &              &                      &              &              &              &              &         &              \\
\midrule
$\vert ECE_\mathcal{M}-CE\vert$ & one-vs-rest &  $\mathbf{2.79_{1}}$ &    $3.7_{5}$ &   $4.05_{8}$ &   $3.04_{2}$ &            $3.37_{4}$ &    $4.02_{7}$ &           $3.34_{3}$ &            $4.05_{9}$ &    $4.34_{10}$ &      $4.0_{6}$ \\
         & confidence &           $4.53_{2}$ &   $4.98_{4}$ &   $6.35_{7}$ &   $4.74_{3}$ &            $5.19_{5}$ &    $5.68_{6}$ &  $\mathbf{4.32_{1}}$ &            $7.51_{9}$ &    $7.63_{10}$ &     $6.76_{8}$ \\
 \midrule
$rankcorrel(ECE)$ & one-vs-rest &          $0.363_{5}$ &  $0.148_{7}$ &  $0.382_{4}$ &  $0.254_{6}$ &  $\mathbf{0.471_{1}}$ &   $0.058_{9}$ &          $0.124_{8}$ &           $0.466_{2}$ &    $0.457_{3}$ &  $-0.006_{10}$ \\
         & confidence &          $0.594_{2}$ &  $0.453_{4}$ &   $0.57_{3}$ &  $0.343_{7}$ &           $0.393_{6}$ &  $-0.026_{9}$ &          $0.432_{5}$ &  $\mathbf{0.622_{1}}$ &    $0.232_{8}$ &  $-0.084_{10}$ \\
\bottomrule
\end{tabular}
\end{adjustbox}
\end{table}
\begin{table}
\caption{Comparison of calibration evaluators in estimating the true calibration error \inserted{with }$\vert ECE_\mathcal{M}-CE\vert$ (measured in thousandths) and the ranking of 6 calibrators (Spearman $rankcorrel(ECE)$) on CIFAR-5m. The value of $c$ was found by estimating the true calibration map on $10^6$ unseen data with equal size binning with 100 bins with flat bin tops instead of isotonic regression.}
    \label{table:cifar5m_ECE_ranking_cgt1}
    \centering
    \begin{adjustbox}{width=1\textwidth}
\begin{tabular}{ll|llllllllp{4em}l}
\toprule
           &  &            $ES_{15}$ & $ES_{sweep}$ &    $ES_{CV}$ &                $PL3$ &    $PL$ &      Platt &         beta &              isotonic & Spline natural & $IOP_{OI}$ \\
Metric & Data &                      &              &              &                      &              &              &              &              &         &              \\
\midrule
$\vert ECE_\mathcal{M}-CE\vert$ & one-vs-rest &           $3.98_{4}$ &    $5.0_{8}$ &   $5.01_{9}$ &  $\mathbf{3.69_{1}}$ &      $4.7_{7}$ &    $4.69_{6}$ &   $4.18_{5}$ &   $3.74_{2}$ &             $3.8_{3}$ &    $5.55_{10}$ \\
         & confidence &  $\mathbf{4.83_{1}}$ &   $6.07_{5}$ &    $6.9_{9}$ &           $5.25_{3}$ &     $5.87_{4}$ &    $6.79_{8}$ &    $4.9_{2}$ &   $6.46_{7}$ &            $6.12_{6}$ &    $8.19_{10}$ \\
\midrule
$rankcorrel(ECE)$ & one-vs-rest &          $0.194_{4}$ &  $0.01_{10}$ &  $0.144_{6}$ &          $0.126_{8}$ &    $0.345_{2}$ &   $0.157_{5}$ &  $0.134_{7}$ &  $0.226_{3}$ &  $\mathbf{0.548_{1}}$ &    $0.088_{9}$ \\
         & confidence &          $0.478_{2}$ &  $0.261_{7}$ &  $0.396_{4}$ &          $0.241_{8}$ &     $0.39_{5}$ &  $-0.084_{9}$ &  $0.308_{6}$ &  $0.477_{3}$ &  $\mathbf{0.575_{1}}$ &  $-0.228_{10}$ \\
\bottomrule
\end{tabular}
\end{adjustbox}
\end{table}

\begin{table}
\caption{Comparison of calibration evaluators in estimating the reliability diagram with $\vert\ch_\mathcal{M}-c^*\vert$ on CIFAR-5m. Aggregate over different data sizes. In total, a single cell value is the average of 6x4x5=120 calibration map estimates, where 4 is number of 2 class problems (1vsRest and confidence). Otherwise like Table~\ref{table:chat_dist_c}. Table is shown as two lines for better readability.}
\label{table:chat_dist_c_ndata}
\centering

\begin{adjustbox}{width=1\textwidth}
\begin{tabular}{ll|lllllll|lllll}
\toprule
       & binning &     $ES_{10}$ &     $ES_{15}$ &     $ES_{20}$ &     $ES_{25}$ &     $ES_{30}$ & $ES_{sweep}$ &     $ES_{CV}$ &   $PL3^{CE}$ &   $PL3^{MSE}$ &       $PL_{NN}^{CE}$ & $PL_{NN}^{MSE}$ &    $PL_{DE}$ \\
Model & {} &               &               &               &               &               &              &               &              &               &                      &                 &              \\
\midrule
resnet110 & 1000 &   $11.98_{9}$ &  $13.29_{12}$ &  $14.16_{17}$ &  $16.04_{18}$ &  $17.05_{19}$ &  $11.93_{8}$ &  $13.04_{11}$ &   $9.64_{3}$ &  $12.13_{10}$ &           $8.61_{2}$ &     $10.12_{4}$ &  $11.66_{7}$ \\
       & 3000 &    $7.83_{5}$ &   $8.49_{12}$ &   $9.09_{14}$ &   $9.33_{15}$ &   $9.89_{19}$ &   $8.16_{7}$ &   $8.32_{10}$ &   $6.89_{3}$ &    $8.16_{6}$ &   $\mathbf{6.7_{1}}$ &      $7.44_{4}$ &   $8.24_{9}$ \\
       & 10000 &    $5.91_{5}$ &   $6.51_{11}$ &    $6.08_{6}$ &   $6.37_{10}$ &   $6.69_{12}$ &  $7.02_{14}$ &     $6.3_{9}$ &   $5.47_{3}$ &    $5.83_{4}$ &  $\mathbf{4.78_{1}}$ &      $5.32_{2}$ &   $6.21_{7}$ \\
       \hline
densenet40 & 1000 &  $12.89_{14}$ &  $13.59_{16}$ &   $15.4_{17}$ &   $16.4_{18}$ &  $16.84_{19}$ &  $11.57_{9}$ &   $10.98_{6}$ &  $10.74_{5}$ &  $12.17_{12}$ &           $8.06_{2}$ &      $9.64_{4}$ &  $11.02_{7}$ \\
       & 3000 &   $8.84_{12}$ &   $9.22_{13}$ &   $9.52_{16}$ &   $10.5_{19}$ &  $11.44_{20}$ &    $7.6_{9}$ &   $8.54_{10}$ &   $6.21_{2}$ &    $7.49_{8}$ &  $\mathbf{5.71_{1}}$ &      $6.54_{4}$ &   $6.77_{6}$ \\
       & 10000 &   $6.52_{12}$ &   $7.14_{19}$ &   $6.22_{11}$ &   $6.75_{15}$ &   $7.33_{20}$ &  $6.59_{13}$ &     $5.6_{6}$ &   $4.29_{2}$ &    $4.92_{5}$ &  $\mathbf{4.17_{1}}$ &      $4.48_{3}$ &    $4.8_{4}$ \\
\bottomrule
\end{tabular}
\end{adjustbox}
\begin{adjustbox}{width=1\textwidth}
\begin{tabular}{ll|llllp{3.5em}p{3.5em}p{3.5em}ll}
\toprule
       &          &        Platt &         beta &      isotonic &           KDE & Spline natural &  Spline cubic & Spline parabolic &           $IOP_{OI}$ &     $IOP_{OP}$ \\
Model & {} &              &              &               &               &                &               &                  &                      &                \\
\midrule
resnet110 & 1000 &  $10.68_{6}$ &  $10.18_{5}$ &  $13.98_{16}$ &  $51.75_{20}$ &   $13.45_{14}$ &   $13.9_{15}$ &     $13.42_{13}$ &   $\mathbf{8.6_{1}}$ &  $205.17_{21}$ \\
       & 3000 &   $8.24_{8}$ &   $6.89_{2}$ &   $8.79_{13}$ &  $46.42_{21}$ &    $9.52_{18}$ &    $9.5_{17}$ &      $9.44_{16}$ &          $8.34_{11}$ &   $10.01_{20}$ \\
       & 10000 &  $7.79_{18}$ &  $6.84_{13}$ &    $6.22_{8}$ &  $43.76_{21}$ &    $7.49_{15}$ &   $7.58_{16}$ &      $7.58_{17}$ &          $8.98_{20}$ &    $8.41_{19}$ \\
\hline
densenet40 & 1000 &  $11.21_{8}$ &   $9.53_{3}$ &  $12.88_{13}$ &  $77.23_{20}$ &   $11.76_{10}$ &  $12.99_{15}$ &     $12.15_{11}$ &  $\mathbf{7.52_{1}}$ &  $180.88_{21}$ \\
       & 3000 &   $6.82_{7}$ &   $6.71_{5}$ &   $8.73_{11}$ &  $69.87_{21}$ &    $9.43_{15}$ &  $10.11_{18}$ &       $9.7_{17}$ &           $6.51_{3}$ &    $9.36_{14}$ \\
       & 10000 &   $5.76_{7}$ &   $5.82_{8}$ &    $5.83_{9}$ &  $66.78_{21}$ &    $7.07_{16}$ &   $7.11_{17}$ &      $7.11_{18}$ &          $6.74_{14}$ &    $6.01_{10}$ \\
\bottomrule
\end{tabular}
\end{adjustbox}

\end{table}
\begin{table}
\caption{Comparison of calibration evaluators in estimating the true calibration error \inserted{with }$\vert ECE_\mathcal{M}-CE\vert$ on CIFAR-5m. Aggregate over different data sizes. In total, a single cell value is the average of 6x4x5=120 calibration map estimates, where 4 is number of 2 class problems (1vsRest and confidence). Otherwise like Table~\ref{table:ECE_abs}. Table is shown as two lines for better readability.}
\label{table:ECE_abs_ndata}
\centering

\begin{adjustbox}{width=1\textwidth}
\begin{tabular}{ll|lllllll|lllll}
\toprule
       & binning &            $ES_{10}$ &            $ES_{15}$ &            $ES_{20}$ &            $ES_{25}$ &            $ES_{30}$ & $ES_{sweep}$ &    $ES_{CV}$ &   $PL3^{CE}$ &  $PL3^{MSE}$ & $PL_{NN}^{CE}$ & $PL_{NN}^{MSE}$ &    $PL_{DE}$ \\
Model & {} &                      &                      &                      &                      &                      &              &              &              &              &                &                 &              \\
\midrule
resnet110 & 1000 &           $4.39_{6}$ &           $4.09_{3}$ &           $4.63_{7}$ &          $4.85_{11}$ &           $5.5_{13}$ &   $4.34_{5}$ &  $6.93_{19}$ &   $3.79_{2}$ &  $5.58_{14}$ &    $5.15_{12}$ &     $4.84_{10}$ &   $4.69_{8}$ \\
       & 3000 &           $2.46_{3}$ &            $2.5_{4}$ &  $\mathbf{2.22_{1}}$ &           $2.45_{2}$ &           $2.53_{5}$ &   $3.7_{12}$ &  $3.77_{14}$ &   $2.63_{7}$ &  $3.51_{11}$ &    $3.16_{10}$ &      $3.14_{9}$ &   $3.03_{8}$ \\
       & 10000 &           $1.97_{7}$ &          $2.03_{11}$ &            $1.7_{3}$ &  $\mathbf{1.39_{1}}$ &           $1.57_{2}$ &  $2.53_{13}$ &   $1.99_{8}$ &   $1.73_{4}$ &    $2.0_{9}$ &     $1.87_{6}$ &     $2.02_{10}$ &  $2.29_{12}$ \\
       \hline
densenet40 & 1000 &           $5.09_{6}$ &           $4.62_{2}$ &           $5.66_{8}$ &          $5.91_{11}$ &          $6.56_{13}$ &   $4.78_{3}$ &  $6.98_{14}$ &  $5.76_{10}$ &  $7.56_{15}$ &     $5.24_{7}$ &      $4.78_{4}$ &   $5.66_{9}$ \\
       & 3000 &           $2.55_{2}$ &  $\mathbf{2.41_{1}}$ &           $2.55_{3}$ &           $2.75_{6}$ &           $2.73_{5}$ &   $3.4_{11}$ &  $4.41_{15}$ &   $2.65_{4}$ &  $3.62_{12}$ &    $3.29_{10}$ &       $3.1_{9}$ &   $2.97_{8}$ \\
       & 10000 &           $1.67_{6}$ &           $1.67_{5}$ &           $1.58_{4}$ &  $\mathbf{1.42_{1}}$ &           $1.47_{2}$ &  $2.76_{14}$ &  $1.94_{10}$ &   $1.58_{3}$ &  $1.97_{11}$ &     $1.89_{9}$ &      $1.75_{8}$ &   $1.75_{7}$ \\
\bottomrule
\end{tabular}
\end{adjustbox}
\begin{adjustbox}{width=1\textwidth}
\begin{tabular}{ll|llllp{3.5em}p{3.5em}p{3.5em}ll}
\toprule
       &          &        Platt &                 beta &     isotonic &           KDE & Spline natural & Spline cubic & Spline parabolic &           $IOP_{OI}$ &     $IOP_{OP}$ \\
Model & {} &              &                      &              &               &                &              &                  &                      &                \\
\midrule
resnet110 & 1000 &   $4.26_{4}$ &  $\mathbf{3.12_{1}}$ &  $6.82_{16}$ &  $43.48_{20}$ &    $6.71_{15}$ &  $6.92_{18}$ &      $6.83_{17}$ &           $4.83_{9}$ &  $200.97_{21}$ \\
       & 3000 &  $4.05_{15}$ &           $2.58_{6}$ &  $3.71_{13}$ &  $39.81_{21}$ &    $4.43_{18}$ &  $4.16_{16}$ &      $4.23_{17}$ &          $4.57_{19}$ &     $4.7_{20}$ \\
       & 10000 &  $3.64_{18}$ &          $2.87_{16}$ &   $1.87_{5}$ &  $38.07_{21}$ &    $2.78_{14}$ &  $2.89_{17}$ &      $2.84_{15}$ &           $5.1_{20}$ &    $4.93_{19}$ \\
       \hline
densenet40 & 1000 &   $6.2_{12}$ &           $4.81_{5}$ &  $8.59_{19}$ &  $71.22_{20}$ &     $7.6_{16}$ &  $8.57_{18}$ &      $7.86_{17}$ &  $\mathbf{3.99_{1}}$ &  $176.45_{21}$ \\
       & 3000 &   $3.7_{13}$ &            $2.9_{7}$ &  $4.66_{16}$ &  $64.68_{21}$ &    $5.06_{17}$ &  $5.72_{20}$ &      $5.33_{18}$ &          $3.93_{14}$ &     $5.5_{19}$ \\
       & 10000 &  $3.25_{15}$ &           $2.6_{13}$ &  $2.53_{12}$ &  $62.48_{21}$ &    $3.69_{16}$ &  $3.87_{20}$ &      $3.79_{18}$ &          $3.76_{17}$ &    $3.84_{19}$ \\
\bottomrule
\end{tabular}
\end{adjustbox}

\end{table}

\begin{table}
\caption{Comparison of calibration evaluators in estimating the reliability diagram with $\vert\ch_\mathcal{M}-c^*\vert$ on CIFAR-5m. Table compares piecewise linear methods with MSE and cross-entropy loss and $PL_{DE}$ method with degree 1 and 2. Otherwise like Table~\ref{table:chat_dist_c}.}
\label{table:chat_dist_c_CE_MSE}
\centering
\begin{adjustbox}{width=1\textwidth}
\begin{tabular}{ll|ll|ll|ll}
\toprule
       & binning &   $PL3^{CE}$ &  $PL3^{MSE}$ &        $PL_{NN}^{CE}$ & $PL_{NN}^{MSE}$ &    $PL_{DE}$ & $PL_{DE}^{2}$ \\
Model & {} &              &              &                       &                 &              &               \\
\midrule
resnet110 & cars vs rest &   $3.52_{2}$ &   $4.57_{6}$ &   $\mathbf{3.18_{1}}$ &      $3.64_{3}$ &   $3.92_{5}$ &    $3.77_{4}$ \\
       & cats vs rest &   $7.72_{2}$ &    $9.7_{5}$ &   $\mathbf{7.44_{1}}$ &       $8.3_{3}$ &  $10.17_{6}$ &    $9.47_{4}$ \\
       & dogs vs rest &    $7.0_{2}$ &   $8.93_{4}$ &   $\mathbf{6.37_{1}}$ &      $7.73_{3}$ &   $8.97_{5}$ &    $9.57_{6}$ \\
       & confidence &  $11.09_{3}$ &  $11.63_{4}$ &   $\mathbf{9.79_{1}}$ &     $10.84_{2}$ &  $11.76_{6}$ &   $11.66_{5}$ \\
\hline
densenet40 & cars vs rest &   $3.32_{3}$ &   $4.53_{6}$ &   $\mathbf{2.43_{1}}$ &      $3.21_{2}$ &   $4.21_{5}$ &    $4.02_{4}$ \\
       & cats vs rest &   $7.16_{3}$ &   $8.49_{6}$ &   $\mathbf{5.72_{1}}$ &      $6.61_{2}$ &    $7.6_{4}$ &    $8.25_{5}$ \\
       & dogs vs rest &   $6.79_{3}$ &   $7.66_{6}$ &   $\mathbf{5.02_{1}}$ &      $6.17_{2}$ &   $6.94_{4}$ &    $6.95_{5}$ \\
       & confidence &  $11.05_{2}$ &  $12.09_{6}$ &  $\mathbf{10.75_{1}}$ &     $11.56_{4}$ &  $11.37_{3}$ &   $11.99_{5}$ \\
\hline
wide32 & cars vs rest &   $3.89_{4}$ &   $5.06_{6}$ &   $\mathbf{2.62_{1}}$ &      $3.22_{2}$ &   $3.66_{3}$ &    $3.91_{5}$ \\
       & cats vs rest &   $7.43_{3}$ &   $9.29_{6}$ &   $\mathbf{6.75_{1}}$ &      $7.41_{2}$ &    $8.8_{4}$ &    $9.01_{5}$ \\
       & dogs vs rest &   $8.19_{3}$ &   $9.41_{6}$ &   $\mathbf{6.01_{1}}$ &      $6.89_{2}$ &   $8.53_{5}$ &    $8.43_{4}$ \\
       & confidence &  $10.03_{3}$ &  $10.64_{6}$ &    $\mathbf{9.2_{1}}$ &     $10.05_{5}$ &   $9.76_{2}$ &   $10.04_{4}$ \\
\hline
       & avg rank &        $2.8$ &        $5.6$ &                 $1.0$ &           $2.7$ &        $4.3$ &         $4.7$ \\
\bottomrule
\end{tabular}
\end{adjustbox}
\end{table}
\begin{table}
\caption{Comparison of calibration evaluators in estimating the true calibration error \inserted{with }$\vert ECE_\mathcal{M}-CE\vert$ on CIFAR-5m. Table compares piecewise linear methods with MSE and cross-entropy loss and $PL_{DE}$ method with degree 1 and 2. Otherwise like Table~\ref{table:ECE_abs}.}
\label{table:ECE_abs_CE_MSE}
\centering
\begin{adjustbox}{width=1\textwidth}
\begin{tabular}{ll|ll|ll|ll}
\toprule
       & binning &           $PL3^{CE}$ &          $PL3^{MSE}$ &       $PL_{NN}^{CE}$ &      $PL_{NN}^{MSE}$ &   $PL_{DE}$ & $PL_{DE}^{2}$ \\
Model & {} &                      &                      &                      &                      &             &               \\
\midrule
resnet110 & cars vs rest &  $\mathbf{1.36_{1}}$ &           $2.15_{6}$ &           $2.03_{5}$ &           $1.71_{2}$ &   $1.8_{3}$ &    $2.01_{4}$ \\
       & cats vs rest &  $\mathbf{2.58_{1}}$ &           $4.08_{6}$ &            $3.6_{5}$ &           $3.17_{2}$ &  $3.29_{3}$ &    $3.33_{4}$ \\
       & dogs vs rest &   $\mathbf{2.9_{1}}$ &            $4.0_{6}$ &           $3.55_{2}$ &           $3.74_{3}$ &  $3.96_{4}$ &    $3.99_{5}$ \\
       & confidence &  $\mathbf{4.02_{1}}$ &           $4.56_{5}$ &           $4.39_{3}$ &           $4.72_{6}$ &  $4.29_{2}$ &    $4.52_{4}$ \\
\hline
densenet40 & cars vs rest &  $\mathbf{1.38_{1}}$ &           $2.62_{6}$ &           $1.49_{3}$ &           $1.45_{2}$ &  $2.14_{4}$ &    $2.19_{5}$ \\
       & cats vs rest &           $3.38_{2}$ &           $4.82_{6}$ &            $3.6_{4}$ &  $\mathbf{3.38_{1}}$ &  $3.45_{3}$ &    $3.66_{5}$ \\
\bottomrule
\end{tabular}
\end{adjustbox}
\end{table}

\begin{table}
\caption{Comparison of calibration evaluators in estimating the reliability diagram with $\vert\ch_\mathcal{M}-c^*\vert$ on CIFAR-5m. Table compares equal-size binning with equal-width binning. Otherwise like Table~\ref{table:chat_dist_c}.}
\label{table:chat_dist_c_es_ew}
\centering
\begin{adjustbox}{width=1\textwidth}
\begin{tabular}{ll|ll|ll|ll|ll}
\toprule
       & binning &    $EW_{10}$ &             $ES_{10}$ &    $EW_{15}$ &            $ES_{15}$ &    $EW_{20}$ &            $ES_{20}$ &            $EW_{CV}$ &             $ES_{CV}$ \\
Model & {} &              &                       &              &                      &              &                      &                      &                       \\
\midrule
resnet110 & cars vs rest &   $4.77_{5}$ &            $4.38_{4}$ &   $5.13_{7}$ &           $3.78_{2}$ &   $5.41_{8}$ &  $\mathbf{3.71_{1}}$ &           $4.94_{6}$ &             $3.9_{3}$ \\
       & cats vs rest &  $11.28_{5}$ &   $\mathbf{8.94_{1}}$ &  $12.03_{7}$ &           $10.4_{4}$ &  $12.74_{8}$ &          $10.37_{3}$ &          $11.49_{6}$ &           $10.04_{2}$ \\
       & dogs vs rest &  $10.49_{6}$ &   $\mathbf{8.28_{1}}$ &  $11.86_{7}$ &           $9.24_{3}$ &  $12.96_{8}$ &           $9.24_{4}$ &          $10.16_{5}$ &            $9.07_{2}$ \\
       & confidence &  $13.11_{2}$ &  $\mathbf{12.69_{1}}$ &  $14.95_{6}$ &          $14.29_{4}$ &   $16.7_{8}$ &          $15.77_{7}$ &          $14.73_{5}$ &           $13.86_{3}$ \\
\hline
densenet40 & cars vs rest &   $4.67_{3}$ &             $7.6_{8}$ &   $4.99_{5}$ &           $4.95_{4}$ &   $5.25_{7}$ &           $5.15_{6}$ &           $4.65_{2}$ &   $\mathbf{3.79_{1}}$ \\
       & cats vs rest &  $10.28_{5}$ &           $10.04_{4}$ &  $11.43_{6}$ &          $11.93_{7}$ &  $12.69_{8}$ &           $9.61_{3}$ &  $\mathbf{8.56_{1}}$ &            $9.12_{2}$ \\
\bottomrule
\end{tabular}
\end{adjustbox}
\end{table}
\begin{table}
\caption{Comparison of calibration evaluators in estimating the true calibration error \inserted{with }$\vert ECE_\mathcal{M}-CE\vert$ on CIFAR-5m. Table compares equal-size binning with equal-width binning. Otherwise like Table~\ref{table:ECE_abs}.}
\label{table:ECE_abs_es_ew}
\centering
\begin{adjustbox}{width=1\textwidth}
\begin{tabular}{ll|ll|ll|ll|ll}
\toprule
       & binning &            $EW_{10}$ &            $ES_{10}$ &   $EW_{15}$ &            $ES_{15}$ &   $EW_{20}$ &            $ES_{20}$ &   $EW_{CV}$ &   $ES_{CV}$ \\
Model & {} &                      &                      &             &                      &             &                      &             &             \\
\midrule
resnet110 & cars vs rest &  $\mathbf{1.46_{1}}$ &           $1.99_{6}$ &  $1.46_{2}$ &           $1.85_{5}$ &  $1.63_{3}$ &           $1.85_{4}$ &  $2.86_{8}$ &  $2.84_{7}$ \\
       & cats vs rest &           $2.81_{6}$ &  $\mathbf{2.32_{1}}$ &  $2.75_{5}$ &           $2.35_{2}$ &  $2.75_{4}$ &           $2.37_{3}$ &  $4.09_{8}$ &  $3.16_{7}$ \\
       & dogs vs rest &           $2.98_{3}$ &           $3.29_{4}$ &  $3.49_{6}$ &  $\mathbf{2.83_{1}}$ &  $3.47_{5}$ &           $2.88_{2}$ &  $4.04_{7}$ &  $4.24_{8}$ \\
       & confidence &  $\mathbf{4.02_{1}}$ &           $4.16_{3}$ &  $4.03_{2}$ &           $4.45_{6}$ &  $4.26_{4}$ &           $4.29_{5}$ &  $7.18_{8}$ &  $6.68_{7}$ \\
\hline
densenet40 & cars vs rest &           $1.46_{4}$ &           $1.23_{2}$ &  $1.56_{5}$ &  $\mathbf{1.11_{1}}$ &  $1.72_{6}$ &           $1.29_{3}$ &  $2.67_{8}$ &  $2.31_{7}$ \\
       & cats vs rest &           $2.91_{3}$ &           $3.28_{4}$ &  $3.28_{5}$ &  $\mathbf{2.63_{1}}$ &   $3.8_{6}$ &           $2.71_{2}$ &  $4.59_{7}$ &  $4.75_{8}$ \\
\bottomrule
\end{tabular}
\end{adjustbox}
\end{table}

\begin{table}
\caption{Comparison of calibration evaluators in estimating the reliability diagram with $\vert\ch_\mathcal{M}-c^*\vert$ on CIFAR-5m. Table compares using the cross-validation with regularisation to no regularisation. Otherwise like Table~\ref{table:chat_dist_c}.}
\label{table:chat_dist_c_trick}
\centering
\begin{adjustbox}{width=1\textwidth}
\begin{tabular}{ll|ll|ll|ll|ll}
\toprule
       & binning & $PL_{NN}^{CE}-notrick$ &       $PL_{NN}^{CE}$ & $PL3^{CE}-notrick$ &   $PL3^{CE}$ & $PL_{DE}-notrick$ &    $PL_{DE}$ & $ES_{CV}-notrick$ &    $ES_{CV}$ \\
Model & {} &                        &                      &                    &              &                   &              &                   &              \\
\midrule
resnet110 & cars vs rest &    $\mathbf{3.12_{1}}$ &           $3.18_{2}$ &         $3.47_{3}$ &   $3.52_{4}$ &        $3.89_{6}$ &   $3.92_{8}$ &        $3.83_{5}$ &    $3.9_{7}$ \\
       & cats vs rest &              $7.5_{2}$ &  $\mathbf{7.44_{1}}$ &         $7.81_{4}$ &   $7.72_{3}$ &       $10.15_{7}$ &  $10.17_{8}$ &       $10.04_{6}$ &  $10.04_{5}$ \\
       & dogs vs rest &             $6.56_{2}$ &  $\mathbf{6.37_{1}}$ &          $7.0_{3}$ &    $7.0_{4}$ &        $8.87_{5}$ &   $8.97_{6}$ &        $9.29_{8}$ &   $9.07_{7}$ \\
       & confidence &             $9.92_{2}$ &  $\mathbf{9.79_{1}}$ &        $10.71_{3}$ &  $11.09_{4}$ &       $11.53_{5}$ &  $11.76_{6}$ &       $14.16_{8}$ &  $13.86_{7}$ \\
\hline
densenet40 & cars vs rest &             $2.43_{2}$ &  $\mathbf{2.43_{1}}$ &          $3.3_{3}$ &   $3.32_{4}$ &        $4.22_{8}$ &   $4.21_{7}$ &        $3.71_{5}$ &   $3.79_{6}$ \\
       & cats vs rest &             $5.89_{2}$ &  $\mathbf{5.72_{1}}$ &         $7.16_{3}$ &   $7.16_{4}$ &         $7.7_{6}$ &    $7.6_{5}$ &        $9.25_{8}$ &   $9.12_{7}$ \\
\bottomrule
\end{tabular}
\end{adjustbox}
\end{table}
\begin{table}
\caption{Comparison of calibration evaluators in estimating the true calibration error \inserted{with }$\vert ECE_\mathcal{M}-CE\vert$ on CIFAR-5m. Table compares using the cross-validation with regularisation to no regularisation. Otherwise like Table~\ref{table:ECE_abs}.}
\label{table:ECE_abs_trick}
\centering
\begin{adjustbox}{width=1\textwidth}
\begin{tabular}{ll|ll|ll|ll|ll}
\toprule
       & binning & $PL_{NN}^{CE}-notrick$ & $PL_{NN}^{CE}$ &   $PL3^{CE}-notrick$ &           $PL3^{CE}$ & $PL_{DE}-notrick$ &            $PL_{DE}$ & $ES_{CV}-notrick$ &   $ES_{CV}$ \\
Model & {} &                        &                &                      &                      &                   &                      &                   &             \\
\midrule
resnet110 & cars vs rest &              $1.9_{5}$ &     $2.03_{6}$ &  $\mathbf{1.35_{1}}$ &           $1.36_{2}$ &        $1.78_{3}$ &            $1.8_{4}$ &        $2.81_{7}$ &  $2.84_{8}$ \\
       & cats vs rest &              $3.7_{8}$ &      $3.6_{7}$ &           $2.77_{2}$ &  $\mathbf{2.58_{1}}$ &        $3.31_{6}$ &           $3.29_{5}$ &        $3.09_{3}$ &  $3.16_{4}$ \\
       & dogs vs rest &             $3.14_{3}$ &     $3.55_{4}$ &  $\mathbf{2.72_{1}}$ &            $2.9_{2}$ &         $3.8_{5}$ &           $3.96_{6}$ &        $4.26_{8}$ &  $4.24_{7}$ \\
       & confidence &              $4.2_{4}$ &     $4.39_{6}$ &  $\mathbf{3.88_{1}}$ &           $4.02_{2}$ &        $4.17_{3}$ &           $4.29_{5}$ &        $5.56_{7}$ &  $6.68_{8}$ \\
\hline
densenet40 & cars vs rest &             $1.44_{3}$ &     $1.49_{4}$ &  $\mathbf{1.37_{1}}$ &           $1.38_{2}$ &        $2.14_{5}$ &           $2.14_{6}$ &        $2.26_{7}$ &  $2.31_{8}$ \\
       & cats vs rest &    $\mathbf{3.35_{1}}$ &      $3.6_{6}$ &           $3.47_{4}$ &           $3.38_{2}$ &        $3.49_{5}$ &           $3.45_{3}$ &        $4.71_{7}$ &  $4.75_{8}$ \\
\bottomrule
\end{tabular}
\end{adjustbox}
\end{table}

\begin{table}
\caption{Comparison standard deviations of calibration evaluators in estimating the reliability diagram with $\vert\ch_\mathcal{M}-c^*\vert$ on CIFAR-5m. Otherwise like Table~\ref{table:chat_dist_c}. Table is shown as two lines for better readability.}
\label{table:chat_dist_c_std}
\centering

\begin{adjustbox}{width=1\textwidth}
\begin{tabular}{ll|lllllll|lllll}
\toprule
       & binning &    $ES_{10}$ &    $ES_{15}$ &    $ES_{20}$ &    $ES_{25}$ &    $ES_{30}$ &         $ES_{sweep}$ &    $ES_{CV}$ &   $PL3^{CE}$ &  $PL3^{MSE}$ & $PL_{NN}^{CE}$ &      $PL_{NN}^{MSE}$ &    $PL_{DE}$ \\
Model & {} &              &              &              &              &              &                      &              &              &              &                &                      &              \\
\midrule
resnet110 & cars vs rest &   $0.42_{7}$ &   $0.29_{2}$ &   $0.38_{5}$ &  $0.49_{13}$ &   $0.42_{8}$ &          $0.62_{16}$ &  $0.53_{14}$ &  $0.54_{15}$ &  $0.64_{17}$ &     $0.39_{6}$ &          $0.71_{19}$ &  $0.66_{18}$ \\
       & cats vs rest &  $0.83_{14}$ &   $0.51_{4}$ &   $0.71_{8}$ &   $0.39_{2}$ &   $0.55_{5}$ &          $0.98_{19}$ &  $0.75_{11}$ &   $0.9_{16}$ &  $1.13_{20}$ &     $0.68_{7}$ &           $0.68_{6}$ &  $0.94_{18}$ \\
       & dogs vs rest &  $1.39_{12}$ &    $1.0_{7}$ &  $1.94_{18}$ &  $1.61_{15}$ &  $1.88_{17}$ &           $0.88_{5}$ &  $2.22_{21}$ &   $0.96_{6}$ &  $1.56_{14}$ &     $0.69_{3}$ &          $1.18_{10}$ &  $1.22_{11}$ \\
       & confidence &   $0.97_{8}$ &  $1.52_{12}$ &  $1.86_{16}$ &  $1.69_{14}$ &  $1.76_{15}$ &           $0.91_{6}$ &  $3.05_{21}$ &   $0.57_{3}$ &   $1.11_{9}$ &     $0.93_{7}$ &          $1.14_{10}$ &  $1.95_{17}$ \\
       \hline
densenet40 & cars vs rest &  $2.52_{20}$ &  $0.97_{16}$ &  $2.09_{18}$ &  $2.01_{17}$ &   $2.2_{19}$ &          $0.83_{12}$ &  $0.76_{10}$ &   $0.65_{8}$ &   $0.47_{5}$ &     $0.34_{4}$ &           $0.51_{6}$ &   $0.9_{14}$ \\
       & cats vs rest &  $1.37_{16}$ &  $1.38_{17}$ &  $0.73_{10}$ &  $1.67_{19}$ &  $1.54_{18}$ &           $1.8_{20}$ &  $0.77_{12}$ &   $0.58_{7}$ &   $1.3_{15}$ &      $0.3_{2}$ &           $0.35_{3}$ &   $0.56_{6}$ \\
\bottomrule
\end{tabular}
\end{adjustbox}
\begin{adjustbox}{width=1\textwidth}
\begin{tabular}{ll|llllp{3.5em}p{3.5em}p{3.5em}ll}
\toprule
       &          &                Platt &                 beta &             isotonic &          KDE &       Spline natural & Spline cubic &     Spline parabolic &           $IOP_{OI}$ &   $IOP_{OP}$ \\
Model & {} &                      &                      &                      &              &                      &              &                      &                      &              \\
\midrule
resnet110 & cars vs rest &          $0.77_{20}$ &          $0.47_{11}$ &  $\mathbf{0.16_{1}}$ &  $4.94_{21}$ &           $0.46_{9}$ &  $0.48_{12}$ &          $0.47_{10}$ &           $0.35_{4}$ &    $0.3_{3}$ \\
       & cats vs rest &           $0.9_{17}$ &          $0.73_{10}$ &           $0.41_{3}$ &  $1.27_{21}$ &           $0.71_{9}$ &  $0.83_{13}$ &          $0.84_{15}$ &  $\mathbf{0.38_{1}}$ &  $0.79_{12}$ \\
       & dogs vs rest &           $0.79_{4}$ &           $1.02_{8}$ &           $1.15_{9}$ &  $1.39_{13}$ &          $1.95_{19}$ &  $1.83_{16}$ &          $2.01_{20}$ &  $\mathbf{0.49_{1}}$ &   $0.58_{2}$ \\
       & confidence &  $\mathbf{0.37_{1}}$ &           $0.38_{2}$ &           $1.2_{11}$ &   $0.74_{5}$ &          $2.53_{20}$ &  $2.16_{18}$ &           $2.2_{19}$ &          $1.64_{13}$ &    $0.7_{4}$ \\
       \hline
densenet40 & cars vs rest &           $0.33_{3}$ &           $0.65_{9}$ &  $\mathbf{0.29_{1}}$ &  $3.86_{21}$ &          $0.89_{13}$ &  $0.82_{11}$ &          $0.96_{15}$ &           $0.31_{2}$ &   $0.53_{7}$ \\
       & cats vs rest &          $0.78_{13}$ &           $0.68_{9}$ &          $0.74_{11}$ &  $3.57_{21}$ &           $0.47_{5}$ &   $0.66_{8}$ &  $\mathbf{0.29_{1}}$ &          $0.99_{14}$ &   $0.45_{4}$ \\
\bottomrule
\end{tabular}
\end{adjustbox}

\end{table}
\begin{table}
\caption{Comparison of standard deviations of calibration evaluators in estimating the true calibration error \inserted{with }$\vert ECE_\mathcal{M}-CE\vert$ on CIFAR-5m. Otherwise like Table~\ref{table:ECE_abs}. Table is shown as two lines for better readability.}
\label{table:ECE_abs_std}
\centering

\begin{adjustbox}{width=1\textwidth}
\begin{tabular}{ll|lllllll|lllll}
\toprule
       & binning &    $ES_{10}$ &    $ES_{15}$ &    $ES_{20}$ &    $ES_{25}$ &    $ES_{30}$ & $ES_{sweep}$ &            $ES_{CV}$ &   $PL3^{CE}$ &  $PL3^{MSE}$ &       $PL_{NN}^{CE}$ & $PL_{NN}^{MSE}$ &    $PL_{DE}$ \\
Model & {} &              &              &              &              &              &              &                      &              &              &                      &                 &              \\
\midrule
resnet110 & cars vs rest &  $0.73_{14}$ &  $0.65_{11}$ &  $0.66_{12}$ &   $0.6_{10}$ &  $0.72_{13}$ &  $1.13_{19}$ &          $1.46_{20}$ &    $0.5_{6}$ &  $0.81_{17}$ &           $0.41_{3}$ &      $0.33_{2}$ &  $0.74_{15}$ \\
       & cats vs rest &  $1.14_{17}$ &   $0.71_{6}$ &   $0.76_{8}$ &  $0.92_{12}$ &   $0.67_{5}$ &  $1.22_{19}$ &          $1.69_{21}$ &  $0.89_{10}$ &   $0.67_{4}$ &          $1.12_{16}$ &     $1.16_{18}$ &  $1.38_{20}$ \\
       & dogs vs rest &  $1.35_{15}$ &   $0.58_{4}$ &  $1.41_{16}$ &  $0.89_{10}$ &  $1.43_{17}$ &  $1.19_{12}$ &          $1.26_{13}$ &   $0.63_{6}$ &  $1.65_{18}$ &           $0.39_{2}$ &      $0.69_{7}$ &  $1.08_{11}$ \\
       & confidence &   $0.86_{3}$ &   $1.28_{7}$ &   $1.16_{5}$ &  $1.72_{16}$ &   $2.9_{21}$ &  $1.37_{10}$ &          $1.86_{17}$ &  $1.51_{11}$ &   $1.24_{6}$ &           $1.36_{9}$ &      $0.78_{2}$ &  $1.69_{15}$ \\
       \hline
densenet40 & cars vs rest &  $0.61_{12}$ &  $0.72_{18}$ &  $0.63_{15}$ &   $0.59_{8}$ &   $0.54_{6}$ &  $0.76_{19}$ &          $0.64_{16}$ &   $0.6_{10}$ &  $0.63_{14}$ &   $\mathbf{0.2_{1}}$ &       $0.3_{4}$ &  $0.88_{20}$ \\
       & cats vs rest &  $0.69_{10}$ &  $1.02_{17}$ &   $0.35_{2}$ &  $0.94_{16}$ &  $0.87_{14}$ &  $1.21_{18}$ &          $1.62_{19}$ &   $0.7_{11}$ &  $1.71_{20}$ &          $0.72_{13}$ &       $0.6_{7}$ &  $0.71_{12}$ \\
\bottomrule
\end{tabular}
\end{adjustbox}
\begin{adjustbox}{width=1\textwidth}
\begin{tabular}{ll|llllp{3.5em}p{3.5em}p{3.5em}ll}
\toprule
       &          &                Platt &                 beta &             isotonic &                  KDE & Spline natural &         Spline cubic &     Spline parabolic &           $IOP_{OI}$ &   $IOP_{OP}$ \\
Model & {} &                      &                      &                      &                      &                &                      &                      &                      &              \\
\midrule
resnet110 & cars vs rest &          $0.84_{18}$ &          $0.78_{16}$ &            $0.5_{5}$ &          $4.96_{21}$ &     $0.54_{7}$ &           $0.56_{8}$ &           $0.57_{9}$ &  $\mathbf{0.19_{1}}$ &   $0.43_{4}$ \\
       & cats vs rest &           $0.9_{11}$ &           $0.75_{7}$ &  $\mathbf{0.43_{1}}$ &           $0.83_{9}$ &    $0.97_{13}$ &          $1.01_{15}$ &          $1.01_{14}$ &           $0.54_{3}$ &   $0.53_{2}$ \\
       & dogs vs rest &  $\mathbf{0.29_{1}}$ &           $0.59_{5}$ &          $1.29_{14}$ &           $0.83_{9}$ &    $1.69_{19}$ &           $1.9_{21}$ &          $1.83_{20}$ &           $0.56_{3}$ &   $0.78_{8}$ \\
       & confidence &          $1.68_{14}$ &           $1.04_{4}$ &           $1.32_{8}$ &  $\mathbf{0.51_{1}}$ &    $2.14_{19}$ &          $2.12_{18}$ &          $1.68_{13}$ &          $1.66_{12}$ &  $2.18_{20}$ \\
       \hline
densenet40 & cars vs rest &           $0.55_{7}$ &          $0.63_{13}$ &           $0.28_{3}$ &          $3.85_{21}$ &    $0.61_{11}$ &            $0.6_{9}$ &          $0.66_{17}$ &           $0.23_{2}$ &   $0.49_{5}$ \\
       & cats vs rest &          $0.92_{15}$ &           $0.61_{8}$ &           $0.53_{6}$ &          $3.43_{21}$ &     $0.51_{5}$ &  $\mathbf{0.35_{1}}$ &            $0.4_{4}$ &           $0.62_{9}$ &   $0.38_{3}$ \\
\bottomrule
\end{tabular}
\end{adjustbox}

\end{table}

\subsection{Results of Synthetic Experiments}


The results of synthetic experiments indicate that beta calibration and Platt scaling are very good when the parametric family is able to approximate \modified{$c^*$} closely (Table~\ref{table:syn_uniform_calmap}). However, these methods can fail for more difficult shapes (i.e `stairs'). Piecewise linear methods are following in performance and the binning methods are in last place. To achieve a good score for all the shapes, the piecewise linear methods should be picked. The repeating results for isotonic in Table~\ref{table:syn_uniform_calmap} are not a bug, but a peculiarity arising from the way synthetic data were generated.


Performance of \modified{with regards to} $\vert ECE_\mathcal{M}-CE\vert$ is similar to 
$\vert\ch_\mathcal{M}-c^*\vert$ (Table~\ref{table:syn_uniform_ece}). The best measure is $PL_{DE}$ and measures Platt scaling and $ES_{CV}$ are following.
Again, Platt scaling and beta calibration are failing at `stairs' dataset.

In Table~\ref{table:syn_uniform_ece_sq} the performance of estimating 
$\vert ECE_\mathcal{M}-CE\vert^2$ is compared so that KCE could also be included. The unbiased version of KCE with RBFKernel is used. KCE performs worse than the other selected methods.

In Table~\ref{table:syn_uniform_spearman} Spearman correlation is used to see how well the measures are able to rank different models. All the measures are really good, the only exception is `stairs', where Platt scaling and beta calibration are not as good. 

\begin{table}
\caption{Comparison of calibration evaluators in estimating the reliability diagram with $\vert\ch_\mathcal{M}-c^*\vert$. Evaluated on synthetic data. The distance has been calculated on $10^6$ test data. The table displays thousandths.}
\label{table:syn_uniform_calmap}
\centering
\begin{adjustbox}{width=1\textwidth}
\begin{tabular}{l|lll|lllll|lll}
\toprule
binning &     $ES_{15}$ & $ES_{sweep}$ &    $ES_{CV}$ &   $PL3^{CE}$ &  $PL3^{MSE}$ &        $PL_{NN}^{CE}$ & $PL_{NN}^{MSE}$ &    $PL_{DE}$ &                 Platt &                  beta &      isotonic \\
\midrule
square   &  $24.76_{10}$ &  $20.46_{9}$ &  $17.33_{8}$ &   $9.98_{2}$ &  $10.98_{3}$ &           $13.07_{6}$ &     $12.78_{5}$ &  $14.88_{7}$ &   $\mathbf{9.48_{1}}$ &           $11.04_{4}$ &  $25.45_{11}$ \\
sqrt     &  $24.78_{10}$ &  $20.43_{9}$ &  $16.87_{8}$ &  $13.29_{4}$ &  $14.44_{7}$ &           $13.43_{5}$ &     $14.06_{6}$ &  $13.04_{3}$ &           $11.82_{2}$ &  $\mathbf{11.18_{1}}$ &  $25.45_{11}$ \\
beta1    &   $25.3_{10}$ &  $20.98_{9}$ &  $18.96_{8}$ &  $14.76_{4}$ &  $15.27_{5}$ &           $16.87_{6}$ &     $17.55_{7}$ &  $13.69_{3}$ &  $\mathbf{11.32_{1}}$ &           $12.49_{2}$ &  $25.45_{11}$ \\
beta2    &  $25.54_{11}$ &  $21.52_{8}$ &  $21.58_{9}$ &  $14.51_{4}$ &  $14.41_{3}$ &           $15.26_{5}$ &     $15.31_{6}$ &  $18.49_{7}$ &  $\mathbf{12.87_{1}}$ &           $14.33_{2}$ &  $25.45_{10}$ \\
stairs   &   $26.79_{9}$ &  $23.24_{6}$ &  $23.55_{7}$ &  $18.72_{3}$ &  $19.32_{4}$ &  $\mathbf{17.89_{1}}$ &     $17.95_{2}$ &  $20.09_{5}$ &          $50.44_{11}$ &          $36.14_{10}$ &   $25.45_{8}$ \\
\hline
avg rank &        $10.0$ &        $8.2$ &        $8.0$ &        $3.4$ &        $4.4$ &                 $4.6$ &           $5.2$ &        $5.0$ &                 $3.2$ &                 $3.8$ &        $10.2$ \\
\bottomrule
\end{tabular}
\end{adjustbox}
\end{table}
\begin{table}
\caption{Comparison of calibration evaluators in estimating the true calibration error \inserted{with }$\vert ECE_\mathcal{M}-CE\vert$. Evaluated on synthetic data. The table displays thousandths.}
\label{table:syn_uniform_ece}
\centering
\begin{adjustbox}{width=1\textwidth}
\begin{tabular}{l|lll|lllll|lll}
\toprule
binning &            $ES_{15}$ & $ES_{sweep}$ &            $ES_{CV}$ &   $PL3^{CE}$ & $PL3^{MSE}$ & $PL_{NN}^{CE}$ & $PL_{NN}^{MSE}$ &            $PL_{DE}$ &                Platt &          beta &      isotonic \\
\midrule
square   &           $7.4_{10}$ &   $7.07_{9}$ &           $6.59_{6}$ &   $5.91_{4}$ &  $6.31_{5}$ &     $7.01_{8}$ &      $6.98_{7}$ &           $5.84_{3}$ &  $\mathbf{5.79_{1}}$ &    $5.81_{2}$ &   $8.66_{11}$ \\
sqrt     &           $6.94_{9}$ &   $6.56_{8}$ &  $\mathbf{5.92_{1}}$ &   $6.37_{4}$ &  $6.39_{5}$ &    $7.12_{10}$ &      $6.49_{7}$ &           $6.21_{2}$ &           $6.34_{3}$ &    $6.49_{6}$ &    $9.6_{11}$ \\
beta1    &          $7.78_{10}$ &   $7.19_{7}$ &           $7.11_{6}$ &   $7.72_{9}$ &  $6.83_{3}$ &     $7.46_{8}$ &      $6.96_{5}$ &  $\mathbf{5.79_{1}}$ &           $6.76_{2}$ &    $6.95_{4}$ &  $10.38_{11}$ \\
beta2    &           $7.18_{2}$ &   $8.04_{8}$ &            $8.1_{9}$ &  $8.34_{10}$ &  $7.52_{4}$ &     $7.75_{5}$ &      $7.23_{3}$ &  $\mathbf{6.88_{1}}$ &           $7.83_{6}$ &    $7.97_{7}$ &   $9.13_{11}$ \\
stairs   &  $\mathbf{7.12_{1}}$ &   $7.46_{4}$ &           $7.29_{2}$ &   $7.59_{7}$ &  $7.58_{6}$ &     $8.11_{8}$ &      $7.32_{3}$ &           $7.47_{5}$ &         $29.27_{11}$ &  $24.98_{10}$ &    $10.3_{9}$ \\
\hline
avg rank &                $6.4$ &        $7.2$ &                $4.8$ &        $6.8$ &       $4.6$ &          $7.8$ &           $5.0$ &                $2.4$ &                $4.6$ &         $5.8$ &        $10.6$ \\
\bottomrule
\end{tabular}
\end{adjustbox}
\end{table}
\begin{table}
\caption{Comparison of calibration evaluators in estimating the true calibration error \inserted{with }$\vert ECE_\mathcal{M}-CE\vert^2$. Evaluated on synthetic data. The table displays thousandths.}
\label{table:syn_uniform_ece_sq}
\centering
\begin{adjustbox}{width=1\textwidth}
\begin{tabular}{l|llll|lllll|lll}
\toprule
binning &   $ES_{15}$ & $ES_{sweep}$ &            $ES_{CV}$ &          KCE &           $PL3^{CE}$ &          $PL3^{MSE}$ & $PL_{NN}^{CE}$ & $PL_{NN}^{MSE}$ &            $PL_{DE}$ &        Platt &         beta &     isotonic \\
\midrule
square   &  $0.91_{9}$ &  $0.97_{10}$ &           $0.81_{7}$ &  $1.62_{12}$ &  $\mathbf{0.69_{1}}$ &           $0.77_{4}$ &     $0.81_{5}$ &      $0.81_{6}$ &           $0.83_{8}$ &    $0.7_{2}$ &   $0.74_{3}$ &  $1.17_{11}$ \\
sqrt     &  $0.81_{6}$ &   $0.77_{3}$ &           $0.76_{2}$ &  $1.56_{12}$ &           $0.85_{8}$ &  $\mathbf{0.74_{1}}$ &    $0.91_{10}$ &      $0.79_{5}$ &           $0.78_{4}$ &   $0.89_{9}$ &   $0.85_{7}$ &  $1.39_{11}$ \\
beta1    &   $0.9_{7}$ &   $0.92_{8}$ &            $0.8_{2}$ &   $4.0_{12}$ &          $0.97_{10}$ &           $0.81_{3}$ &     $0.94_{9}$ &      $0.82_{4}$ &  $\mathbf{0.74_{1}}$ &   $0.85_{5}$ &   $0.88_{6}$ &  $1.41_{11}$ \\
beta2    &  $1.03_{5}$ &   $1.07_{9}$ &           $0.96_{2}$ &  $3.93_{12}$ &          $1.15_{10}$ &           $1.06_{7}$ &      $1.0_{3}$ &      $1.02_{4}$ &   $\mathbf{0.9_{1}}$ &   $1.06_{6}$ &   $1.06_{8}$ &  $1.23_{11}$ \\
stairs   &   $0.9_{2}$ &   $0.91_{3}$ &  $\mathbf{0.88_{1}}$ &  $4.11_{12}$ &           $1.02_{7}$ &           $0.93_{4}$ &     $1.06_{8}$ &       $1.0_{6}$ &           $0.99_{5}$ &  $3.31_{11}$ &  $3.11_{10}$ &   $1.41_{9}$ \\
\hline
avg rank &       $5.8$ &        $6.6$ &                $2.8$ &       $12.0$ &                $7.2$ &                $3.8$ &          $7.0$ &           $5.0$ &                $3.8$ &        $6.6$ &        $6.8$ &       $10.6$ \\
\bottomrule
\end{tabular}
\end{adjustbox}
\end{table}
\begin{table}
\caption{Spearman correlations on synthetic data with uniform distribution.}
\label{table:syn_uniform_spearman}
\centering
\begin{adjustbox}{width=1\textwidth}
\begin{tabular}{l|lll|lllll|llll}
\toprule
binning &              $ES_{15}$ &           $ES_{sweep}$ &      $ES_{CV}$ &    $PL3^{CE}$ &            $PL3^{MSE}$ & $PL_{NN}^{CE}$ & $PL_{NN}^{MSE}$ &     $PL_{DE}$ &          Platt &                   beta &               isotonic \\
\midrule
square   &           $0.9965_{2}$ &          $0.9898_{10}$ &  $0.9811_{11}$ &  $0.9952_{6}$ &           $0.9965_{2}$ &   $0.9919_{9}$ &    $0.9954_{4}$ &  $0.9948_{8}$ &   $0.9954_{4}$ &           $0.9952_{7}$ &  $\mathbf{0.9965_{1}}$ \\
sqrt     &          $0.9877_{10}$ &  $\mathbf{0.9937_{1}}$ &   $0.984_{11}$ &  $0.9931_{2}$ &           $0.9921_{3}$ &   $0.9878_{9}$ &    $0.9893_{8}$ &  $0.9916_{5}$ &   $0.9919_{4}$ &           $0.9913_{6}$ &           $0.9894_{7}$ \\
beta1    &           $0.9956_{3}$ &            $0.995_{4}$ &   $0.9919_{7}$ &  $0.9842_{8}$ &           $0.9702_{9}$ &  $0.9601_{11}$ &    $0.964_{10}$ &  $0.9945_{5}$ &   $0.9945_{6}$ &  $\mathbf{0.9974_{1}}$ &           $0.9961_{2}$ \\
beta2    &           $0.9971_{7}$ &          $0.9942_{11}$ &   $0.9979_{5}$ &  $0.9984_{4}$ &  $\mathbf{0.9999_{1}}$ &   $0.9987_{3}$ &    $0.9996_{2}$ &  $0.9956_{9}$ &   $0.9971_{6}$ &           $0.9958_{8}$ &          $0.9942_{10}$ \\
stairs   &  $\mathbf{0.9965_{1}}$ &           $0.9956_{2}$ &   $0.9895_{6}$ &  $0.9888_{8}$ &            $0.989_{7}$ &   $0.9887_{9}$ &    $0.9926_{5}$ &  $0.9946_{3}$ &  $0.9549_{11}$ &          $0.9777_{10}$ &           $0.9939_{4}$ \\
\hline
avg rank &                  $4.7$ &                  $5.6$ &          $8.0$ &         $5.6$ &                  $4.5$ &          $8.2$ &           $5.9$ &         $6.0$ &          $6.3$ &                  $6.4$ &                  $4.8$ \\
\bottomrule
\end{tabular}
\end{adjustbox}
\end{table}

\subsection{Results of Real Experiments}

Table~\ref{table:real_data_biases} also shows the biases on the real data, to be compared with pseudo-real results. To achieve as valid results as possible, the pseudo-real data has the subset containing the architectures (Resnet110, WideNet32, Densenet40) and calibrated with the same methods (TempS, VecS, MSODIR, dirL2, dirODIR). The only difference is that real experiments are trained and evaluated on CIFAR-10, but the pseudo-real experiments are trained on CIFAR-5m dataset.
To calculate average bias, one needs to subtract the average true calibration error from the average estimates - the same constant for each value in the row).
For real data, we do not know this constant, but we have subtracted a value which makes the row average bias match with the corresponding row average bias of pseudo-real data. 
Thus, the true average biases are a constant shift away from these.
However, as this does not affect the ranks, we can still compare whether the ranking of biases across different methods agrees between the synthetic and real data.
There is a strong agreement, increasing confidence that the stronger methods on pseudo-real data in Tables~\ref{table:chat_dist_c} and \ref{table:ECE_abs} are also stronger on real data.

\begin{figure}
 \centering
\includegraphics[width=\textwidth]{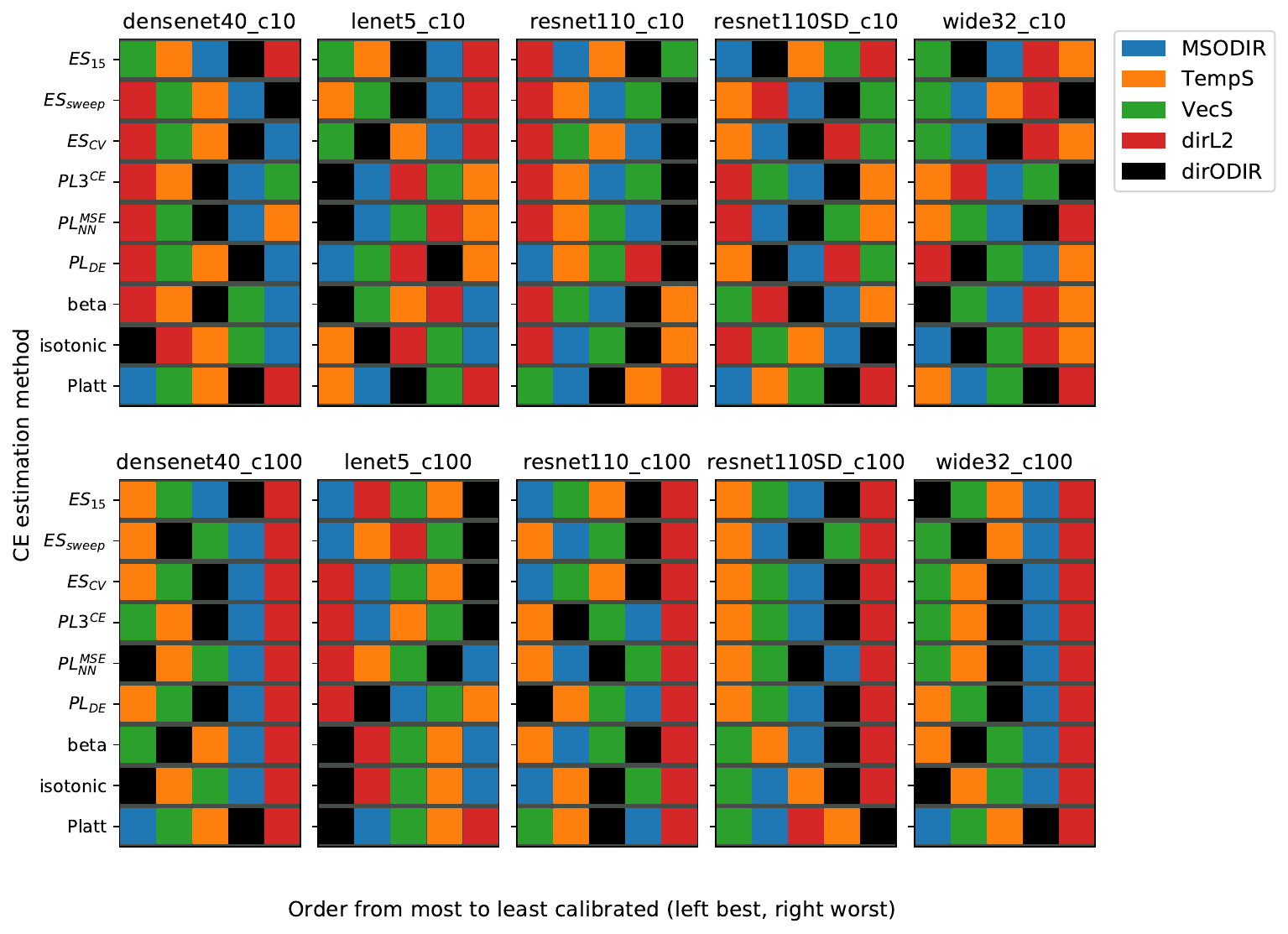}
\caption{Ordering of models from the most to the least calibrated by estimated calibration error on real dataset models (shown as subfigures), according to different evaluation methods (shown as rows within the subfigures). Test set size is 10k for each dataset.}
\label{fig:real_data_ordering}
\end{figure}

Figure~\ref{fig:real_data_ordering} depicts how different evaluation methods order models by calibration. The results are shown for 10 different real model-dataset combinations. The test set size is 10k for each example. For each model-dataset combination there are 5 models after applying post-hoc calibration methods. 

It can be seen from the figure that for all cases, the choice of the evaluation method has consequences on the ordering of models. For example, for resnet110SD\_c100, $beta$, $isotonic$ and $Platt$ order vector scaling (VecS) as the most calibrated model, while other methods order temperature scaling (TempS) as the most calibrated model.

\begin{table}
\caption{Bias calculated on the pseudo-real (CIFAR-5m) and real data. The asterisk (*) denotes that the center of real biases is unknown but they have been shifted to match the average bias of pseudo-real data covering the same 5 calibration functions and 3 models (DenseNet40, ResNet110, WideNet32). The table displays thousandths.}
\label{table:real_data_biases}
\centering
\begin{adjustbox}{width=0.8\textwidth}
\begin{tabular}{ll|lll|lll|lll}
\toprule
           & binning &    $ES_{15}$ & $ES_{sweep}$ &   $ES_{CV}$ &   $PL3^{CE}$ & $PL_{NN}^{MSE}$ &   $PL_{DE}$ &         beta &    isotonic &        Platt \\
{} & $n_{data}$ &              &              &             &              &                 &             &              &             &              \\
\midrule
CIFAR-5m & 1000 &   $2.28_{7}$ &   $1.16_{9}$ &  $2.64_{4}$ &   $4.34_{3}$ &      $1.23_{8}$ &  $4.37_{2}$ &   $2.61_{5}$ &  $8.04_{1}$ &   $2.40_{6}$ \\
           & 3000 &   $0.29_{7}$ &  $-0.50_{9}$ &  $0.89_{4}$ &   $1.69_{3}$ &      $0.86_{5}$ &  $1.70_{2}$ &   $0.31_{6}$ &  $4.76_{1}$ &  $-0.40_{8}$ \\
           & 10000 &  $-0.52_{6}$ &  $-0.83_{7}$ &  $0.23_{3}$ &  $-0.05_{5}$ &      $0.09_{4}$ &  $0.45_{2}$ &  $-0.95_{8}$ &  $2.45_{1}$ &  $-1.46_{9}$ \\
\hline
Real Data* & 1000 &   $2.43_{5}$ &   $0.03_{9}$ &  $1.55_{8}$ &   $6.47_{2}$ &      $1.99_{7}$ &  $3.87_{3}$ &   $2.16_{6}$ &  $8.11_{1}$ &   $2.46_{4}$ \\
           & 3000 &   $0.53_{6}$ &  $-1.37_{9}$ &  $0.69_{5}$ &   $2.17_{2}$ &      $0.25_{8}$ &  $1.76_{3}$ &   $0.71_{4}$ &  $4.46_{1}$ &   $0.40_{7}$ \\
           & 10000 &  $-0.25_{7}$ &  $-1.42_{9}$ &  $0.14_{3}$ &  $-0.03_{4}$ &     $-0.22_{6}$ &  $0.20_{2}$ &  $-0.11_{5}$ &  $1.65_{1}$ &  $-0.55_{8}$ \\
\bottomrule
\end{tabular}
\end{adjustbox}
\end{table}\

\subsection{Running Time of the Experiments}
The experiments were run on the pseudo-real data combining 13 calibration maps (6 was used for final results), 5 seeds, 3 models and 3 data sizes and 4 2-class experiments (3 1vsRest and 1 confidence). In total 2340 combinations for all the ECE methods, isotonic, beta, platt, kernel methods and $PL$ methods took around 6500 hours. 
Running the experiments on the synthetic data combining 5 calibration maps, 5 seeds, 21 derivates and 3 data sizes, for all the ECE methods, isotonic, beta, platt, kernel methods and $PL$ methods took around 5000 hours.
The experiments were run on the real data combining 11 model-dataset combinations, with three different data sizes with different number of subsets (total 10) and 5 calibration methods, in total 700 combinations, which took under 350 hours. 
The scripts were run in a high performance computing center using CPU processing power (Intel(R) Xeon(R) CPU E5-2660 v2 @ 2.20GHz) with up to 6GB of RAM.

\section{Limitations and Future Work}

While piecewise linear methods in both the original probability space and in the logit-logit space have improved the state-of-the-art as calibrators and as evaluators of calibration, there is room for improvement: (1) we have not experimented on non-neural classification methods to identify whether PL or PL3 or some other method would work well there; (2) when estimating calibration error, PL and PL3 would benefit from debiasing methods to further improve performance; (3) the running time of PL and PL3 should be reduced, e.g. by only considering less bins (usually 5 or less bins were used) and using 3- or 5-fold CV instead of 10-fold CV; (4) new pseudo-real datasets should be created from generative models trained on other datasets than CIFAR-10, so that true calibration maps could be estimated and calibration methods compared; and (5) new calibration map families could be created based on results with new pseudo-real datasets.


\end{appendices}


\end{document}